\documentclass[11pt, letterpaper]{template}

\usepackage[all]{hypcap}
\usepackage[authoryear, round]{natbib}
\definecolor{citecolor}{HTML}{0071bc}
\usepackage{hyperref}[citecolor=citecolor]
\usepackage{url} 
\hypersetup{
    colorlinks=true,
    citecolor=blue,
    linkcolor=magenta,
    urlcolor=magenta
}

\definecolor{darkblue}{rgb}{0, 0, 0.5}
\usepackage{booktabs}       
\usepackage{amsfonts}       
\usepackage{nicefrac}       
\usepackage{microtype}      
\usepackage{xcolor}         

\usepackage{lineno}

\usepackage{enumitem}       
\usepackage{subfig}         
\usepackage{multirow}       %
\usepackage{xspace}         
\usepackage{wrapfig}        
\usepackage{makecell}       
\usepackage{colortbl}
\usepackage{bbm}            
\usepackage{adjustbox}

\newboolean{showsection}
\setboolean{showsection}{true}
\makeatletter
\@namedef{ver@everyshi.sty}{}
\makeatother

\definecolor{custom_green}{rgb}{0.0, 0.5, 0.0}
\definecolor{custom_red}{rgb}{1.0, 0.01, 0.24}
\definecolor{custom_blue}{HTML}{C9DAF7}
\definecolor{custom_purple}{HTML}{D9D1E9}
\definecolor{customL}{HTML}{FF3E3E}  
\definecolor{customu}{HTML}{FF914D}  
\definecolor{customm}{HTML}{FFC94D}  
\definecolor{customi}{HTML}{B6E24D}  
\definecolor{customn}{HTML}{4DDC95}  
\definecolor{customa}{HTML}{4DB8FF}  
\definecolor{customD}{HTML}{FF007F}  
\definecolor{customi2}{HTML}{FF00FF}  
\definecolor{customM}{HTML}{FF9900}  
\definecolor{customO}{HTML}{0099FF}  
\definecolor{customO2}{HTML}{66FF66}  
\definecolor{title_blue}{HTML}{204899} 
\definecolor{cite_blue}{HTML}{044dc1}  
\definecolor{cite_purple}{HTML}{7406a7}  
\hypersetup{
    colorlinks = true,
    citecolor = {cite_blue},
    linkcolor = magenta,
    urlcolor = magenta,
}
\usepackage[most,skins,theorems]{tcolorbox}
\tcbset{
  aibox/.style={
    width=\linewidth,
    top=8pt,
    bottom=4pt,
    colback=blue!6!white,
    colframe=black,
    colbacktitle=black,
    enhanced,
    center,
    attach boxed title to top left={yshift=-0.1in,xshift=0.15in},
    boxed title style={boxrule=0pt,colframe=white,},
  }
}
\newtcolorbox{AIbox}[2][]{aibox,title=#2,#1}

\title{
\centering {\raisebox{-0.45em}{\includegraphics[height=1.6em]{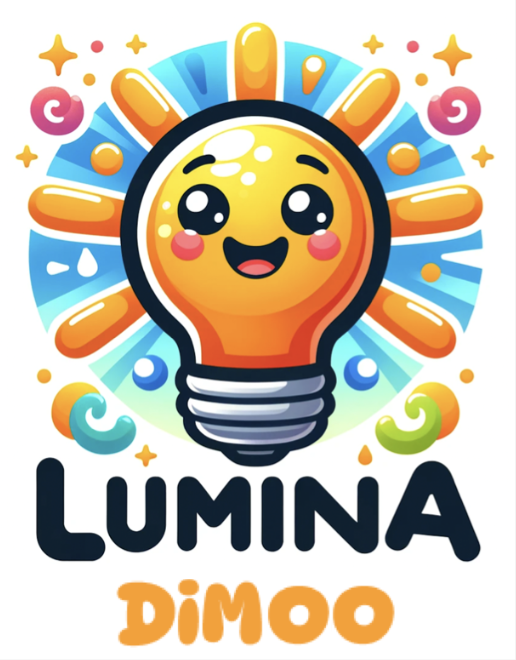}} 
\fontsize{26}{24}\selectfont \textcolor{customL}{L} \textcolor{customu}{u} \textcolor{customm}{m} \textcolor{customi}{i} \textcolor{customn}{n} \textcolor{customa}{a} \textcolor{black}{-} \textcolor{customD}{D} \textcolor{customi2}{i} \textcolor{customM}{M} \textcolor{customO}{O} \textcolor{customO2}{O}} \\
An Omni Diffusion Large Language Model for Multi-Modal Generation and Understanding}

\author[1,2,5,$\clubsuit$]{Yi Xin}
\author[1,4,$\clubsuit$]{Qi Qin}
\author[1,3]{Siqi Luo}
\author[1,3]{Kaiwen Zhu}
\author[1,7]{Juncheng Yan}
\author[3]{Yan Tai}
\author[1,3]{Jiayi Lei}
\author[1]{Yuewen Cao}
\author[1]{\\Keqi Wang}
\author[2]{Yibin Wang}
\author[1]{Jinbin Bai}
\author[1]{Qian Yu}
\author[1]{Dengyang Jiang}
\author[1]{Yuandong Pu}
\author[5]{Haoxing Chen}
\author[6]{Le Zhuo}
\author[1]{\\Junjun He}
\author[1]{Gen Luo}
\author[1]{Tianbin Li}
\author[1]{Ming Hu}
\author[1]{Jin Ye}
\author[1]{Shenglong Ye}
\author[1]{Bo Zhang}
\author[4]{Chang Xu}
\author[1]{Wenhai Wang}
\author[1,6]{\\Hongsheng Li}
\author[1,3]{Guangtao Zhai}
\author[6,1]{Tianfan Xue}
\author[1,$\dag$]{Bin Fu}
\author[3,2,$\dag$]{Xiaohong Liu}
\author[1,$\dag$]{Yu Qiao}
\author[1,$\dag$]{Yihao Liu}

\affil[1]{Shanghai AI Laboratory}
\affil[2]{Shanghai Innovation Institute}
\affil[3]{Shanghai Jiao Tong University}
\affil[4]{The University of Sydney}
\affil[5]{Nanjing University}
\affil[6]{The Chinese University of Hong Kong}
\affil[7]{Tsinghua University}

\correspondingauthor{$^\clubsuit$ Equal Contribution.}

\begin{abstract}
\vspace{-0.05in}
\textcolor{title_blue}{\textbf{Abstract—}}We introduce Lumina-DiMOO, an open-source foundational model for seamless multi-modal generation and understanding. 
Lumina-DiMOO sets itself apart from prior unified models by utilizing a fully discrete diffusion modeling to handle inputs and outputs across various modalities.
This innovative approach allows Lumina-DiMOO to achieve higher sampling efficiency compared to previous autoregressive (AR) or hybrid AR-Diffusion paradigms and adeptly support a broad spectrum of multi-modal tasks, including text-to-image generation, image-to-image generation (e.g., image editing, subject-driven generation, and image inpainting, etc.), as well as image understanding. 
Lumina-DiMOO achieves state-of-the-art performance on multiple benchmarks, surpassing existing open-source unified multi-modal models. 
To foster further advancements in multi-modal and discrete diffusion model research, we release our code and checkpoints to the community.

\end{abstract}

\begin{document}

\maketitle

\begin{figure*}[!bht]
    \centering
    \begin{picture}(0,345)
    \put(-238,1){
    \includegraphics[width=1.0\linewidth]{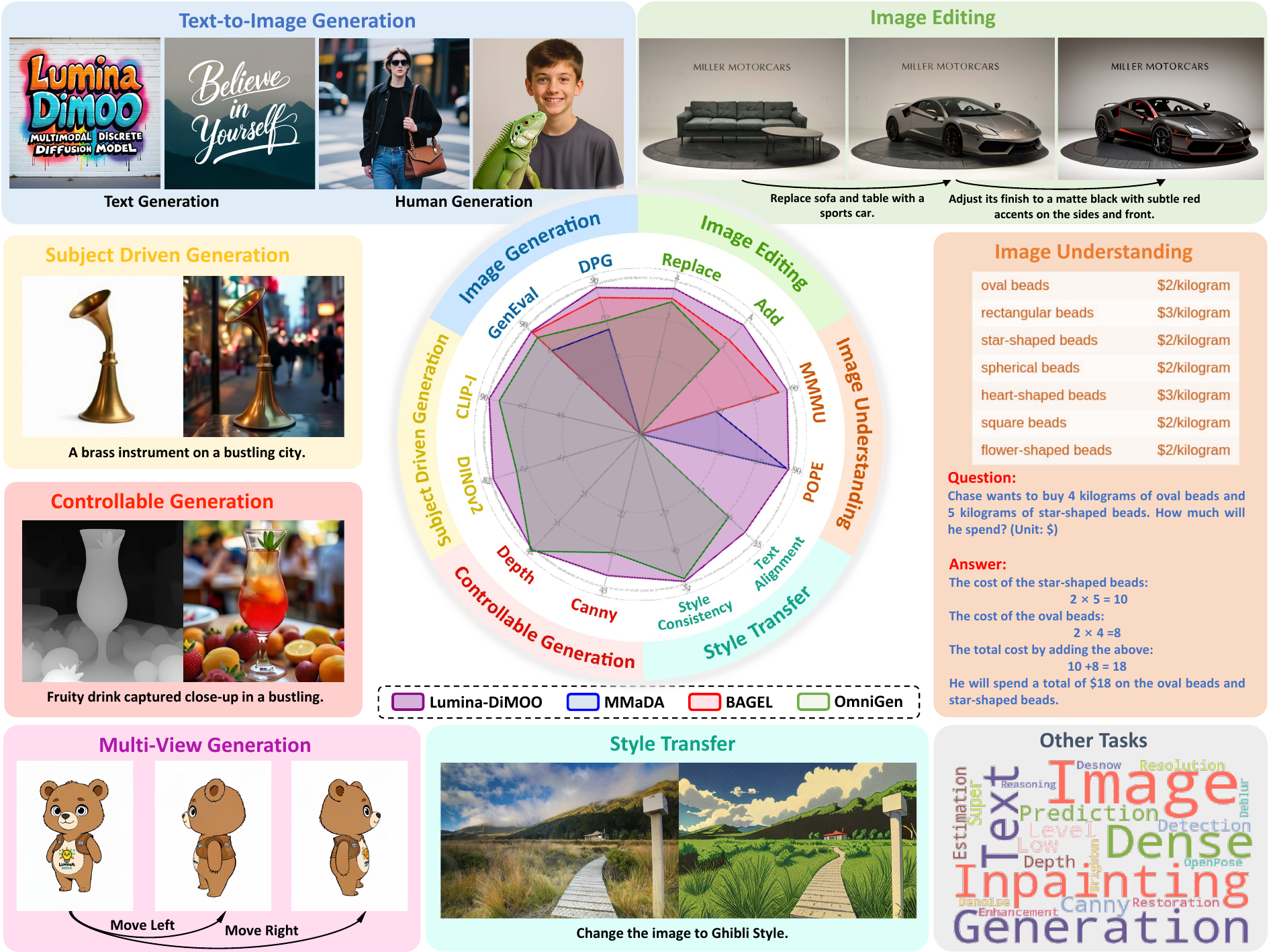}
    }
    \end{picture}
    \vspace{-0.1in}
    \caption{\centering \textbf{Overview of Lumina-DiMOO's Versatile Multi-Modal Capabilities and Superior Performance.}}
    \label{fig:abs_case}
    \vspace{-0.5in}
\end{figure*}
\section{Introduction}
Recent advancements in Large Language Models (LLMs) have markedly improved their ability to tackle multi-modal understanding tasks. 
Efforts such as the LLaVA series~\citep{liu2023llava,liu2023improvedllava,liu2024llavanext}, QwenVL series~\citep{Qwen-VL,Qwen2-VL,Qwen2.5-VL}, and InternVL series~\citep{chen2024internvl,chen2024far,zhu2025internvl3,wang2025internvl3_5} demonstrated remarkable exceptional visual comprehension performance. 
Concurrently, progress in text-to-image generation models, including diffusion-based methods \citep{2023SDXL,dalle3,xie2024sana,flux2024,zhuo2024lumina,yi2024towards,bai2024meissonic,LuminaImage2024} and more recent autoregressive approaches \citep{wang2024emu3,liu2024lumina,xin2025lumina,chen2025januspro,xin2025resurrect}, has significantly advanced the generation of high-quality images. 
Building upon these foundational models, various downstream tasks has been explored, such as image editing~\citep{yu2025anyedit}, multi-view generation~\citep{huang2024mvadapter}, and controllable generation~\citep{zhang2023adding}. 
These advancements have accelerated the convergence towards unified multi-modal generation and understanding modeling, aiming to integrate diverse capabilities into a single, end-to-end architecture, thereby contributing to the pursuit of artificial general intelligence (AGI).

\begin{figure}
    \centering
    \includegraphics[width=1.0\linewidth]{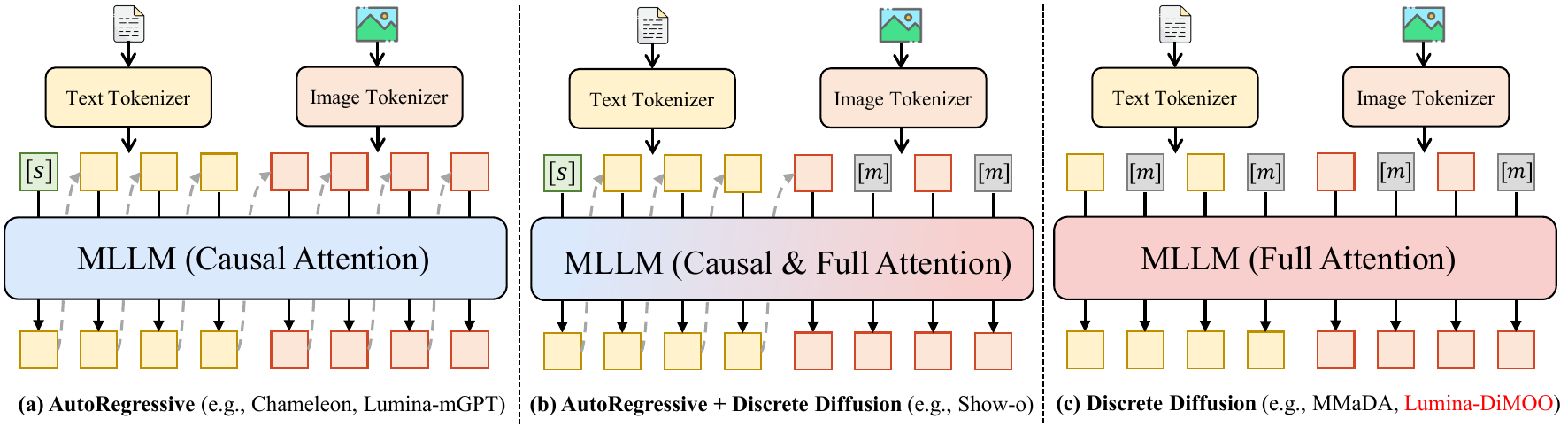}
    \caption{\textbf{Characteristics Comparison Among Existing Unified Models.} The overall architecture has transitioned from the initial pure autoregressive (AR), which also involved adding Diffusion Head after AR, to a combination of AR and discrete diffusion, and ultimately to the current model using pure discrete diffusion.}
    \label{fig:difference}
    \vspace{-0.1in}
\end{figure}

To develop unified multi-modal models, various paradigms have been explored. As shown in Figure~\ref{fig:difference}\textcolor{magenta}{(a)}, the earliest models, e.g., Chameleon \citep{team2024chameleon} and Lumina-mGPT \citep{liu2024lumina}, relied on a purely autoregressive (AR) architecture. 
However, these models faced two key challenges: 1) Their next-token prediction paradigm resulted in extremely slow generation speeds, often requiring several minutes, which significantly affected user experience. 2) Their image generation quality was suboptimal. 
To improve quality, approaches such as MetaQueries~\citep{pan2025transfer} and BLIP3-o~\citep{chen2025blip3} added a diffusion head after the AR process to decode image tokens, enhancing quality but sacrificing the unified model concept. 
Conversely, Show-o \citep{showo} aimed to increase speed by adopting an AR+Discrete Diffusion strategy, as illustrated in Figure~\ref{fig:difference}\textcolor{magenta}{(b)}. 
However, optimal solutions were not achieved due to incomplete exploration of text-based discrete diffusion. Recent advances~\citep{llada,llada1.5} in discrete diffusion modeling for text have made unified multi-modal discrete diffusion models more feasible, as depicted in Figure~\ref{fig:difference}\textcolor{magenta}{(c)}. 
Our concurrent work, MMaDA~\citep{yang2025mmada}, has preliminarily demonstrated the potential of a comprehensive discrete diffusion architecture for unifying text-to-image generation and image understanding. 
Nevertheless, its performance remains limited, and it lacks full support for downstream generation tasks.

In this paper, we introduce \textbf{Lumina-DiMOO}, \textit{\textbf{an open-source and unified diffusion large language model}}, which possesses \textbf{\textit{versatile multi-modal capabilities}}. These capabilities encompass text-to-image generation, supporting both arbitrary and high-resolution, and a range of image-to-image tasks, including image editing, style transfer, subject-driven generation, controllable generation, multi-view generation, and dense prediction, alongside advanced image understanding, as shown in Figure~\ref{fig:abs_case}. 

Lumina-DiMOO’s unique discrete diffusion architecture \textbf{\textit{substantially enhances inference speed relative to previous unified AR or hybrid AR-Diffusion models}}. 
For example, it achieves a 32x speed improvement in text-to-image generation compared to the representative AR model—Lumina-mGPT 2.0 \citep{xin2025lumina}. 
Furthermore, during inference, we note that tokens with high maximal logit values often share similar representations with previous steps. 
To capitalize on this, we introduce a training-free Max Logit-based Cache (ML-Cache) method for Lumina-DiMOO, resulting in an additional 2x boost in sampling speed. 

Beyond its speed advantages, the discrete diffusion architecture enables Lumina-DiMOO to execute zero-shot inpainting. 
This capability can be extened to \textbf{\textit{a novel application—Interactive Retouching}}, which allows users to freely refine specific areas through precise annotations, offering flexibility that is difficult for other methods to achieve.

We evaluate Lumina-DiMOO’s capabilities across various multi-modal generation and understanding benchmarks, where it \textbf{\textit{surpasses the leading open-source unified models and sets new standards in this field}}. 
Notably, Lumina-DiMOO has achieved the first place among open-source multi-modal models on the newly released UniGenBench~\citep{wang2025pref} leaderboard\footnote{Leaderboard Link: \url{https://huggingface.co/spaces/CodeGoat24/UniGenBench_Leaderboard}.}, which is evaluated and maintained by the Tencent Hunyuan team. 
Extensive qualitative comparisons further demonstrate Lumina-DiMOO’s superior performance. These results position Lumina-DiMOO as a strong foundation model for future research and applications in general-purpose multi-modal intelligence.

\section{Related Work}
\label{sec:related_work}
\paragraph{\textbf{Diffusion Large Language Model.}}
Recent advancements in diffusion-based large language models (dLLMs) are built upon the theory of discrete diffusion, as developed by works like \citep{d3pm, mdlm, sedd, radd}. 
Among various discrete diffusion methodologies, masked diffusion has emerged as the \textit{de facto} standard due to its simplicity and effectiveness~\citep{d3pm, sedd}. 
This approach introduces a special \texttt{[mask]} state in the forward process—transforming data into \texttt{[mask]}—and recovers data in the reverse process, similar to BERT~\citep{devlin2018bert}. 
dLLMs offer distinctive advantages such as bidirectional attention, iterative refinement, flexible generation order, parallel decoding, and infilling capabilities.
These features contribute to their strong reasoning abilities~\citep{dream, dcolt, tokenorder，zhou2024lidarptq，pillarhist}, high efficiency~\citep{llada, dimple}, and enhanced inference controllability.
Recent innovations in the field include LLaDA~\citep{llada}, which scales dLLMs to 8B parameters with performance comparable to LLaMA3 8B~\citep{llama3}, and LLaDA 1.5~\citep{llada1.5}, which reduces reinforcement learning variance to better align models with human preferences. 
Multi-modal capabilities have also been explored through models like Dimple~\citep{dimple}, LLaDA-V~\citep{lladav}, and LaViDa~\citep{lavida}.
These models, although having not yet achieved peak performance, unveil a promising alternative pathway beyond autoregressive models.

\vspace{-0.15in}
\paragraph{\textbf{Unified Generation and Understanding.}} 
Unifying multi-modal generation and understanding has been a long-standing goal. One typical approach relies on using separate continuous diffusion models, where the LLM regresses image features that are subsequently decoded into images via diffusion process~\citep{emu2, pan2025transfer, chen2025blip3}.
While this method achieves decent visual generation, the reliance on external models compromises true modality unification and introduces bottlenecks that hinder seamless interaction across modalities~\citep{bagel,liquid}.
To address this, another line of research integrates diffusion within the LLM transformer, sharing parameters for both generation and understanding. 
These unified models use a single transformer to generate text autoregressively and images through either continuous~\citep{transfusion, ma2024janusflow, mogao} or discrete diffusion~\citep{showo}. 
However, there still exist heavy modality-specific designs, complicating the model and reducing the unity.
In pursuit of more simplified unification,  some works tokenize all modalities into discrete tokens, enabling uniform autoregressive processing. 
For example, Emu3~\citep{wang2024emu3}  and Lumina-mGPT~\citep{liu2024lumina} demonstrate versatility across tasks such as visual question answering and mixed-modal generation. 
However, these autoregressive models face inefficiencies from raster-scan generation orders and the inherently slow token-by-token decoding.
Multi-modal dLLMs offer a promising solution to these challenges. Their inherent flexibility in generation order and support for parallel decoding enable higher efficiency. 
Concurrent with our work, MMaDA~\citep{yang2025mmada} has preliminarily validated the feasibility of discrete diffusion on unified generation and understanding.

\begin{figure}[!t]
    \centering
    \vspace{-0.15in}
    \includegraphics[width=1.0\linewidth]{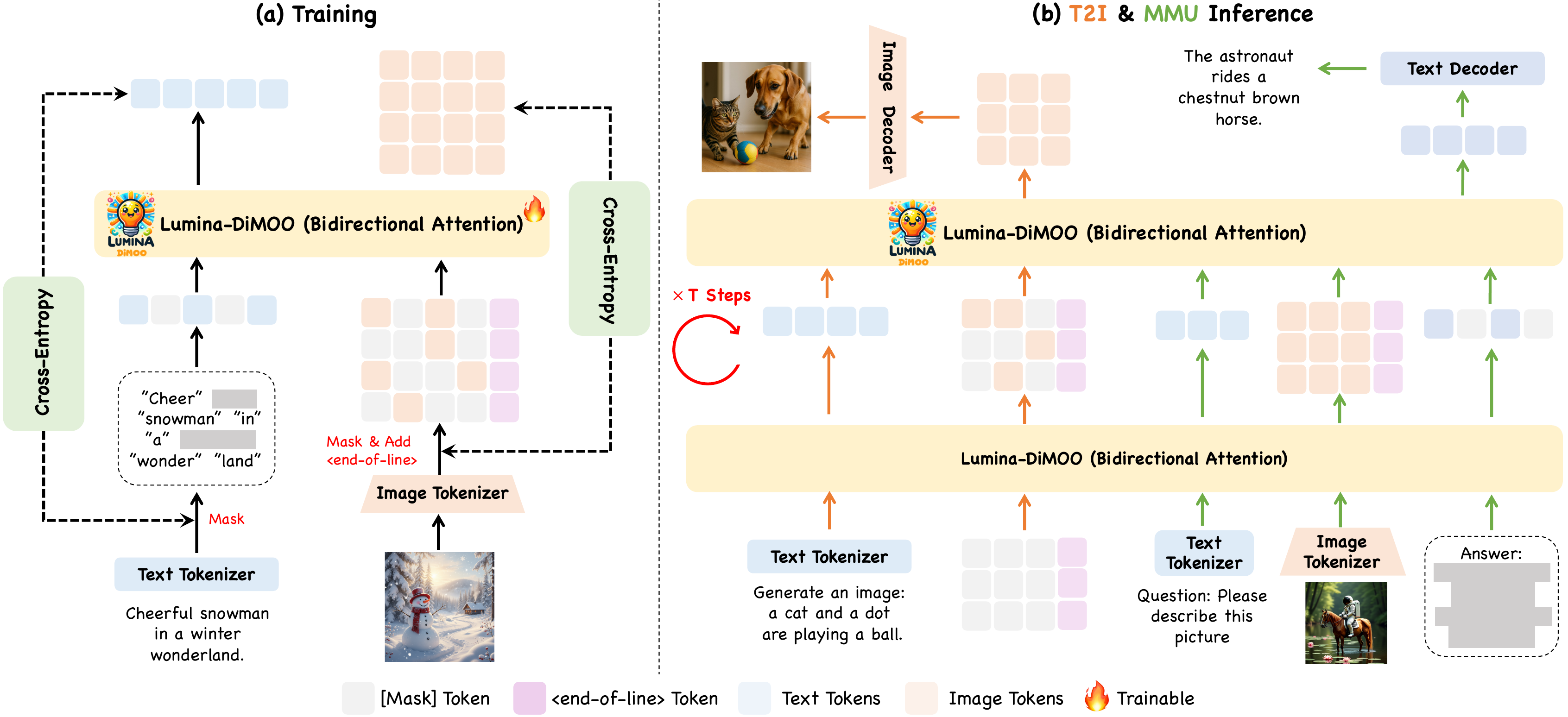}
    \caption{\textbf{An Overview of Lumina-DiMOO’s Discrete Diffusion Modeling.} (a) Training: Lumina-DiMOO is trained on text and image tokens with mask. (b) Inference: Lumina-DiMOO predicts the masked tokens, refining its output progressively.}
    \label{fig:framework}
    \vspace{-0.1in}
\end{figure}
\section{Lumina-DiMOO}
\label{sec:method}
\subsection{Foundation Image Tokenizer}
The discrete image tokenizer is a fundamental component in discrete diffusion modeling paradigms, crucial to the ultimate performance of visual generation and understanding tasks. Therefore, selecting a tokenizer capable of high-fidelity image reconstruction is essential. 
Although SBER-MoVQGAN~\citep{razzhigaev2023kandinsky}, validated in Lumina-mGPT 2.0~\citep{xin2025lumina}, is considered the state-of-the-art open-source tokenizer, its 8$\times$8 downsampling results in excessively long token sequences for high-resolution images, posing significant computational challenges. 
To balance performance with efficiency, we choose the tokenizer from aMUSEd-VQ~\citep{patil2024amused}, which uses a 16$\times$16 downsampling factor. 
We also explored other 16×16 downsampling tokenizers, such as Chameleon-VQ~\citep{team2024chameleon} and Open-MAGVIT2~\citep{luo2024open}. 
However, Chameleon-VQ produces slightly inferior reconstructions, and although Open-MAGVIT2 performs well in reconstruction, its token format doesn’t align with our modeling needs. 
A drawback of the aMUSEd-VQ tokenizer is its lack of semantic information about the image, which poses challenges for image understanding tasks. We address this by scaling the understanding data.

\subsection{Model Design}
\paragraph{Unified Discrete Diffusion Modeling.}
We adopt a unified discrete diffusion framework that not only simplifies the modeling complexity but also introduces a unified optimization objective to model both textual and visual modalities, as shown in Figure~\ref{fig:framework}\textcolor{magenta}{(a)}. 
Specifically, 
let $\mathbf{x}=(x_1,\ldots,x_L)$ denote a mixed text-image sequence drawn from the joint vocabulary (text tokens, image tokens, and special tokens, details in subsequent paragraph).
A mask set $\mathcal{M}\subseteq \{1,\ldots,L\}$ is sampled by a mask ratio $m\in(0,1]$, where the length of $\mathcal{M}$ is $\left \lfloor L\times m \right \rfloor$.
Tokens (in sequence $\mathbf{x}$) at these indices (in mask set $\mathcal{M}$) are replaced with a special \texttt{[Mask]} token. 
Therefore, the input sequence $\tilde{\mathbf{x}}$ (with \texttt{[Mask]}) construction process for the model is as follows:
\begin{equation}
\tilde{x}_i=\left\{\begin{matrix}
 [Mask] & if~~i\in\mathcal{M}, \\
 x_i & otherwise.
\end{matrix}\right.
\end{equation}

\vspace{-0.1in}
Then, the model $p_\theta$ predicts, in parallel, the original tokens at masked positions conditioned on the unmasked context and optional condition tokens $c$ (\textit{e.g.}, text prompt):
\begin{equation}
p_\theta\!\big(\mathbf{\tilde{x}}_{\mathcal{M}}\mid \mathbf{\tilde{x}}_{\overline{\mathcal{M}}},\,c\big)
=
\prod_{i\in\mathcal{M}} p_\theta\!\big(\tilde{x}_i \mid \mathbf{\tilde{x}}_{\overline{\mathcal{M}}},\,c\big).
\end{equation}

\vspace{-0.1in}
Training minimizes the masked cross-entropy over randomly sampled mask ratios $m$ applied to both text and image positions:
\begin{equation}
\mathcal{L}(\theta)
=
\mathbb{E}_{\mathbf{\mathbf{x}},\,m,\,\mathcal{M}}
\!\left[
-\sum_{i\in\mathcal{M}}
\log p_\theta\!\big(x_i \mid \mathbf{\mathbf{\tilde{x}}}_{\overline{\mathcal{M}}},\,c\big)
\right].
\end{equation}

\vspace{-0.1in}
At inference, generation starts from fully masked tokens and proceeds for $T$ refinement steps via parallel prediction-sampling-remasking, as shown in Figure~\ref{fig:framework}\textcolor{magenta}{(b)} (see Section~\ref{sec:sampling} for details).

\vspace{-0.15in}
\paragraph{Effective Initialization from Pre-Trained dLLM.} 
In the realm of autoregressive (AR) multi-modal generation and understanding, a successful paradigm involves initializing models with powerful pre-trained LLMs~\citep{liquid,showo}. 
These existing LLMs are ideal starting points for training as they already possess robust text semantic understanding and generation capabilities, which can greatly reduce training resource requirements. 
Inspired by this paradigm, Lumina-DiMOO is developed on a pre-trained dLLM, seamlessly integrating multi-modal generation and understanding within a discrete diffusion framework. 
Specifically, we utilize LLaDA-Base~\citep{llada} as our base model without any structural modifications. To demonstrate the effectiveness of this paradigm, we conduct an ablation analysis in Section~\ref{sec:initial}.

\vspace{-0.15in}
\paragraph{Multi-Modal Tokenization.} To expand the multi-modal capabilities, we make a key modification to the vocabulary. The original LLaDA model operates with 126,345 text tokens. 
We expand this by integrating 8,192 visual tokens from the pre-trained aMUSEd-VQ codebook. 
Additionally, we introduce special tokens, such as <IMAGE> and </IMAGE>, to explicitly define the boundaries of visual elements within the token sequence. 
As a result, Lumina-DiMOO’s combined vocabulary now includes 126,345 LLaDA text tokens, 8,192 aMUSEd-VQ visual tokens, and a set of special tokens. 
Detailed descriptions of these special tokens are provided in Table~\ref{tab:special_tokens}. For Lumina-DiMOO, only the newly introduced visual and special tokens require learning.

\begin{table*}[t]
\centering
\renewcommand{\arraystretch}{1.1}  
\caption{\centering \textbf{Detailed Description of the Special Tokens.}}
\vspace{-0.1in}
\begin{tabular}{>{\centering\arraybackslash}m{5.4cm} |m{10.2cm}}  
\toprule
<IMAGE> and </IMAGE> & The beginning and ending identifiers of an image.\\
\hline
<canny> and </canny> & The beginning and ending identifiers of a canny detection map image. \\
\hline
<depth> and </depth> & The beginning and ending identifiers of a depth map image. \\
\hline
<openpose> and </openpose> & The beginning and ending identifiers of a skeleton map image. \\
\hline
<hed> and </hed> & The beginning and ending identifiers of an edge detection map image. \\
\hline
<system> and </system> & The beginning and ending identifiers of a system prompt, which are usually descriptions of task prompts. \\
\hline
<user> and </user> &The beginning and ending identifiers of a user prompt, which are typically correspond to the user’s instructions or requirements.  \\
\hline
<answer> and </answer> & The beginning and ending identifiers of the model’s response. \\
\hline
<end-of-line> & The identifier for the end of a line in an image.  \\
\hline
<uncondition> & Identifiers for CFG (Classifier-Free Guidance) applied to image generation.  \\
\bottomrule
\end{tabular}
\label{tab:special_tokens}
\end{table*}

\vspace{-0.15in}
\paragraph{Arbitrary Resolution Image Representation.} For a versatile multi-modal generation and understanding model, the capability to process images of arbitrary resolutions is essential. 
However, our foundational model, LLaDA, which uses 1D RoPE designed for text, encounters challenges when applied to inherently 2D image tokens. A key issue is that images with different aspect ratios, such as 512$\times$1024 and 1024$\times$512 would be flattened into sequences of the same length, losing their distinct aspect ratios in a 1D format. 
To overcome this, we introduce a <end-of-line> token after the last image token of each row, serving as an explicit delimiter of the structure. 
This addition allows the original 2D shape of the image to be correctly parsed and reconstructed from the 1D sequence without requiring a new positional embedding design. 
This modification is crucial for enabling Lumina-DiMOO to effectively handle images with arbitrary resolution. 
In contrast, MMaDA~\citep{yang2025mmada}, sharing the same architecture with Lumina-DiMOO, only processes images at a fixed resolution of 512$\times$512.

\subsection{Inference}
\subsubsection{Sampling Strategies}
\label{sec:sampling}
\paragraph{Parallel Sampling for Image Generation.}
For image generation, we treat the entire set of image tokens to be generated (excluding special <end-of-line> tokens) as a single generation block. 
Following MaskGIT~\citep{chang2022maskgit}, we partition the image generation process into four stages. 
Generation starts from a sequence in which all image tokens are masked, i.e. $x_{t=0}$ and proceeds decoding for $T$ timesteps. At each timestep $t$, our decoding operates as follows:

\textbf{1. Predict.} Conditioned on the user prompt $c$, Lumina-DiMOO predicts, in parallel, probabilities 
$p_\theta\!\big(\tilde{x}_i^t \mid \mathbf{\tilde{x}}^t_{\overline{\mathcal{M}}},\,c\big)\in\mathbb{R}^{L'_t\times K}$
for all masked tokens, where $L'_t$ is the number of masked image tokens at timestep $t$ and $K$ is the size of the full vocabulary.

\textbf{2. Sample.} We first restrict the predicted probabilities $p_{\theta}\in\mathbb{R}^{L'_t\times K}$ from the entire vocabulary to the image-token subset $p_{\theta}\in\mathbb{R}^{L'_t\times K'}$ ($K'=8{,}192$ denotes the size of the image vocabulary). Then, for each masked image token, we sample its value with the highest probability within the image codebook and take the corresponding probability as the confidence in the timestep $t$. For image tokens that have already been decoded, we set their confidence as $-\infty$ to prevent them from participating in the re-masking step.

\textbf{3. Mask Schedule.} We use a cosine sampling schedule to determine the number of tokens to re-mask at the timestep $t$.
Specifically,
\begin{equation}
\label{eq:remask-schedule}
k_t \;=\; \Big\lceil \; \cos\!\frac{\pi t}{2T} \cdot L'_t \Big\rceil,   
\end{equation}
where $T$ is the total number of timesteps, and $k_t$ is the number of tokens to re-mask at timestep $t$.

\textbf{4. Remask.} After determining the number of re-masked tokens using the masking schedule, we select the re-masked image tokens with a top-$k$ rule according to each token's confidence obtained in the step 2 (Sample stage).

After predefined $T$ decoding timesteps, all image tokens are predicted. In addition, we employ classifier-free guidance (CFG), a commonly used strategy in the field of image generation.

\vspace{-0.1in}
\paragraph{Block-Wise Parallel Sampling for Image Understanding.}
Unlike image generation, which produces image tokens, image understanding predicts text tokens.
Following LLaDA~\citep{llada} and MMaDA~\citep{yang2025mmada}, we adopt a semi-autoregressive strategy.
Concretely, starting from a fully masked text sequence, we partition the sequence into multiple blocks.
\textbf{\textit{Within each block we perform parallel token prediction, while across blocks we decode sequentially in order.}}
While this design can enrich output details (e.g., MMaDA), it also makes the results highly sensitive to the sampling steps and the overall generation length.
In the extreme case—exemplified by MMaDA—where each step predicts only two tokens, the semi-autoregressive procedure effectively degenerates to standard next-token prediction.
Moreover, a major drawback of block-wise inference is inefficiency: the semi-autoregressive procedure always generates the full predefined length in a next-block manner, even though the model often terminates its response earlier. 
This mismatch leads to substantial redundant computation. To mitigate this, we introduce an early stopping strategy, which halts inference immediately once the current block has been completed and an </answer> token is detected, thereby reducing unnecessary steps and improving efficiency.

\subsubsection{Acceleration Sampling via Max Logit-based Cache}
\begin{wrapfigure}[20]{r}{0.5\textwidth}
	\vspace{-0.85cm}
	\flushleft 
	\includegraphics[width=0.5\textwidth]{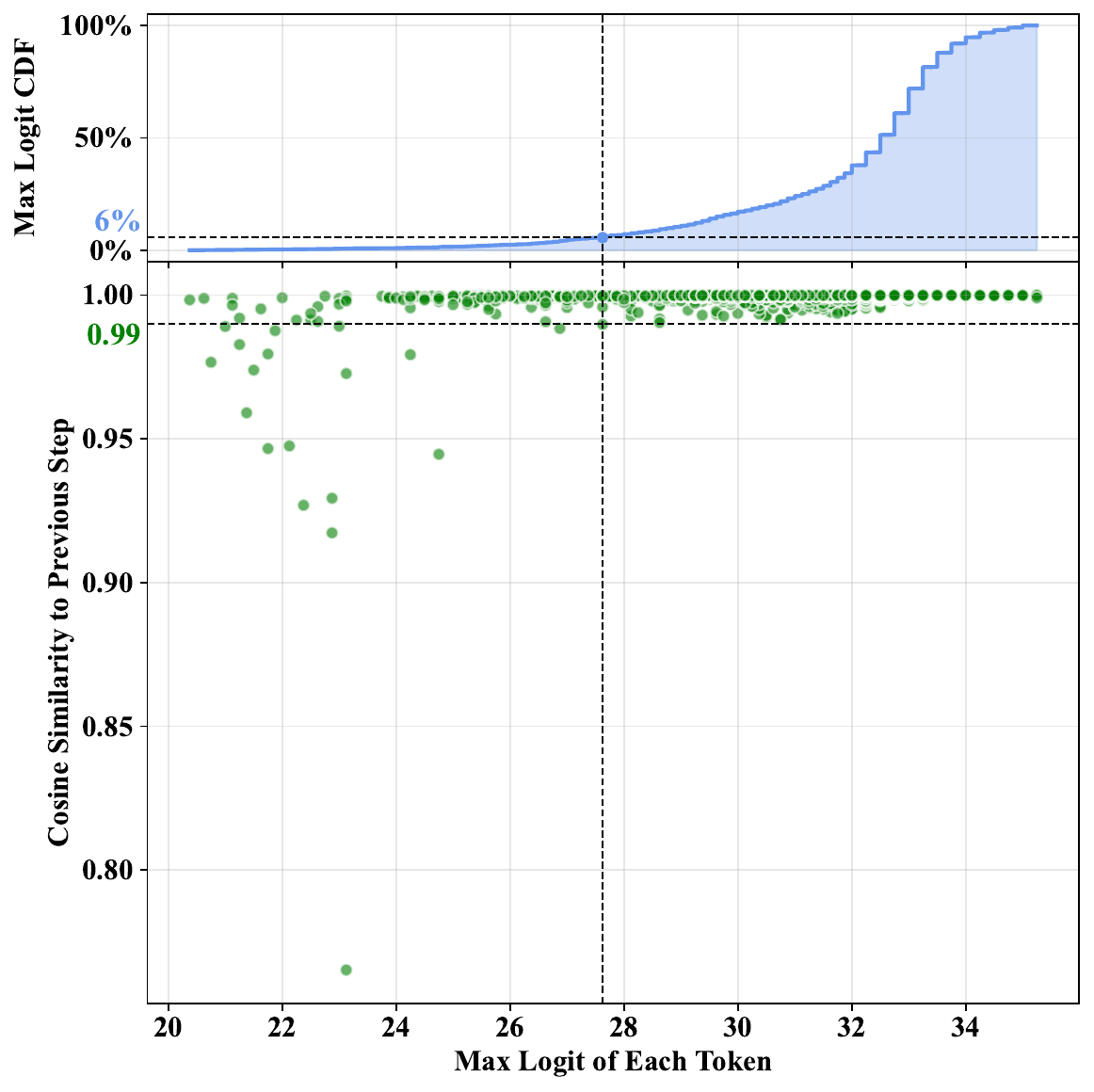}
        \vspace{-0.7cm}
	\caption{\textbf{Example of token logits statistics, illustrating that tokens with high maximal logit tend to have stable representations.}}
	\label{fig:cache_obs}
\end{wrapfigure}

Compared to AR or hybrid AR-Diffusion models, although Lumina-DiMOO could reduce generation steps by parallel sampling, each step is significantly more costly due to bidirectional attention.
Note that we cannot just excessively force fewer steps to compensate for the cost, because this will introduce compounding decoding error~\citep{jumpyoursteps, copula} and degrade the generation quality. Therefore, it is crucial to improve the computational efficiency of each step.

Autoregressive models can be losslessly accelerated through KV-Cache. Although the bidirectional attention prohibits Lumina-DiMOO from this desideratum, we can leverage the idea of caching to achieve lossy acceleration.
It turns out that the representations remain stable across steps for most tokens~\citep{dkvcache, dllmcache, fastdllm, recap}. 
Given this, we can safely skip the computation of these tokens and directly reuse the representations in the previous step. The key challenge then becomes accurately identifying these tokens.
In experiments, we find that for a token in a step, if its maximal logit is high, then the logits tend to be highly similar to those in the previous step. Figure~\ref{fig:cache_obs} shows an example, where logits of tokens with top 94\% maximal logit have over 0.99 cosine similarity.
In view of this, we use the maximal logit as the proxy to identify reusable tokens. 

Specifically, we use a hyperparameter $\texttt{cache\_ratio} \in [0, 1)$ to denote the ratio of reused tokens. In a step where we decide to reuse previous representations, we select tokens with top $\texttt{cache\_ratio} \times 100\%$ maximal logit as the reused tokens, while the remaining tokens will be computed.
We only feed the tokens to compute into the unmasking network. While computing bidirectional attention, the K and V representations of tokens to reuse are approximated by those used in the previous step. In sampling, the logits of tokens to reuse are also approximated by those in the previous step.

Another problem is which step to reuse previous representations. We use two hyperparameters \texttt{warmup\_ratio} and \texttt{refresh\_interval} to decide, similar to existing works~\citep{dkvcache, dllmcache, fastdllm, recap}.
In the beginning $\texttt{warmup\_ratio} \times 100\%$ steps, we compute all tokens to avoid error from inaccurate estimation due to the poor context.
Moreover, we compute all tokens every \texttt{refresh\_interval} steps to alleviate the error accumulation.
These mechanisms could reduce the approximation error and allow flexible tuning of efficiency-quality trade-offs.

\begin{table}[!t]
\small
\centering
\caption{\centering \textbf{Detailed Hyperparameter and Configuration of the Training Recipe Across Different Stages.}}
\label{tab:training_recipe}
\renewcommand{\arraystretch}{1.25} 
\setlength{\tabcolsep}{5pt}
\begin{adjustbox}{width=\textwidth}
\begin{tabular}{l|cccc}
\toprule
\multirow{2}{*}{\textbf{Hyperparameters}} & \textbf{Stage-I} & \textbf{Stage-II} & \textbf{Stage-III} & \textbf{Stage-IV}\\

& \textbf{(Pre-Training)} &\textbf{(Mid-Training)} &\textbf{(Supervised Fine-Tuning)} & \textbf{(Self-GRPO)} \\

\hline
Learning Rate  & $2.0\times10^{-4}$ &$2.0\times10^{-4}$ & $2.0\times10^{-5}$ &  $3.0\times10^{-6}$\\

LR Scheduler   & Constant & Constant & Constant & Constant\\

Weight Decay   & 0.1 & 0.1 & 0.1 & 0.1\\

Gradient Norm Clip  & 1.0 & 1.0 & 1.0 & 1.0\\

Optimizer       & \multicolumn{3}{c}{AdamW ($\beta_1=0.9$, $\beta_2=0.95$)} & AdamW ($\beta_1=0.9$, $\beta_2=0.99$) \\

Batch Size  &1,024 &512 &512 &48 \\

Training GPUs  &64$\times$A800 &64$\times$A800 &64$\times$A800 &8$\times$H20 \\

\hline
Gen. Resolution     & 256$\rightarrow$512 & 1024 (512 for I2I) & 1024 (512 for I2I) & 1024\\
Arbitrary Resolution &\ding{51} &\ding{51} &\ding{51} &\ding{53} \\

\hline
\multirow{2}{*}{Under. Resolution}  & \multirow{2}{*}{256$\rightarrow$512} &Dynamical \& Native &Dynamical \& Native & \multirow{2}{*}{1024}\\

& &512$\sim$1024 &512$\sim$1024 \\

Arbitrary Resolution &\ding{51} &\ding{51} &\ding{51} &\ding{53} \\
\bottomrule
\end{tabular}
\end{adjustbox}
\end{table}
\section{Training Pipeline}
The training pipeline comprises four stages, with details of each stage outlined in Table~\ref{tab:training_recipe}. 
Notably, the Self-GRPO stage is specifically designed for Lumina-DiMOO, capitalizing primarily on the discrete diffusion mechanism and the unified generation and understanding model.

\subsection{Stage-I: Multi-Modal Pre-Training for Image-Text Alignment} 
The multi-modal pre-training stage serves as a crucial bridge to transition Lumina-DiMOO from a unimodal text model to a proficient multi-modal model. 
The core goals of this stage are to cultivate visual capability and to align text and visual representations. 
To achieve this, we design a unified input format where text-image pairs are concatenated into a single sequence formatted as:
\begin{center}
\textcolor{black}{<|startoftext|>} \textcolor{red}{$\{$text tokens$\}$} \textcolor{black}{<|endoftext|>} \textcolor{black}{<|IMAGE|>} \textcolor{red}{$\{$image tokens$\}$} \textcolor{black}{<|/IMAGE|>}
\end{center}

\vspace{-0.1in}
Here, <|startoftext|> and <|endoftext|> are the begin-of-sequence and end-of-sequence tokens defined in the original text tokenizer. 
\textit{\textbf{During training, we employ a random masking strategy, where portions of text and image tokens are masked (\textcolor{red}{red area} indicates tokens that can be masked), and Lumina-DiMOO learns to predict them based on unmasked tokens.}}
To address the challenges of learning complexity associated with long visual token sequences, we introduce a progressive training schedule. 
The training begins with low-resolution (arbitrary resolution around 256$\times$256, $\sim$256 tokens), then advances to medium-resolution (arbitrary resolution around 512$\times$512, $\sim$1024 tokens).

\subsection{Stage-II: Mid-Training for Diverse Tasks}
In contrast to typical unified model training, we introduce an additional mid-training stage designed to achieve two goals: first, to integrate a diverse suite of image-to-image tasks into Lumina-DiMOO, and second, to enhance its comprehension of specialized visual data.  
The image-to-image tasks include image editing, subject-driven generation, controllable generation, style transfer (using a reference image), and multi-view generation, to name a few. 
Concurrently, the model's enhanced understanding extends to complex visual formats such as tables, charts, user interfaces, mathematical equations, and geometric structures. 
\textbf{\textit{Unlike Stage-I training, this stage focuses solely on calculating the loss for the target image in text-to-image and image-to-image tasks, or the target text in image understanding tasks.}}

\vspace{-0.15in}
\paragraph{Efficient Mid-Training.} The nature of image-to-image tasks, which typically process two or more images, results in substantially longer token sequences compared to single-image tasks such as text-to-image generation and image understanding.
This will result in low training efficiency. To address this issue, we set the image resolution for image-to-image tasks to 512. 
In contrast, for text-to-image tasks, a higher resolution of 1024 is adopted to better capture finer details. 
For image understanding tasks, we implement a dynamic resolution strategy: maintaining the original image resolution within 512 to 1024, downscaling images exceeding 1024 to 1024, and upscaling those below 512 to 512.

\subsection{Stage-III: Supervised Fine-Tuning for Instruction Following}

In the supervised fine-tuning stage, the primary objective is to enhance two key aspects of Lumina-DiMOO: its ability to align with user instructions and the overall quality of its multi-modal generation and understanding.
To achieve these objectives, we construct a large collection of high-quality <System Prompt, User Prompt, Answer> triples. 
\textit{\textbf{During training, the system prompt and user prompt remain unchanged, while the tokens in the answer are masked and the loss is computed independently.}} 
The processing of image resolution in this stage is consistent with that in Stage II.

\subsection{Stage-IV: Self-Improving via GRPO}
\begin{figure}
    \centering
    \includegraphics[width=1\linewidth]{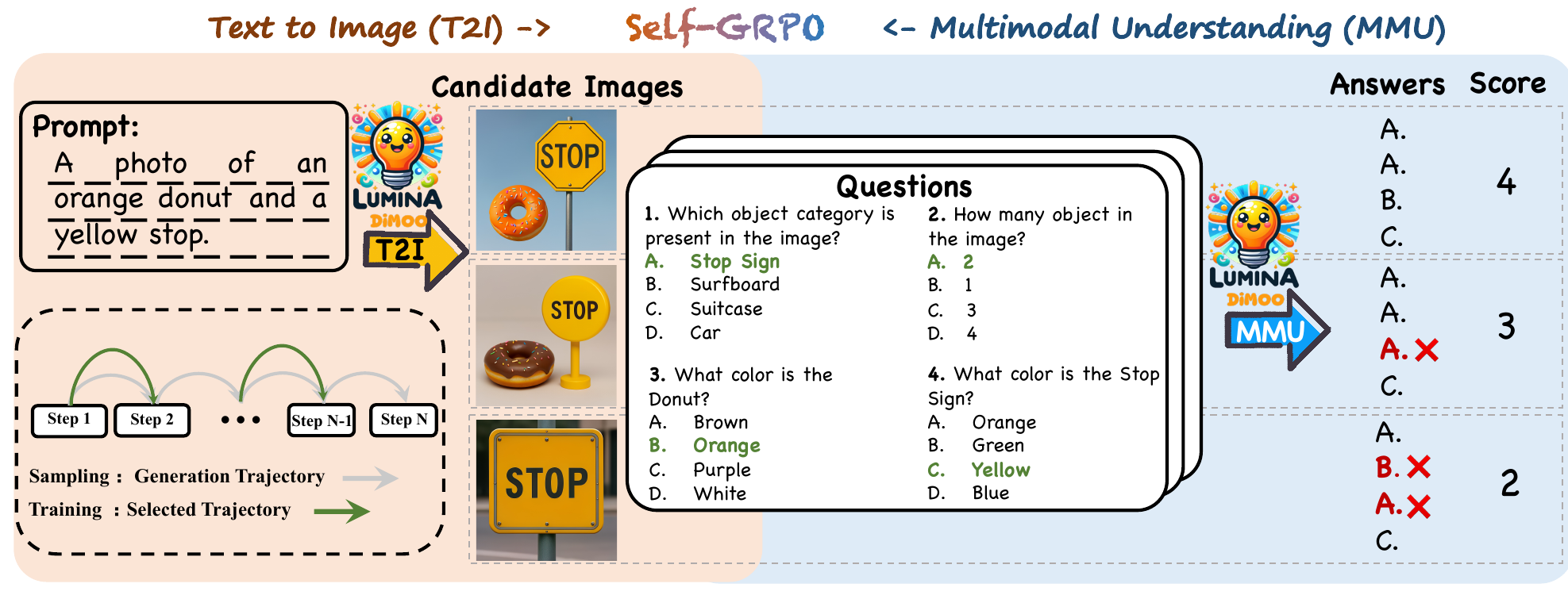}
    \vspace{-0.2in}
    \caption{\textbf{Overview of the Proposed Self-GRPO Framework.} Self-GRPO unifies text-to-image (T2I) generation and multi-modal understanding (MMU) under trajectory-consistent reinforcement learning.}
    \label{fig:self-grpo}
    \vspace{-0.15in}
\end{figure}
Finally, to fully leverage the unified nature of generation and understanding, we propose Self-GRPO, a self-improving reinforcement learning framework that jointly optimizes text-to-image (T2I) generation and multi-modal understanding (MMU). 
Unlike prior work that relies solely on answer-level MMU supervision (e.g., UniRL~\citep{mao2025unirl}) or ignores generation-inference alignment (e.g., UniGRPO~\citep{yang2025mmada}), Self-GRPO integrates structured semantic feedback and ensures trajectory-consistent training, as shown in the lower left of Figure~\ref{fig:self-grpo}.

The GRPO~\citep{guo2025deepseek} strategy requires computing the outputs of the old policy $\pi_{\theta_{old}}$ and then optimizing the current policy model $\pi_{\theta}$. 
Lumina-DiMOO supports high-resolution (1024$\times$1024) image generation with image sequences of length $L=4096$, while unified T2I and MMU tasks further increase the sequence length. 
Computing both $\pi_{\theta_{old}}$ and $\pi_{\theta}$ while storing the corresponding activations and gradients imposes a substantial memory burden. 
Following UniRL~\citep{mao2025unirl}, we eliminate the old policy $\pi_{\theta_{old}}$ to reduce memory consumption. 
At each training step, given a prompt $p$, we sample $G$ candidate images as token sequences
$\{x^{(g)}\}_{g=1}^{G}$ (each of length $L$) from the current policy $p_\theta$.
Given a set of questions $\{q_n\}_{n=1}^N$, the model then answers each $q_n$ conditioned on the generated image $x^{(g)}$ to obtain per-sample T2I and MMU losses, $\ell^{(g)}_{\text{T2I}}$ and $\ell^{(g)}_{\text{MMU}}$. Combining these, we optimize the reward-weighted objective:

\vspace{-0.25in}
\begin{equation}
L(\theta) = - \sum_{g=1}^{G} w^{(g)} \left( \ell^{(g)}_{\text{T2I}} + \ell^{(g)}_{\text{MMU}} \right)
\;+\; \beta\, \mathrm{KL}\!\left( p_\theta \,\Vert\, p_\theta^{\text{ref}} \right).
\label{eq:grpo}
\vspace{-0.15in}
\end{equation}

Rewards $r^{(g)}$ are defined as the number of correct answers across $\{q_n\}$ and are normalized with a softmax temperature $\alpha$:
\vspace{-0.2in}
\begin{equation}
w^{(g)} = \frac{\exp\!\big(\alpha\,(r^{(g)}-\bar r)\big)}
{\sum_{j=1}^{G}\exp\!\big(\alpha\,(r^{(j)}-\bar r)\big)},
\quad
\bar r = \frac{1}{G}\sum_{j=1}^{G} r^{(j)}.
\label{eq:weights}
\vspace{-0.1in}
\end{equation}

Reinforcement learning involves two distinct processes: output sampling and reward updating, where the latter follows the sampling trajectory to assign rewards and compute gradients accordingly. 
Unlike autoregressive MLLMs, Lumina-DiMOO performs multi-step forward passes and re-masking during image generation. Consequently, it is necessary to design sophisticated strategies to preserve trajectory consistency. Since the primary content can be generated in early timesteps during T2I generation~\citep{chang2022maskgit}, we propose a step trajectory following strategy to improve memory efficiency. 
Specifically, Self-GRPO retains the complete sampling trajectory but computes gradients only from the model outputs at selected timesteps $\mathcal{T}_\text{sel}$. The T2I log-likelihood is defined as:

\vspace{-0.2in}
\begin{equation}
\ell^{(g)}_{\text{T2I}} = 
\frac{1}{|\mathcal{T}_\text{sel}|} 
\sum_{t \in \mathcal{T}_\text{sel}} 
\log p_\theta \left( x^{(g)}_t \mid x^{(g)}_{<t}, p \right).
\label{eq:t2i}
\vspace{-0.1in}
\end{equation}

To evaluate the MMU capability of the model, we compute the average log-likelihood of $N$ predicted answers, where each answer $y_n^{(g)}$ is generated conditioned on the corresponding question $q_n$ and image tokens $x^{(g)}$:

\vspace{-0.25in}
\begin{equation}
\ell^{(g)}_{\text{MMU}} = \frac{1}{N} \sum_{n=1}^{N}
\log p_\theta \left( y_n^{(g)} \mid x^{(g)}, q_n \right).
\label{eq:mmu}
\vspace{-0.1in}
\end{equation}

Self-GRPO therefore unifies vision and language under a trajectory-consistent framework. 
By combining KL-regularized policy updates, memory-efficient training, and multi-modal reward supervision, it closes the training loop between generation and understanding.

\section{Data Construction}
\paragraph{Stage-I: Pre-Training Data.} 
\label{sec:pretain-data}
We collect approximately 80 million high-quality text-image pairs, sourced from diverse and reliable datasets, including 30 million pairs from re-captioned public collections (i.e., LAION-400M~\citep{schuhmann2021laion} for image understanding pre-training and CC12M~\citep{changpinyo2021conceptual}) and 50 million from Lumina-Image 2.0~\citep{LuminaImage2024}, Lumina-mGPT 2.0~\citep{xin2025lumina} for image generation pre-training.

\vspace{-0.15in}
\paragraph{Stage-II: Mid-Training Data.} In this stage, we incorporate an additional 3 million images from several challenging domains: MMTable~\citep{zheng2024multimodal} and TinyChart~\citep{zhang2024tinychart} for table and chart comprehension, AutoGeo~\citep{huang2025autogeo} and MAVIS~\citep{zhang2024mavis} for understanding math equations and geometric structures, and MultiUI~\citep{liuharnessing} for user interface parsing, which are all captioned using Qwen2.5-VL~\citep{Qwen2.5-VL}. 
For image-to-image tasks, data is sourced from several datasets, including UltraEdit~\citep{zhao2024ultraedit}, OmniEdit~\citep{wei2024omniedit}, OminiControl~\citep{tan2024ominicontrol}, and Lumina-mGPT 2.0~\citep{xin2025lumina}. 

\vspace{-0.15in}
\paragraph{Stage-III: Supervised Fine-Tuning Data.} 
For image understanding, we construct a high-quality dataset of 15 million samples, combining 2 million from MAmmoTH-VL dataset~\citep{guo2024mammoth} and 13 million from InternVL-2.5-SFT dataset~\citep{chen2024expanding}. For visual generation, we utilize a total of 15 million samples, aggregated from Lumina-Image 2.0~\citep{LuminaImage2024} (selecting only the highest quality data), Blip3o-60k~\citep{chen2025blip3}, ShareGPT-4o-Image~\citep{chen2025sharegpt}, and additional in-house synthetic data. 
For image-to-image tasks, we incorporate data for \textit{Subject-Driven Generation}, \textit{Controllable Generation}, \textit{Dense Prediction}, and \textit{Style Transfer}, each comprising 200K examples from VisualCloze~\citep{li2025visualcloze}. 
Additionally, there are 500K instruction-guided \textit{Image Editing} samples from UniWorld~\citep{lin2025uniworld} and 200K examples for \textit{Low-Level Vision} tasks from Lumina-OmniLV~\citep{pu2025lumina}, focusing on enhancements like super-resolution, dehazing, and denoising, etc. 
However, we find that Lumina-DiMOO perform poorly on low-level tasks. For \textit{Multi-View Generation}, we use data consistent with Lumina-mGPT~\citep{liu2024lumina}.

\vspace{-0.15in}
\paragraph{Stage-IV: Self-GRPO Data.} In this stage, only text prompt data is required. We utilize prompt from the subset of GenRef~\citep{zhuo2025reflection}, which resemble GenEval’s prompt templates. 
From each prompt, we extract (entity, relation, value) triples using DSG~\citep{2024dsg} method. These triples are then used to craft single-choice questions for semantic alignment supervision in Self-GRPO. 
To generate distractor options, we maintain global candidate pools for entities, relations, quantities, and colors. 
For each question, distractors are selected to be semantically close to the correct answer, ensuring that the resulting QA tasks are both challenging and informative.

\section{Evaluation}
\label{sec:eval}
\subsection{Performance of Text-to-Image Generation}
For evaluating text-to-image generation capabilities, we conduct evaluations using five publicly available benchmarks—GenEval~\citep{ghosh2024geneval}, DPG~\citep{hu2024ella}, UniGenBench~\citep{wang2025pref}, OneIG-EN~\citep{chang2025oneig}, and TIIF~\citep{wei2025tiif}. 
These benchmarks offer a comprehensive framework to measure the model’s proficiency in generating high-quality, semantically consistent images from textual prompts. 
Additionally, we perform qualitative comparisons with state-of-the-art models to complement these automatic evaluation metrics, ensuring a robust analysis of performance.

\subsubsection{Quantitative Results}
\begin{table}[t]
    \centering
    \setlength{\tabcolsep}{2pt}
    \renewcommand{\arraystretch}{1.25}
    \scriptsize
    \caption{\textbf{Evaluation of Text-to-Image Generation on GenEval~\citep{ghosh2024geneval} Benchmark}. ``Und.'' and ``Gen.'' denote ``understanding'' and ``generation'', respectively. We highlight the \textcolor{red}{best} and the \textcolor{blue}{second} results.
    }
    \vspace{-0.2cm}
    \scalebox{0.858}{
    \begin{tabular}{lcc|cccccc|c}
        \toprule
        \textbf{Method} & \textbf{Architecture} & \textbf{\# Params.} & \textbf{Single Obj.} & \textbf{Two Obj.} & \textbf{Counting} & \textbf{Colors} & \textbf{Position} & \textbf{Attribute} & \textbf{Overall $\uparrow$} \\
        \midrule
        \rowcolor{gray!10} \multicolumn{10}{c}{\textbf{Gen. Only}} \\ \hline 
        LlamaGen~\citep{sun2024autoregressive} &AR &0.8B & $0.71$ & $0.34$ & $0.21$ & $0.58$ & $0.07$ & $0.04$ & $0.32$ \\
        
        PixArt-$\alpha$~\citep{chen2023pixart} &Diffusion &0.6B &$0.98$ & $0.50$ & $0.44$ & $0.80$ & $0.08$ & $0.07$ & $0.48$ \\
        
        SDv$2.1$~\citep{rombach2022high} &Diffusion &0.9B &$0.98$ & $0.51$ & $0.44$ & $0.85$ & $0.07$ & $0.17$ & $0.50$ \\
        
        Emu$3$-Gen ~\citep{wang2024emu3} &AR & 8B  &$0.98$ & $0.71$ & $0.34$ & $0.81$ & $0.17$ & $0.21$ & $0.54$ \\
        
        SDXL~\citep{2023SDXL} &Diffusion &2.6B  &$0.98$ & $0.74$ & $0.39$ & $0.85$ & $0.15$ & $0.23$ & $0.55$ \\
        
        DALL-E $3$~\citep{dalle3} &- &-  & $0.96$ & $0.87$ & $0.47$ & $0.83$ & $0.43$ & $0.45$ & $0.67$ \\
        
        SD3-Medium~\citep{esser2024scaling}&Diffusion &2B & \textcolor{blue}{$0.99$} & \textcolor{blue}{$0.94$} & $0.72$ & $0.89$ & $0.33$ & $0.60$ & $0.74$ \\
        
        FLUX.1 [Dev]~\citep{flux2024} &Diffusion &12B &$0.98$ &$0.81$ &$0.74$ &$0.79$ &$0.22$ &$0.45$ &$0.66$ \\
        
        OmniGen~\citep{xiao2024omnigen} &Diffusion &3.8B &$0.98$ &$0.84$ &$0.66$ &$0.74$ &$0.40$ &$0.43$ &$0.68$\\

        SANA-1.5~\citep{xie2025sana} &Diffusion &4.8B &\textcolor{blue}{$0.99$} &$0.85$ &$0.77$ &$0.87$ &$0.34$ &$0.54$ &$0.72$\\    
        
        Lumina-mGPT 2.0~\citep{xin2025lumina} &AR &7B &\textcolor{blue}{$0.99$} &$0.87$ &$0.44$ &$0.85$ &$0.44$ &$0.54$ &$0.69$\\
        
        \midrule
        \rowcolor{gray!10} \multicolumn{10}{c}{\textbf{Und. and Gen.}} \\ \hline 
        SEED-X~\citep{ge2024seed} &AR &17B  & $0.97$ & $0.58$ & $0.26$ & $0.80$ & $0.19$ & $0.14$ & $0.49$ \\
        
        Show-o~\citep{showo} &AR+Discrete Diff. &1.3B &  $0.95$ & $0.52$ & $0.49$ & $0.82$ & $0.11$ & $0.28$ & $0.53$ \\
        
        Janus~\citep{wu2024janus} &AR &1.3B &$0.97$ &$0.68$ &$0.30$ &$0.84$ &$0.46$ &$0.42$ &$0.61$ \\
        
        D-DiT~\citep{li2024dual} &Discrete Diff.+Diff. &2B &  $0.97$ & $0.80$ & $0.54$ & $0.76$ & $0.32$ & $0.50$ & $0.65$ \\
        
        Transfusion~\citep{transfusion} &AR+Diff. &7B & - & - & - & - & - & - & $0.63$ \\
        
        TokenFlow-XL~\citep{liu2024world}&AR &14B &  $0.95$ & $0.60$ & $0.41$ & $0.81$ & $0.16$ & $0.24$ & $0.55$ \\
        
        Chameleon~\citep{team2024chameleon} &AR &7B &  - & - & - & - & - & - & $0.39$ \\
        
        Janus-Pro~\citep{chen2025januspro} & AR &7B &\textcolor{blue}{$0.99$} & $0.89$ & $0.59$ &\textcolor{blue}{$0.90$} & \textcolor{blue}{$0.79$} &$0.66$ & $0.80$ \\

        GPT-4o~\citep{gpt4o} &- &- &\textcolor{blue}{$0.99$} &$0.92$ &\textcolor{red}{$0.85$} &\textcolor{red}{$0.92$} &$0.75$ &$0.61$ &\textcolor{blue}{$0.84$} \\

        BLIP3-o~\citep{chen2025blip3} &AR+Diff. &8B &- &- &- &- &- &- &$0.80$ \\
        
        BAGEL~\citep{bagel} &AR+Diff. &14B &\textcolor{blue}{$0.99$} &\textcolor{blue}{$0.94$} &\textcolor{blue}{$0.81$} &$0.88$ &$0.64$ &$0.63$ &$0.82$ \\

        Uniworld-V1~\citep{lin2025uniworld} &AR+Diff. &20B &\textcolor{blue}{$0.99$} &$0.93$ &$0.79$ &$0.89$ &$0.49$ &\textcolor{blue}{$0.70$} &$0.80$ \\

        OmniGen2~\citep{wu2025omnigen2} &AR+Diff. &7B &\textcolor{red}{$1.0$} &\textcolor{red}{$0.95$} &$0.64$ &$0.88$ &$0.55$ &\textcolor{red}{$0.76$} &$0.80$ \\
        
        MMaDA~\citep{yang2025mmada} &Discrete Diff. &8B & \textcolor{blue}{$0.99$} & $0.76$ &$0.61$ &$0.84$ &$0.20$ &$0.37$ &$0.63$\\
        
        \rowcolor{green!10}\textbf{Lumina-DiMOO (Ours)} &Discrete Diff. &8B 	&\textcolor{red}{$1.0$} &\textcolor{blue}{$0.94$} 	&\textcolor{red}{$0.85$} &$0.89$ &\textcolor{red}{$0.85$} &\textcolor{red}{$0.76$} &\textcolor{red}{$0.88$}\\

        \rowcolor{orange!10}\textbf{Lumina-DiMOO w/ Self-GRPO} &Discrete Diff. &8B 	&$1.0$ &$0.96$(\textcolor{violet}{+2\%}) 	&$0.87$(\textcolor{violet}{+2\%}) &$0.95$(\textcolor{violet}{+6\%}) &$0.85$ &$0.82$(\textcolor{violet}{+6\%}) &$0.91$(\textcolor{violet}{+3\%})\\
        \bottomrule
    \end{tabular}
}
    \label{tab:geneval}
    \vspace{-0.3cm}
\end{table}

\begin{table}[!t]
    \centering
    \setlength{\tabcolsep}{4.5pt}
    \renewcommand{\arraystretch}{1.25}
    \scriptsize
    \caption{\textbf{Evaluation of Text-to-Image Generation on DPG~\citep{hu2024ella} Benchmark}. ``Und.'' and ``Gen.'' denote ``understanding'' and ``generation'', respectively. We highlight the \textcolor{red}{best} and the \textcolor{blue}{second} results. ``$\dagger$'' means the MMaDA results are evaluated by ourselves.
    }
    \vspace{-0.2cm}
    \scalebox{1.025}{
    \begin{tabular}{lcc|ccccc|c}
        \toprule
        \textbf{Method} & \textbf{Architecture} & \textbf{\# Params.} & \textbf{Global} & \textbf{Entity} & \textbf{Attribute} & \textbf{Relation} & \textbf{Other} & \textbf{Overall $\uparrow$} \\
        \midrule
        \rowcolor{gray!10} \multicolumn{9}{c}{\textbf{Gen. Only}} \\ \hline
        
        PixArt-$\alpha$ \citep{chen2023pixart} &Diffusion &0.6B & $74.97$ & $79.32$ & $78.60$ & $82.57$ & $76.96$ & $71.11$ \\
        
        Lumina-Next \citep{zhuo2024lumina} &Diffusion &2B & $82.82$ & $88.65$& $86.44$ & $80.53$ & $81.82$ & $74.63$ \\
        
        SDXL \citep{2023SDXL} &Diffusion &2.6B & $83.27$ & $82.43$ & $80.91$ & $86.76$ & $80.41$ & $74.65$ \\
        
        Emu3-Gen \citep{wang2024emu3} &AR & 8B & $85.21$ & $86.68$ & $86.84$ & $90.22$ & $83.15$ & $80.60$ \\
        
        DALL-E 3 \citep{dalle3} &- &- & \textcolor{red}{$90.97$} & $89.61$ & $88.39$ & $90.58$ &$89.83$ & $83.50$ \\
        
        SD3-Medium \citep{esser2024scaling} &Diffusion &2B & $87.90$ & \textcolor{blue}{$91.01$} & $88.83$ & $80.70$ & $88.68$ & $84.08$ \\	 	
        
        FLUX.1 [Dev]~\citep{flux2024} &Diffusion &12B &$74.35$ &$90.00$ &$88.96$ &$90.87$ &$88.33$ &$83.84$\\
        
        OmniGen~\citep{xiao2024omnigen} &Diffusion &3.8B &$87.90$ &$88.97$ &$88.47$ &$87.95$ &$83.56$ &$81.16$\\

        SANA-1.5~\citep{xie2025sana} &Diffusion &4.8B &- &- &- &- &- &$85.00$\\
        
        Lumina-mGPT 2.0~\citep{xin2025lumina} &AR &7B &- &$88.94$ &$88.08$ &$91.70$ &- &$84.30$\\
        
        \midrule
        \rowcolor{gray!10} \multicolumn{9}{c}{\textbf{Und. and Gen.}} \\ \hline
        
        Show-o~\citep{showo} &AR+Diff. &1.3B &- &- &- &- &-  &$67.48$ \\
        
        TokenFlow-XL~\citep{liu2024world} & AR &14B &$78.72$ &$79.22$ &$81.29$ &$85.22$ &$71.20$ &$73.38$ \\
        
        Janus~\citep{wu2024janus}&AR &1.3B & $82.33$ & $87.38$ & $87.70$ & $85.46$ & $86.41$ & $79.68$ \\
        
        Janus-Pro~\citep{chen2025januspro}& AR & 7B & $86.90$ & $88.90$ &$89.40$ & $89.32$ & $89.48$ & $84.19$ \\

        GPT-4o~\citep{gpt4o} &- &- & $88.89$ & $88.94$ & $89.84$ & \textcolor{blue}{$92.63$} & \textcolor{red}{$90.96$} &\textcolor{blue}{$85.15$} \\

        BLIP3-o~\citep{chen2025blip3}& AR+Diff. &8B &- &- &- &- &- &$81.60$ \\
        
        BAGEL~\citep{bagel} &AR+Diff. &14B &\textcolor{blue}{$88.94$} &$90.37$ 	&\textcolor{red}{$91.29$} 	&$90.82$ &$88.67$  &$85.07$ \\

        Uniworld-V1~\citep{lin2025uniworld} &AR+Diff. &20B &$83.64$ &$88.39$ &$88.44$ &$89.27$ &$87.22$ &$81.38$ \\

        OmniGen2~\citep{wu2025omnigen2} &AR+Diff. &7B &$88.81$ &$88.83$ &\textcolor{blue}{$90.18$} &$89.37$ &\textcolor{blue}{$90.27$} &$83.57$ \\
        
        MMaDA$^\dagger$~\citep{yang2025mmada} &Discrete Diff. &8B &$77.81$ &$78.48$ &$81.74$ &$84.79$ &$63.20$ &$69.97$ \\
        
        \rowcolor{green!10} \textbf{Lumina-DiMOO (Ours)} &Discrete Diff. &8B &$81.46$ 	&\textcolor{red}{$92.08$} 	&$88.98$ 	&\textcolor{red}{$94.31$} 	&$82.00$ 	&\textcolor{red}{$86.04$}\\ 
        
        \bottomrule
    \end{tabular}
}
    \label{tab:dpg}
\end{table}

\begin{table*}[!t]
\centering
\tiny
\setlength{\tabcolsep}{2pt}
\renewcommand{\arraystretch}{1.25}
\caption{\textbf{Evaluation of Text-to-Image Generation on on UniGenBench~\citep{wang2025pref}}. This leadborder is evaluated and maintained by the Tencent Hunyuan team. ``Und.'' and ``Gen.'' denote ``understanding'' and ``generation'', respectively. We highlight the \textcolor{red}{best} and the \textcolor{blue}{second} results.} 
\vspace{-0.1in}
\scalebox{1.24}{
\begin{tabular}{lcccccccccc|c}
\toprule
\textbf{Model} & \textbf{Style} & \textbf{World Know.} & \textbf{Attribute} & \textbf{Action} & \textbf{Relation.} & \textbf{Logic.} & \textbf{Grammar} & \textbf{Compound} & \textbf{Layout} & \textbf{Text} & \textbf{Overall} \\
\midrule
\rowcolor{gray!10}
\multicolumn{12}{c}{\textbf{Gen. Only}} \\
\midrule
SDXL~\citep{2023SDXL}      & 87.40 & 72.63 & 44.34 & 34.22 & 44.92 &  9.55 & 47.33 & 26.68 & 29.85 &  0.57 & 39.75     \\
Playground 2.5~\citep{li2024playground} & 89.50 & 76.11 & 52.78 & 42.68 & 51.52 & 16.59 & 53.21 & 35.44 & 37.13 &  1.15 & 45.61 \\
Emu3~\cite{wang2024emu3}   & 86.80 & 77.06 & 51.39 & 40.11 & 49.75 & 19.32 & 52.94 & 36.86 & 44.78 &  1.15 &46.02 \\
DALL-E-3~\citep{dalle3}    & \textcolor{red}{95.06} & \textcolor{red}{93.51} & \textcolor{blue}{75.97} & \textcolor{blue}{69.83} & \textcolor{blue}{78.06} & \textcolor{red}{48.18} & 68.07 & \textcolor{blue}{70.60} & 66.67 & 25.86 & \textcolor{blue}{69.18}  \\
SD-3.5-Large~\citep{esser2024scaling}   & 88.60 & 88.92 & 68.59 & 62.17 & 69.80 & 32.27 & 58.96 & 58.76 & 69.03 & \textcolor{red}{32.76} &62.99 \\
FLUX.1-dev~\citep{flux2024} & 83.90 & 88.92 & 67.84 & 62.17 & 67.26 & 30.91 & 60.96 & 47.04 & 71.83 & \textcolor{blue}{32.18} &61.30   \\
\hline
\rowcolor{gray!10}
\multicolumn{12}{c}{\textbf{Und. and Gen.}} \\
\midrule
Janus-flow~\citep{ma2024janusflow}   & 86.20 & 62.50 & 47.97 & 43.35 & 50.00 & 21.14 & 60.29 & 45.10 & 46.46 &  0.86 &46.39  \\
BLIP3-o~\citep{chen2025blip3}    & \textcolor{blue}{92.80} & 80.22 & 63.89 & 63.97 & 66.50 & 39.55 & \textcolor{blue}{68.45} & 53.74 & 68.47 &  1.15 &  59.87   \\
Janus-Pro~\citep{chen2025januspro}   & 90.80 & 86.71 & 67.74 & 64.26 & 68.40 & 37.05 & 64.44 & 62.11 & 72.01 &  2.59 &61.61   \\
BAGEL~\citep{bagel}      & 90.20 & 85.60 & 67.74 & 61.98 & 70.69 & 30.23 & 66.44 & 58.12 & \textcolor{blue}{76.49} &  7.76 &  61.53  \\
UniWorld-V1~\cite{lin2025uniworld} &91.10 &82.91 &70.62 &67.21 &67.13 &38.41 &63.77 &54.51 &69.03 &26.44 &63.11\\
OmniGen2~\citep{wu2025omnigen2} &91.90 &86.39 &72.12 &62.83 &68.27 &32.50 &59.89 &56.31 &71.64 &29.02 &63.09\\
MMaDA~\citep{yang2025mmada} &82.40 &56.65 &48.39 &37.83 &50.25 &17.95 &55.75 &32.35 &30.22 &1.15 &41.35\\
\rowcolor{green!10} \textbf{Lumina-DiMOO (Ours)} &89.70 &\textcolor{blue}{90.03} &\textcolor{red}{81.62} &\textcolor{red}{71.12} &\textcolor{red}{78.43} &\textcolor{blue}{45.45} &\textcolor{red}{70.45} &\textcolor{red}{73.32} &\textcolor{red}{82.84} &25.57 &\textcolor{red}{71.12}\\

\bottomrule
\end{tabular} 
}
\label{tab:unibench}
\end{table*}
\begin{table}[!th]
    \centering
    \caption{\textbf{Evaluation of Text-to-Image Generation on OneIG-EN~\citep{chang2025oneig} Benchmark.} The overall score is the average of the five dimensions. ``Und.'' and ``Gen.'' denote ``understanding'' and ``generation'', respectively. We highlight the \textcolor{red}{best} and the \textcolor{blue}{second} results.}
    \vspace{-0.15cm}
    \setlength{\tabcolsep}{5pt}
    \renewcommand{\arraystretch}{1.25}
    \resizebox{0.99\linewidth}{!}{
    \begin{tabular}{lcc|ccccc|c}
        \toprule
        \textbf{Method} &\textbf{Architecture} &\textbf{\# Params.} &\textbf{Alignment}& \textbf{Text} & \textbf{Reasoning} & \textbf{Style}& \textbf{Diversity} & \textbf{Overall} $\uparrow$ \\
        \midrule
        \rowcolor{gray!10} \multicolumn{9}{c}{\textbf{Gen. Only}} \\ \hline

        SD 1.5~\citep{rombach2022high} &Diffusion &0.9B  &0.565 &0.010 &0.207 &0.383 &\textcolor{red}{0.429} &0.319 \\

        SDXL~\citep{2023SDXL} &Diffusion &2.6B & 0.688 & 0.029 & 0.237 & 0.332 & 0.296 &0.316\\
        FLUX.1 [Dev]~\citep{flux2024} &Diffusion &12B & \textcolor{blue}{0.786} & \textcolor{blue}{0.523} & \textcolor{blue}{0.253} & 0.368 & 0.238 &\textcolor{blue}{0.434}\\
        SANA-1.5~\citep{xie2025sana} &Diffusion &4.8B & 0.765 & 0.069 & 0.217 & \textcolor{red}{0.401} & 0.216 &0.334\\
        
        \midrule
        \rowcolor{gray!10} \multicolumn{9}{c}{\textbf{Und. and Gen.}} \\ \hline
        
        Janus-Pro~\citep{chen2025januspro}&AR &7B & 0.553  & 0.001  &   0.139     & 0.276 & \textcolor{blue}{0.365} & 0.267\\

        BLIP3-o~\citep{chen2025blip3} &AR+Diff. &8B &0.711 &0.013 &0.223 &0.361 &0.229 &0.307 \\
        BAGEL~\citep{bagel} &AR &14B & 0.769  & 0.244  &   0.173    & 0.367 & 0.251& 0.361\\
        \rowcolor{green!10} \textbf{Lumina-DiMOO (Ours)} &Discrete Diff. &8B  &\textcolor{red}{0.816} &\textcolor{red}{0.551} &\textcolor{red}{0.276} &\textcolor{blue}{0.400} &0.232 &\textcolor{red}{0.455}\\

        \bottomrule
    \end{tabular}
    }
    \label{tab:oneig_en}
\end{table}

\begin{table}[!t]\centering
\caption{\textbf{Evaluation of Text-to-Image Generation on TIIF~\citep{wei2025tiif} Benchmark.} ``Und.'' and ``Gen.'' denote ``understanding'' and ``generation'', respectively. We highlight the \textcolor{red}{best} and the \textcolor{blue}{second} results.}
\vspace{-0.15cm}
\renewcommand{\arraystretch}{1.7} 
\setlength{\tabcolsep}{3pt}
\centering
\begin{adjustbox}{width=\textwidth}
\begin{tabular}{l|cc|cccccccc|cccccccccccc|cc}
\toprule
\multirow{3}{*}{\textbf{Method}}
  & \multicolumn{2}{c|}{\multirow{2}{*}{\textbf{Overall} $\uparrow$}}
  & \multicolumn{8}{c|}{\textbf{Basic Following}}
  & \multicolumn{12}{c|}{\textbf{Advanced Following}}
  & \multicolumn{2}{c}{\textbf{Designer}} \\

\cmidrule(lr){4-11} \cmidrule(lr){12-23} \cmidrule(lr){24-25}

& & &
  \multicolumn{2}{c}{Avg}                    
  & \multicolumn{2}{c}{Attribute}
  & \multicolumn{2}{c}{Relation}
  & \multicolumn{2}{c|}{Reasoning}
  & \multicolumn{2}{c}{Avg}                  
  & \multicolumn{2}{c}{\makecell{Attribute\\+Relation}}
  & \multicolumn{2}{c}{\makecell{Attribute\\+Reasoning}}
  & \multicolumn{2}{c}{\makecell{Relation\\+Reasoning}}
  & \multicolumn{2}{c}{Style}
  & \multicolumn{2}{c|}{Text}
  & \multicolumn{2}{c}{\makecell{Real\\World}} \\

& short & long &          
  short & long &          
  short & long &          
  short & long &          
  short & long &          
  short & long &          
  short & long &          
  short & long &          
  short & long &          
  short & long &          
  short & long &          
  short & long            
\\

\midrule
\rowcolor{gray!10} \multicolumn{25}{c}{\textbf{Gen. Only}} \\ \hline

SD 3~\citep{esser2024scaling}            &67.46 &66.09 &78.32 &77.75 &\textcolor{blue}{83.33} &79.83 &82.07 &78.82 &71.07 &74.07 &61.46 &59.56 &61.07 &64.07 &68.84 &70.34 &50.96 &57.84 &66.67 &\textcolor{blue}{76.67} &\textcolor{red}{59.83} &20.83 &63.23 &67.34 \\
PixArt-$\Sigma$~\citep{chen2023pixart} &62.00 &58.12 &70.66 &75.25 &69.33 &78.83 &75.07 &77.32 &67.57 &69.57 &57.65 &49.50 &65.20 &56.57 &66.96 &61.72 &66.59 &54.59 &\textcolor{red}{83.33} &70.00 &1.83 &1.83 &62.11 &52.41 \\
Lumina-Next~\citep{zhuo2024lumina} &50.93 &52.46 &64.58 &66.08 &56.83 &59.33 &67.57 &71.82 &69.32 &67.07 &44.75 &45.63 &51.44 &43.20 &51.09 &59.72 &44.72 &54.46 &70.00 &66.67 &0.00 &0.83 &47.56 &49.05 \\
SANA 1.5~\citep{xie2025sana}   &67.15 &65.73 &\textcolor{blue}{79.66} &77.08 &79.83 &77.83 &\textcolor{blue}{85.57} &\textcolor{red}{83.57} &73.57 &69.82 &61.50 &60.67 &65.32 &56.57 &69.96 &\textcolor{blue}{73.09} &62.96 &65.84 &\textcolor{blue}{80.00} &\textcolor{red}{80.00} &17.83 &15.83 &\textcolor{red}{71.07} &68.83 \\

FLUX.1 [dev]~\citep{flux2024}  &\textcolor{blue}{71.09} &\textcolor{red}{71.78} &\textcolor{red}{83.12} &\textcolor{red}{78.65} & \textcolor{red}{87.05} & \textcolor{red}{83.17} & \textcolor{red}{87.25} &\textcolor{blue}{80.39} &75.01 &72.39 &\textcolor{blue}{65.79} &\textcolor{red}{68.54} & \textcolor{blue}{67.07} &\textcolor{red}{73.69} &\textcolor{red}{73.84} &\textcolor{red}{73.34} &\textcolor{red}{69.09} &\textcolor{red}{71.59} & 66.67 & 66.67 &43.83 &\textcolor{red}{52.83} &\textcolor{blue}{70.72} &\textcolor{red}{71.47} \\

\midrule
\rowcolor{gray!10} \multicolumn{25}{c}{\textbf{Und. and Gen.}} \\ \hline

Show-o~\citep{showo} &59.72 &58.86 &73.08 &75.83 &74.83 &79.83 &78.82 &78.32 &65.57 &69.32 &53.67 &50.38 &60.95 &56.82 &68.59 &68.96 &66.46 &56.22 &63.33 &66.67 &3.83 &2.83 &55.02 &50.92 \\   
Janus-Pro-7B~\citep{chen2025januspro} &66.50 &65.02 &79.33 &78.25 &79.33 &\textcolor{blue}{82.33} &78.32 &73.32 &\textcolor{red}{80.32} &\textcolor{red}{79.07} &59.71 &58.82 &66.07 &56.20 &70.46 &70.84 &\textcolor{blue}{67.22} &59.97 &60.00 &70.00 &28.83 &33.83 &65.84 &60.25 \\

\rowcolor{green!10} \bf Lumina-DiMOO (Ours) &\textcolor{red}{71.27} &\textcolor{blue}{68.53} &75.5 &\textcolor{blue}{78.29} &77.00  &81.50 &74.20 &78.21 &\textcolor{blue}{75.29} &\textcolor{blue}{75.16} &\textcolor{red}{70.49} &\textcolor{blue}{68.33} &\textcolor{red}{75.99} &\textcolor{blue}{72.85} &\textcolor{blue}{70.73} &67.30 &65.79 &\textcolor{blue}{69.23} &73.33 &60.00 &\textcolor{blue}{59.28} &\textcolor{blue}{41.63} &69.78 &\textcolor{blue}{70.90} \\
\bottomrule
\end{tabular}\label{tab:tiif}
\end{adjustbox}
\end{table}
\paragraph{Evaluation Results on GenEval Benchmark.} Table~\ref{tab:geneval} presents a comparison of model performance on the GenEval~\citep{ghosh2024geneval} benchmark, which is designed to evaluate object-centric T2I generation using compositional prompts with diverse object attributes. 
Under identical evaluation settings, Lumina-DiMOO achieves an impressive 88\% overall score, surpassing both specialized generation models (FLUX.1 [Dev]: 82\%, Lumina-mGPT 2.0: 69\%) and unified models (Janus-Pro: 80\%, BAGEL: 82\%, and GPT-4o: 84\%), thereby setting new SOTA results. 
This success is largely attributed to Lumina-DiMOO’s enhanced capability in managing positional relationships and binding attributes. 
Compared to MMaDA, which features a similar architecture, Lumina-DiMOO demonstrates a substantial overall improvement of 25\% (88\% vs. 63\%). 
This significant advancement underscores the potential of the discrete diffusion architecture for practical applications. 
In addition, we validate the effectiveness of the proposed Self-GRPO on GenEval. 
Following the Self-GRPO training stage, Lumina-DiMOO demonstrates an overall improvement of 3\% on GenEval, with even more pronounced enhancements in ``Colors'' and ``Attribute''.

\vspace{-0.1in}
\paragraph{Evaluation Results on DPG Benchmark.} Table~\ref{tab:dpg} presents a performance comparison on the DPG~\citep{hu2024ella} benchmark, which includes 1,065 dense prompts designed for a detailed evaluation of various aspects of prompt adherence. 
Overall, Lumina-DiMOO achieves an impressive overall score of 86.04, surpassing all previous models and demonstrating its superior prompt-following abilities. 
In particular, Lumina-DiMOO excels in interpreting prompts that involve entities and relationships, outperforming all other models in the comparison. 
In addition, we evaluate MMaDA under the same settings on the DPG benchmark, its performance proves to be subpar, with a score of just 69.97.

\begin{figure*}[!th]
  \vspace*{-0.08\textwidth}
  \hspace*{-0.05\textwidth}
  \includegraphics[width=1.1\textwidth]{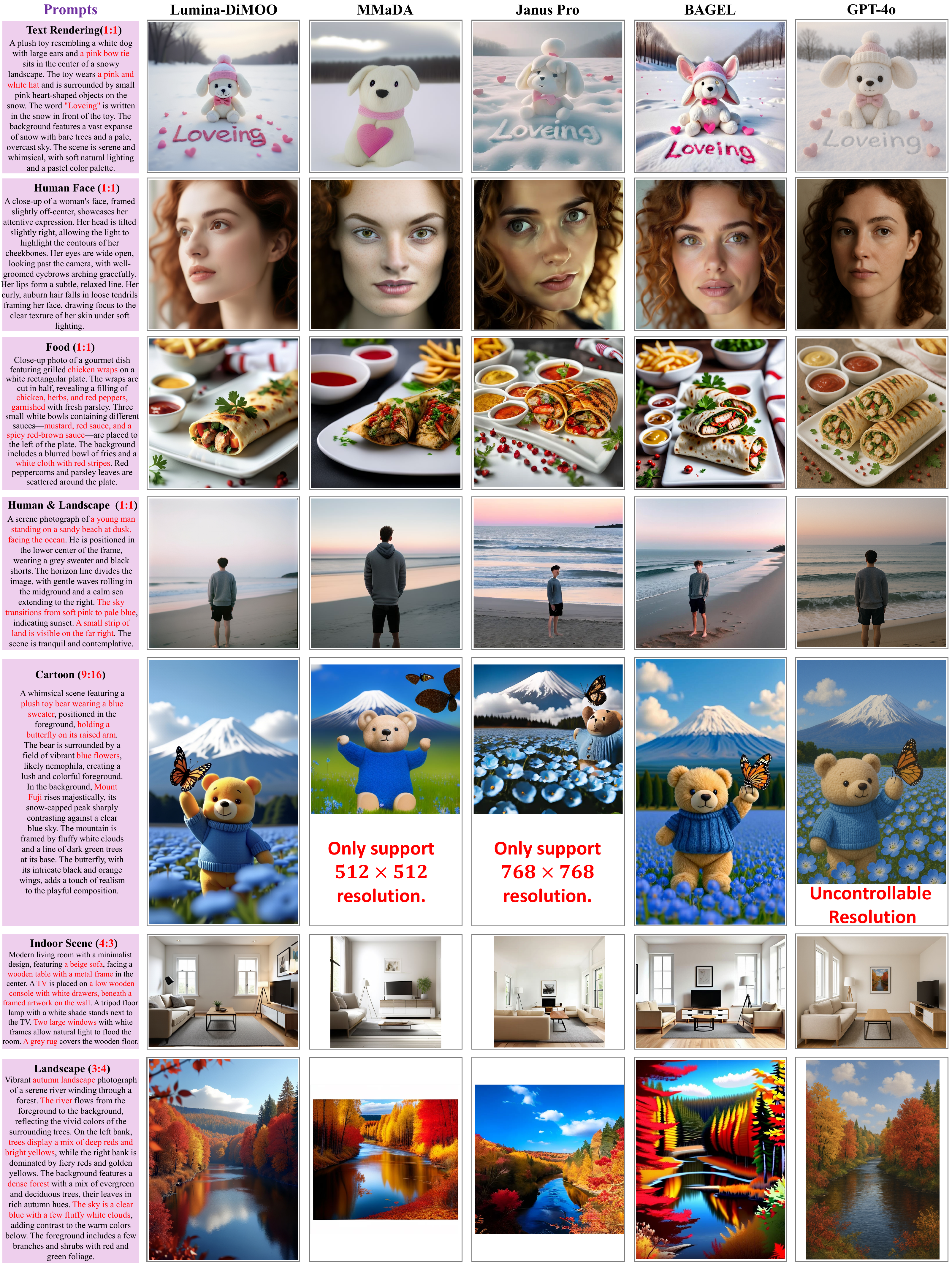}
  \vspace{-0.7cm}
  \caption{\textbf{Qualitative Comparison on Text-to-Image Generation.} We compare Lumina-DiMOO, MMaDA, Janus Pro, BAGEL, and GPT-4o across various common scenarios. Notably, MMaDA and Janus Pro lack support for arbitrary resolution generation.}
  \vspace{-0.7in}
    \label{fig:t2i}
\end{figure*}

\begin{figure*}[!th]
  \vspace*{-0.01\textwidth}
  \includegraphics[width=1.0\textwidth]{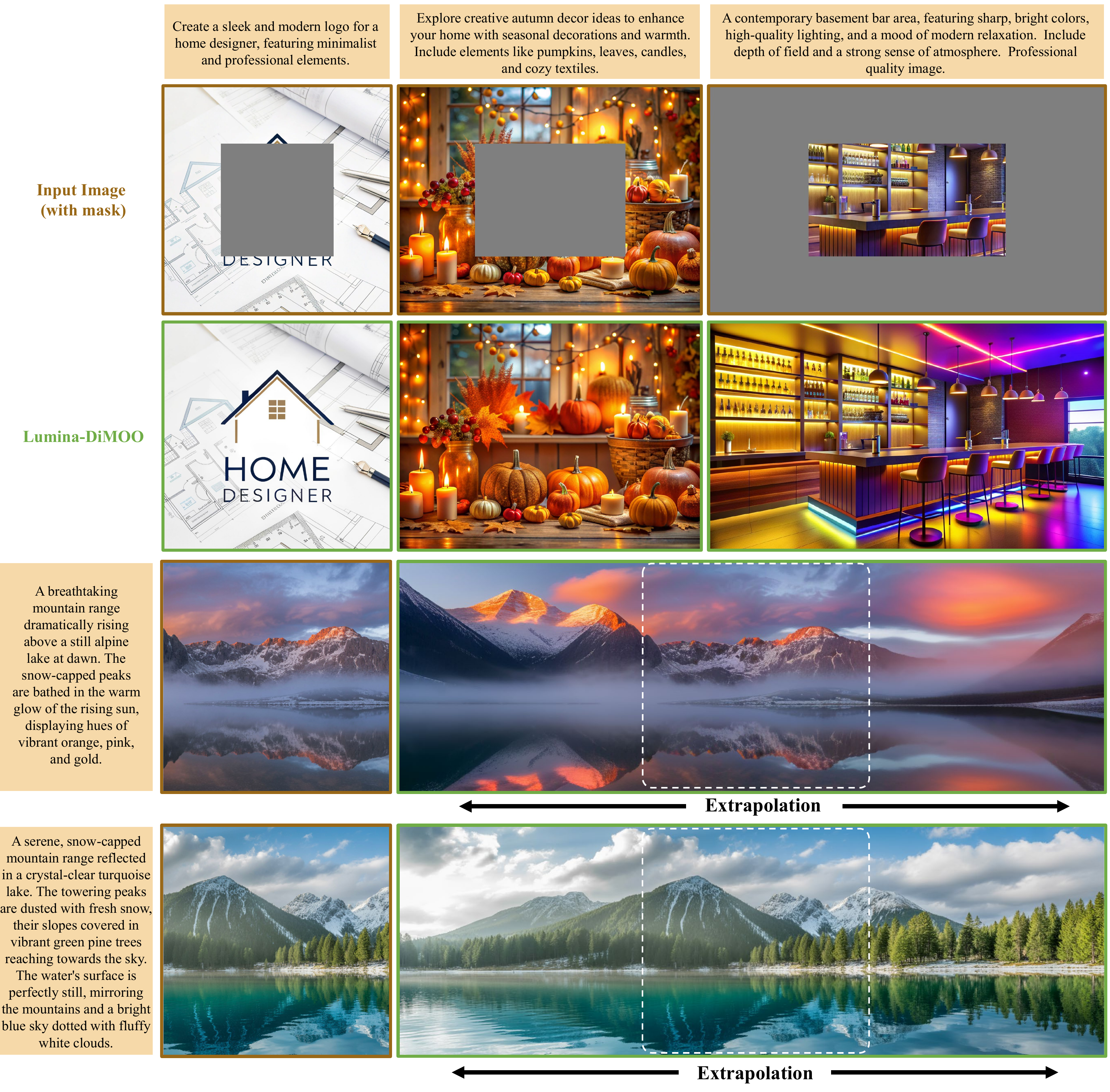}
  \vspace{-0.77cm}
  \caption{\centering \textbf{Qualitative Results on Text-guided Image Inpainting and Extrapolation.}}
    \label{fig:inpainting}
    \vspace{-0.15in}
\end{figure*}

\vspace{-0.15in}
\paragraph{Evaluation Results on UniGenBench Leaderboard.} UniGenBench~\citep{wang2025pref} is a newly unified benchmark for text-to-image generation that integrates diverse prompt themes with a comprehensive suite of fine-grained evaluation criteria. 
\textit{\textbf{The leaderboard is evaluated and maintained by the Tencent Hunyuan team.}} We extract evaluation results for various models from the leaderboard, as presented in Table~\ref{tab:unibench}. 
Lumina-DiMOO ranks among the top performers across all metrics, notably excelling in the ``Layout'' and ``Attribute'' categories, and surpassing all other models in overall evaluation scores. 
For a detailed evaluation across 27 dimensions, please refer to \href{https://huggingface.co/spaces/CodeGoat24/UniGenBench_Leaderboard}{Leaderboard Link}.

\vspace{-0.15in}
\paragraph{Evaluation Results on OneIG-EN Benchmark.} Table~\ref{tab:oneig_en} reports the quantitative results on the OneIG-EN~\citep{chang2025oneig} benchmark, a comprehensive evaluation framework specifically designed to assess the fine-grained performance of text-to-image models across multiple dimensions. 
For a fair comparison, we compute the overall score by averaging the results across all dimensions. 
Overall, Lumina-DiMOO achieves the highest average score and significantly surpasses other unified models such as BAGEL and Janus Pro, showcasing its robust capability in general-purpose image generation. 
Notably, it ranks first in the Alignment, Text, and Reasoning categories, highlighting its exceptional ability to follow prompts accurately and perform advanced reasoning. 

\vspace{-0.15in}
\paragraph{Evaluation Results on TIIF Benchmark.} Table~\ref{tab:tiif} shows the performance comparison on the TIIF testmini~\citep{wei2025tiif}, a benchmark designed to systematically assess the ability of text-to-image models to interpret and follow complex textual instructions. 
Overall, Lumina-DiMOO secures the second position, surpassed only by FLUX.1 [dev], a result that underscores its robust instruction-following capabilities.

\subsubsection{Qualitative Results}
\paragraph{Qualitative Comparisons.}
We conduct a qualitative comparison among Lumina-DiMOO, MMaDA, Janus-Pro 7B, BAGEL, and GPT-4o. As illustrated in Figure~\ref{fig:t2i}, Lumina-DiMOO consistently generates images of significantly higher quality compared to MMaDA and Janus-Pro 7B. 
Moreover, Lumina-DiMOO demonstrates exceptional flexibility in supporting any resolution, whereas MMaDA (limited to a fixed resolution of 512$\times$512), Janus-Pro (restricted to 768$\times$768), and GPT-4o (featuring uncontrollable resolution) show clear limitations in this aspect.

\vspace{-0.15in}
\paragraph{Image Inpainting and Extrapolation.} Due to the mask training paradigm of Lumina-DiMOO, it naturally supports text-guided image inpainting and extrapolation without requiring any fine-tuning. Examples are presented in Figure~\ref{fig:inpainting}. 
As shown on the top of the figure, given an input image with partial mask, Lumina-DiMOO is able to seamlessly inpaint the masked areas. 
Besides, Lumina-DiMOO is capable of extrapolating the original image horizontally or vertically based on the given text prompt (as illustrated in the third and fourth rows). 
These examples clearly highlight the inherent advantages of Lumina-DiMOO over Diffusion, AR or hybrid AR-Diffusion models in downstream applications.

\subsection{Performance of Image-to-Image Generation}
\newcommand{\best}[1]{\textbf{#1}}
\begin{table*}
  [!t] \small
  \centering
  \renewcommand{\arraystretch}{1.25}
  \setlength{\tabcolsep}{2mm} 
  \caption{\textbf{Evaluation of Controllable Generation Ability on Graph-200K~\citep{li2025visualcloze} benchmark.} 
  The methods that train a specialist for each task are marked as {\color{gray}{gray color}}. Except for these methods, we highlight the \textcolor{red}{best} and the \textcolor{blue}{second} results.}
  \resizebox{1.01\linewidth}{!}{
  \begin{tabular}{clccccccc}
    \toprule
    \multirow{2}{*}{\textbf{Condition}}               & \multirow{2}{*}{\textbf{Method}}  &  \multicolumn{2}{c}{\textbf{Controllability}}                          & \multicolumn{4}{c}{\textbf{Quality}} & \textbf{Text Consistency}                     \\
    \cmidrule(lr){3-4} \cmidrule(lr){5-8} \cmidrule(lr){9-9} 
     &  & $\text{F1} \uparrow$ & $\text{RMSE} \downarrow$ & $\text{FID}\downarrow$               & $\text{SSIM}\uparrow$               & $\text{MAN-IQA}\uparrow$            & $\text{MUSIQ}\uparrow$               & $\text{CLIP-Score}\uparrow$          \\
     \midrule 
     
     \multirow{7}{*}{Canny} & \color{gray} ControlNet~\citep{zhang2023adding} & \color{gray} {0.13} & \color{gray} - & \color{gray} 46.06  & \color{gray} 0.34 & \color{gray} 0.31 & \color{gray} 45.45 & \color{gray} 34.10 \\
     & \color{gray} OminiControl~\citep{tan2024ominicontrol}  & \color{gray} 0.47 & \color{gray} - & \color{gray} 29.58 & \color{gray} 0.61 & \color{gray} 0.44 & \color{gray} 61.40 & \color{gray} 34.40 \\
     
     & OmniGen~\citep{xiao2024omnigen} & \textcolor{blue}{0.43} & - &  51.58  &  {0.47} & \textcolor{blue}{0.47} & 62.66 & 33.66  \\
     
     & Lumina-mGPT~\citep{liu2024lumina} & 0.16 & - & 85.03 & 0.23 & \textcolor{red}{0.48} & \textcolor{red}{70.78} & 28.18\\
     
     & OneDiffusion~\citep{le2024diffusiongenerate} & 0.39 & - &32.76 & \textcolor{blue}{0.55} & 0.46 & 59.99 & \textcolor{red}{34.99} \\
     
     & Lumina-mGPT 2.0~\citep{xin2025lumina} & \textcolor{red}{0.49} & - & \textcolor{blue}{30.89} & 0.54 & 0.42 &63.18 & 34.44 \\

    \rowcolor{green!10} \cellcolor{white}
     & \textbf{Lumina-DiMOO (Ours)} &0.38 &- &\textcolor{red}{30.35} &\textcolor{red}{0.65} &0.41 &\textcolor{blue}{64.11} &\textcolor{blue}{34.56} \\
    \midrule \multirow{7}{*}{Depth}          & \color{gray} ControlNet~\citep{zhang2023adding} & \color{gray} - & \color{gray} 23.70 & \color{gray} 36.83 & \color{gray} 0.41 & \color{gray} 0.44 & \color{gray} 60.17 & \color{gray} 34.49 \\
    
     & \color{gray} OminiControl~\citep{tan2024ominicontrol}& \color{gray} - & \color{gray} 21.44 & \color{gray} 36.23 & \color{gray} 0.52 & \color{gray} 0.44 & \color{gray} 60.18 & \color{gray} 34.08 \\
     
     & OmniGen~\citep{xiao2024omnigen} & - &15.07    &  86.08   &    0.26  & \textcolor{red}{0.49} & \textcolor{blue}{64.90} & 29.72   \\
     
     & Lumina-mGPT~\citep{liu2024lumina} & - & 15.71  & 61.44 &0.34 & 0.38 & \textcolor{red}{69.72} & 31.58 \\
     
     & OneDiffusion~\citep{le2024diffusiongenerate} & - & \textcolor{blue}{10.35} & 39.03 & \textcolor{blue}{0.49} & \textcolor{red}{0.49} & 60.49 & \textcolor{red}{34.71}   \\
     
     & Lumina-mGPT 2.0~\citep{xin2025lumina} & - & 17.42 & \textcolor{blue}{36.52} & \textcolor{blue}{0.49} &0.39 & 59.52 &34.03  \\

     \rowcolor{green!10} \cellcolor{white}
     & \textbf{Lumina-DiMOO (Ours)} &- &\textcolor{red}{8.31} &\textcolor{red}{34.38} &\textcolor{red}{0.62} &\textcolor{blue}{0.40} &63.72 &\textcolor{blue}{34.54} \\
    \bottomrule
  \end{tabular}
  }
 
  \label{tab:control}
\end{table*}
We primarily evaluate our model using the Graph-200K~\citep{li2025visualcloze} and ImgEdit~\citep{ye2025imgedit} benchmarks. 
The Graph-200K benchmark enables comprehensive assessment across multiple tasks, including controllable generation, subject-driven generation, and style transfer, with an image style serving as a reference. 
In contrast, the ImgEdit benchmark focuses on evaluating the model’s proficiency in image editing tasks, such as adding, replacing, and removing objects, as well as changing the image style based on text descriptions.

\begin{table*}
  [!t] \small
  \centering
  \renewcommand{\arraystretch}{1.25}
  \setlength{\tabcolsep}{1mm}  
  \caption{\textbf{Evaluation of Style Transfer and Subject-Driven Generation Abilities on Graph-200K~\citep{li2025visualcloze} Benchmark and Image Editing Ability on ImgEdit~\citep{ye2025imgedit} Benchmark.} 
  Image editing metrics are evaluated by GPT-4.1.
  The methods that train a specialist for each task are marked as {\color{gray}{gray color}}. 
  Except for these methods, we highlight the \textcolor{red}{best} and the \textcolor{blue}{second} results.}
  \resizebox{1.0\linewidth}{!}{
  \begin{tabular}{lccccccccc}
    \toprule
     \multirow{2}{*}{\textbf{Method}}  &  \multicolumn{2}{c}{\textbf{Style Transfer (Img Reference)}}                          & \multicolumn{3}{c}{\textbf{Subject-Driven Generation}} & \multicolumn{4}{c}{\textbf{Image Editing}}                     \\
    \cmidrule(lr){2-3} \cmidrule(lr){4-6} \cmidrule(lr){7-10}
     & $\text{Text Alignment} \uparrow$ & $\text{Style Consistency} \uparrow$ & $\text{DINOv2}\uparrow$               & $\text{CLIP-I}\uparrow$               & $\text{CLIP-T}\uparrow$            & $\text{Add}\uparrow$               & $\text{Replace}\uparrow$       &$\text{Remove}\uparrow$ &$\text{Style}\uparrow$   \\
     \midrule 
     \color{gray}OminiControl~\citep{tan2024ominicontrol} & - & - &\color{gray}73.17 &\color{gray}87.70 &\color{gray}33.53\\
     
     \color{gray}InstantStyle~\citep{wang2024instantstyle} &\color{gray}0.27 &\color{gray}0.60 & - & - & -\\

     \color{gray}AnyEdit~\citep{yu2025anyedit} & - & - & - & - & - &\color{gray}3.18 &\color{gray}2.47 &\color{gray}2.23 &\color{gray}2.85\\
     
     OmniGen~\citep{xiao2024omnigen}   &\textcolor{blue}{0.27} &\textcolor{blue}{0.52} &67.73 &83.43 &34.53 &3.47 &2.94 &2.43 &4.19  \\
     
     Lumina-mGPT~\citep{liu2024lumina} &- & - &60.94 &70.63 &30.16 & - & - & - & -\\
     
     OneDiffusion~\citep{le2024diffusiongenerate} & - & - &73.88 &86.91 &\textcolor{red}{34.80} & - & - & - & -\\
     
     Lumina-mGPT 2.0~\citep{xin2025lumina} &- & - &\textcolor{blue}{76.60} &\textcolor{blue}{87.37} &33.90 & - & - & - & -\\

     BAGEL~\citep{bagel} &- &- &- &- &- &\textcolor{blue}{3.56} &3.30 &2.62 &\textcolor{red}{4.49} \\

     UniWorld-V1~\citep{lin2025uniworld} &- &- &- &- &- &\textcolor{red}{3.82} &\textcolor{blue}{3.47} &\textcolor{red}{3.24} &\textcolor{blue}{4.21}   \\
     
     \rowcolor{green!10}
     \textbf{Lumina-DiMOO (Ours)} &\textcolor{red}{0.32} &\textcolor{red}{0.53} &\textcolor{red}{80.57} &\textcolor{red}{89.36} &\textcolor{blue}{34.72} &\textcolor{red}{3.82} &\textcolor{red}{3.83} &\textcolor{blue}{2.76} &4.18 \\
    \bottomrule
  \end{tabular}
  }
  \label{tab:style_subject_editing}
  \vspace{-0.05in}
\end{table*}

\begin{figure*}[!th]
  \vspace*{-0.08\textwidth}
  \hspace*{-0.05\textwidth}
  \includegraphics[width=1.1\textwidth]{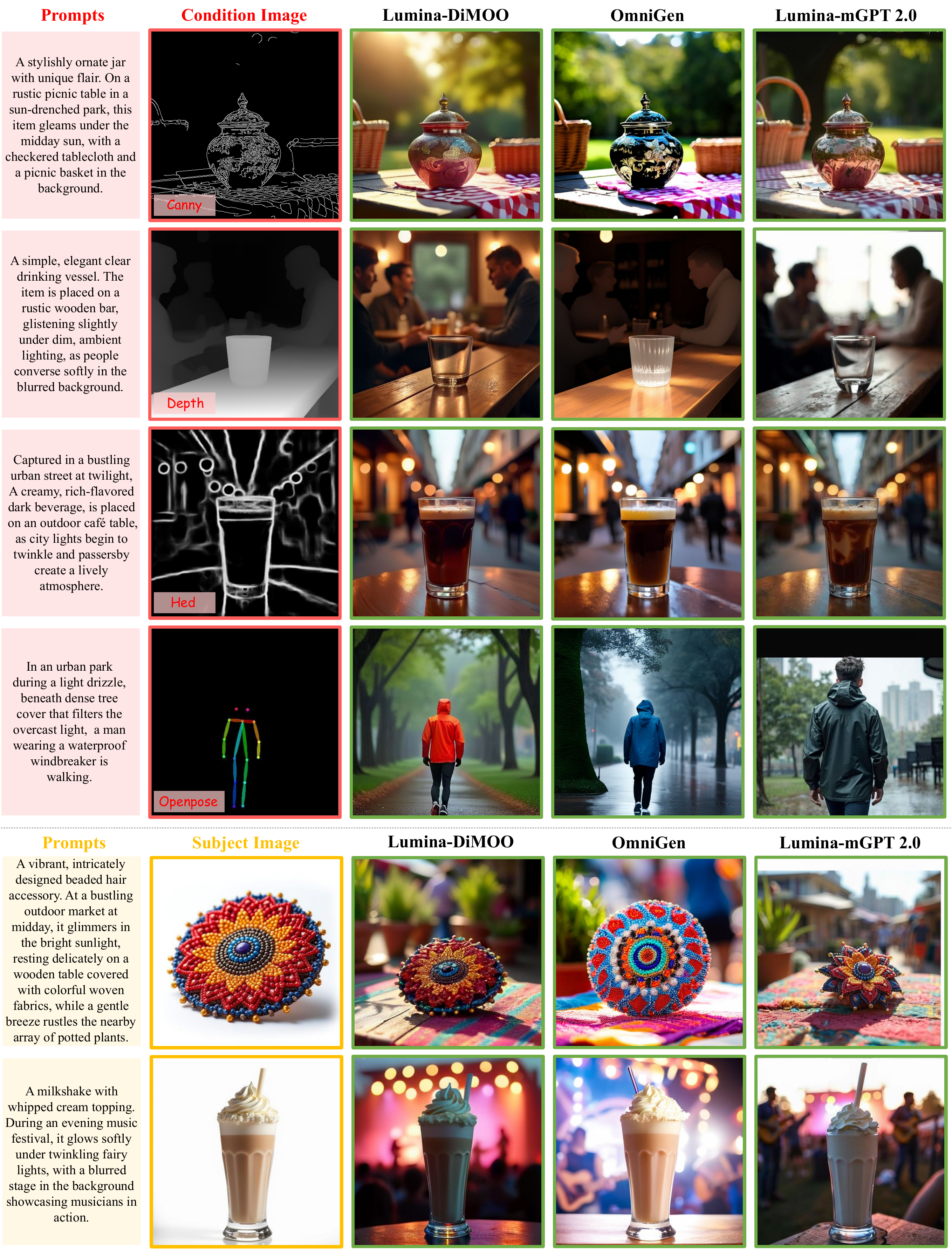}
  \vspace{-0.25in}
  \caption{\textbf{Qualitative Comparison on Controllable and Subject-Driven Generation Tasks.} We compare Lumina-DiMOO, BAGEL, and GPT-4o in object addition, removal, replacement, as well as background and style modification. Lumina-DiMOO performes well in terms of instruction adherence and resolution preservation.}
  \vspace{-0.5in}
    \label{fig:control_subject}
\end{figure*}

\subsubsection{Quantitative Results}
\paragraph{Evaluation Results of Controllable Generation.} For controllable generation, we evaluate the models based on three criteria: controllability (measured through F1-Score and RMSE), visual quality (measured through FID~\citep{heusel2017gans}, SSIM, MAN-IQA~\citep{yang2022maniqa}, and MUSIQ~\citep{ke2021musiq}), and text consistency (measured through CLIP-Score~\citep{radford2021learning}), following the evaluation approach of Graph-200K~\citep{li2025visualcloze}. 
As shown in Table~\ref{tab:control}, Lumina-DiMOO exhibits comparable controllability to other leading universal generative models (OmniGen, OneDiffusion, and Lumina-mGPT 2.0), while achieving superior visual quality and text consistency. 
Notably, when compared to specialized methods (ControlNet and OminiControl), Lumina-DiMOO performs on par with the best results and even outperforms them on the depth-to-image task.

\begin{figure*}[!th]
  \vspace*{-0.065\textwidth}
  \hspace*{-0.07\textwidth}
  \includegraphics[width=1.15\textwidth]{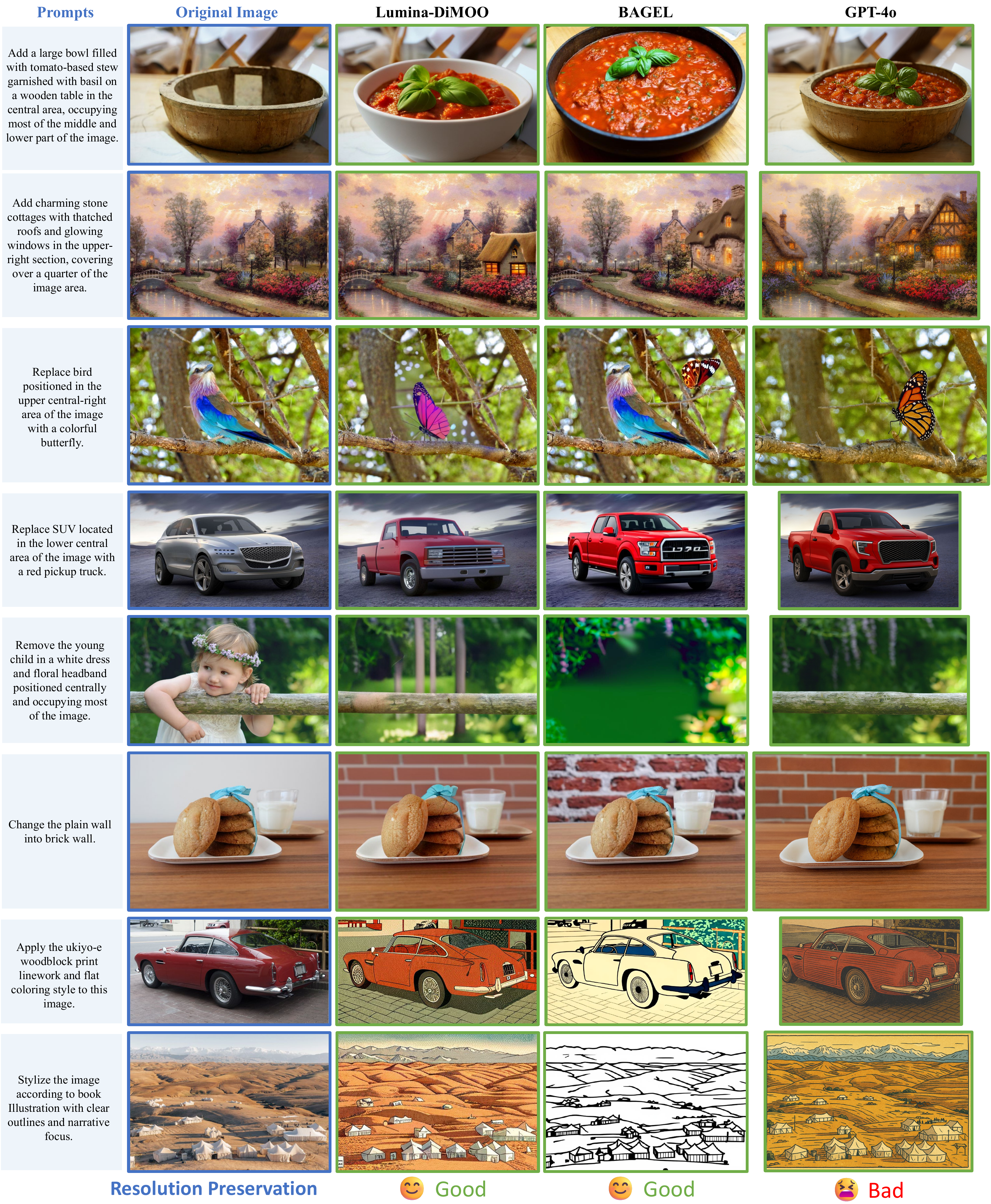}
  \caption{\textbf{Qualitative Comparison on Image Editing Tasks.} We compare Lumina-DiMOO, BAGEL, and GPT-4o in object addition, removal, replacement, as well as background and style modification. Lumina-DiMOO performed well in terms of instruction adherence and resolution preservation.}
  \vspace{-0.5in}
    \label{fig:editing}
\end{figure*}

\begin{figure*}[!t]
  \hspace*{-0.02\textwidth}
  \includegraphics[width=1.02\textwidth]{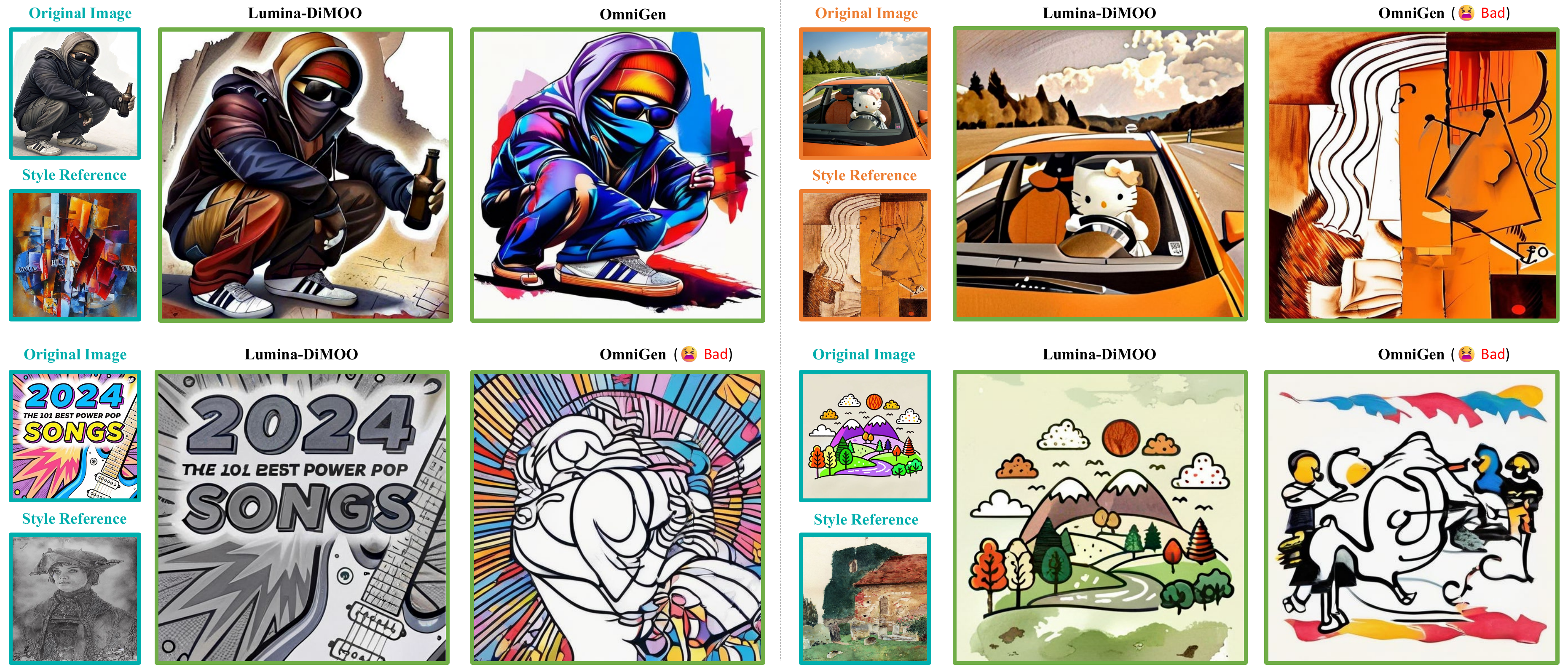}
  \vspace{-0.5cm}
  \caption{\textbf{Qualitative Comparison on Style Transfer Task.} Lumina-DiMOO completely outperforms OmniGen, which performs worse in most cases.}
  \vspace{-0.11in}
    \label{fig:styletransfer}
\end{figure*}

\vspace{-0.15in}
\paragraph{Evaluation Results of Style Transfer.} In the style transfer task, where an image serves as the style reference (as shown in Figure~\ref{fig:styletransfer}), we measure text consistency and style alignment using the CLIP \citep{radford2021learning} models on the Graph-200K benchmark. 
As presented in Table~\ref{tab:style_subject_editing}, Lumina-DiMOO exceeds OmniGen by 5\% and 1\% in text alignment and style consistency, respectively. 
Furthermore, when compared to InstantStyle, a specialized model, Lumina-DiMOO also achieves a 5\% improvement in text alignment, with a 7\% decrease in style alignment.

\vspace{-0.15in}
\paragraph{Evaluation Results of Subject-Driven Generation.} We also evaluate the models on Graph-200K specifically for subject-driven image generation and report semantic alignment using the DINOv2~\citep{oquab2023dinov2}, CLIP-I~\citep{radford2021learning}, and CLIP-T~\citep{radford2021learning} scores. 
As shown in Table~\ref{tab:style_subject_editing}, Lumina-DiMOO consistently demonstrates notable improvements across all these metrics. 
For example, compared to the previous SOTA model Lumina-mGPT 2.0, Lumina-DiMOO achieves improvements of 3.97\%, 1.99\%, and 0.82\% in these three scores.

\vspace{-0.15in}
\paragraph{Evaluation Results of Image Editing.} Table~\ref{tab:style_subject_editing} presents the results on the ImgEdit~\citep{ye2025imgedit} benchmark. 
We primarily focus on testing four common editing tasks: adding, removing, replacing, and changing style (text guidance). 
The evaluation metrics included instruction adherence, image-editing quality, and detail preservation, each scored on a scale from 1 to 5. These scores are assessed by GPT-4.1. 
Lumina-DiMOO performs exceptionally well in adding and replacing objects, surpassing other models (e.g., OmniGen, BAGEL, and UniWorld-V1). 
However, there remains room for improvement in tasks involving removing objects and changing styles.

\subsubsection{Qualitative Results}
We conduct a qualitative comparison on multiple image-to-image tasks between Lumina-DiMOO, OmniGen, Lumina-mGPT 2.0, BAGEL and GPT-4o. 
\textbf{1) Controllable Generation}: As illustrated at the top of Figure~\ref{fig:control_subject}, Lumina-DiMOO demonstrates precise generation capabilities under various control conditions. In contrast, OmniGen exhibits notable shortcomings in depth-to-image tasks, while Lumina mGPT 2.0 shows clear limitations in pose-to-image scenarios.
\textbf{2) Subject-Driven Generation}: As depicted at the
bottom of Figure~\ref{fig:control_subject}, Lumina-DiMOO excels in both object preservation and adherence to text instructions. 
\textbf{3) Style Transfer}: Lumina-DiMOO holds a distinct advantage over OmniGen in preserving the original image during style transfer, while also demonstrating superior comprehension and application of the reference image’s style, as shown in Figure~\ref{fig:styletransfer}. 
\textbf{4) Image Editing}: As shown in Figure~\ref{fig:editing}, Lumina-DiMOO performs well in tasks such as adding, removing, and replacing objects, as well as changing image backgrounds and styles. It also excels in preserving the resolution of the original image. BAGEL, on the other hand, falls slightly behind in object removal and style modification tasks. While GPT-4o demonstrates strong performance in editing tasks, there is still significant room for improvement in maintaining the resolution of the original image.

\begin{table}[!t]
    \centering
    \setlength{\tabcolsep}{3.5pt}
    \renewcommand{\arraystretch}{1.25}
    \scriptsize
    \caption{\textbf{Comparison with State-of-the-arts on Multimodal Understanding Benchmarks}. ``Und.'' and ``Gen.'' denote ``understanding'' and ``generation'', respectively. 
    We highlight the \textcolor{red}{best} and the \textcolor{blue}{second} results.
    }
    \label{sota_result_understanding}
    \resizebox{1.0\linewidth}{!}{
    \begin{tabular}{lcc|cccccc}
        \toprule
        \textbf{Model} & \textbf{Architecture} & \textbf{\# Params.} & \textbf{POPE$ \uparrow$} & \textbf{MME-P$ \uparrow$} & \textbf{MMB$ \uparrow$} & \textbf{SEED$ \uparrow$} & \textbf{MMMU$ \uparrow$} \\
        \midrule
        \rowcolor{gray!10}\multicolumn{8}{c}{\textit{Und. Only}} \\ \hline
        
        MobileVLM~\citep{chu2023mobilevlm} &AR & $1.4$B & $84.5$ & $1196.2$ & $53.2$ & - & -\\
        
        MobileVLM-V2~\citep{chu2024mobilevlm2} &AR & $1.4$B & $84.3$ & $1302.8$ & $57.7$ & - & - \\
        
        LLaVA-Phi~\citep{zhu2024llava} &AR & $2.7$B & $85.0$ & $1335.1$ & $59.8$ & - & -\\
        
        LLaVA~\citep{liu2024visual}&AR & $7$B & $76.3$ & $809.6$ & $38.7$ & $33.5$ & -\\
        
        LLaVA-v$1.5$~\citep{liu2023llava} &AR & $7$B & $85.9$ & $1510.7$ & $64.3$ & $58.6$& $35.4$\\
        
        InstructBLIP~\citep{instructblip}&AR & $7$B & - & - & $36.0$ & $53.4$ & - \\
        
        Qwen-VL-Chat~\citep{Qwen-VL} &AR & $7$B & - & $1487.5$ & $60.6$ & $58.2$ & - \\
        
        IDEFICS-$9$B~\citep{laurencon2023introducing}&AR & $8$B & - & - & $48.2$ & - & - \\
        Emu$3$-Chat~\citep{wang2024emu3} &AR & $8$B & $85.2$ & $1244$ & $58.5$ & $68.2$ & $31.6$ \\
        InstructBLIP~\citep{instructblip} &AR & $13$B & $78.9$ & $1212.8$ & - & - & - \\
        
        \midrule
        \rowcolor{gray!10}\multicolumn{8}{c}{\textit{Und. and Gen.}} \\ \hline

        Show-o~\citep{showo} &AR+Discrete Diff. & $1.3$B & $80.0$ & $1097.2$ & - & - & $26.7$ \\
        
        D-Dit~\citep{li2024dual} &Discrete Diff.+Diff. & $2.0$B & $84.0$ & $1124.7$ & - & - & - \\
        
        TokenFlow-XL~\citep{qu2024tokenflow} &AR &$13$B &\textcolor{blue}{$86.8$} &  $1545.9$ &  $68.9$ &  $68.7$ & $38.7$ \\
        
        VILA-U~\citep{wu2024vila}&AR & $7$B & $85.8$ & $1401.8$ & - & $59.0$ & - \\
        
        Chameleon~\citep{team2024chameleon}&AR & $7$B & - & - & - & - & $22.4$ \\
        
        Janus-Pro~\citep{chen2025januspro} &AR & $7$B & \textcolor{red}{$87.4$} & $1567.1$ & $79.2$ & \textcolor{blue}{$72.1$} & $41.0$ \\

        BLIP3-o~\citep{chen2025blip3} &AR+Diff. & $8$B &- &\textcolor{blue}{$1682.6$} &$83.5$ &$77.5$ &$50.6$ \\
        
        BAGEL~\citep{bagel}&AR+Diff. &14B &- &\textcolor{red}{$1687.0$} &\textcolor{red}{$85.0$} &- 	&\textcolor{blue}{$55.3$} \\

        Uniworld-V1~\citep{lin2025uniworld} &AR+Diff. &$20$B &- &- &$83.5$ &- &\textcolor{red}{$58.6$}\\ 
 
        OmniGen2~\citep{wu2025omnigen2} &AR+Diff. &$7$B &- &- &$79.1$ &- &$53.1$ \\
        
        MMaDA~\citep{yang2025mmada}&Discrete Diff. & $8$B &$86.1$ &$1410.7$ &$68.5$ &$64.2$ &$30.2$\\
        
        \rowcolor{green!10} \textbf{Lumina-DiMOO (Ours)} &Discrete Diff. & $8$B &\textcolor{red}{$87.4$} &$1534.2$ &\textcolor{blue}{$84.5$} &\textcolor{red}{$83.1$} &\textcolor{red}{$58.6$}\\
        \bottomrule
    \end{tabular}
    }
\end{table}
\vspace{-0.15in}
\subsection{Performance of Image Understanding}
To evaluate our model’s multimodal understanding capabilities, we evaluate it on five widely recognized vision-language benchmarks: POPE~\citep{li2023evaluating}, MME-P~\citep{yin2024survey}, MMBench~\citep{liu2024mmbench}, SEED~\citep{li2023seed}, and MMMU~\citep{yue2024mmmu}. 
Together, these benchmarks provide a concise yet comprehensive testbed that encompasses perception, cognition, and multimodal reasoning. 
They also possess strong discriminative power for ranking state-of-the-art models, ensuring a thorough assessment of performance.

\vspace{-0.05in}
\subsubsection{Quantitative Results}
We conduct a comprehensive comparison of Lumina-DiMOO with leading open-source multimodal models, covering both specialized models for visual understanding and general-purpose unified models. 
The results of visual understanding are detailed in Table~\ref{sota_result_understanding}. 
Compared with dedicated understanding-only models such as LLaVA-v1.5, Qwen-VL-Chat, Emu3-Chat, and InstructBLIP, our model achieves superior results across all benchmarks, despite being trained in a unified framework. 
When compared to other unified models (e.g., Show-o, VILA-U, Janus-Pro, BAGEL), Lumina-DiMOO consistently demonstrates outstanding performance, achieving leading scores in the POPE (87.4), SEED (83.1), and MMMU (58.6) benchmarks. 
In particular, Lumina-DiMOO significantly outperforms MMaDA (with similar architecture) across all benchmarks, highlighting the potential of a unified discrete diffusion architecture in bridging generation and understanding tasks.

\begin{figure*}[!h]
  \includegraphics[width=1.02\textwidth]{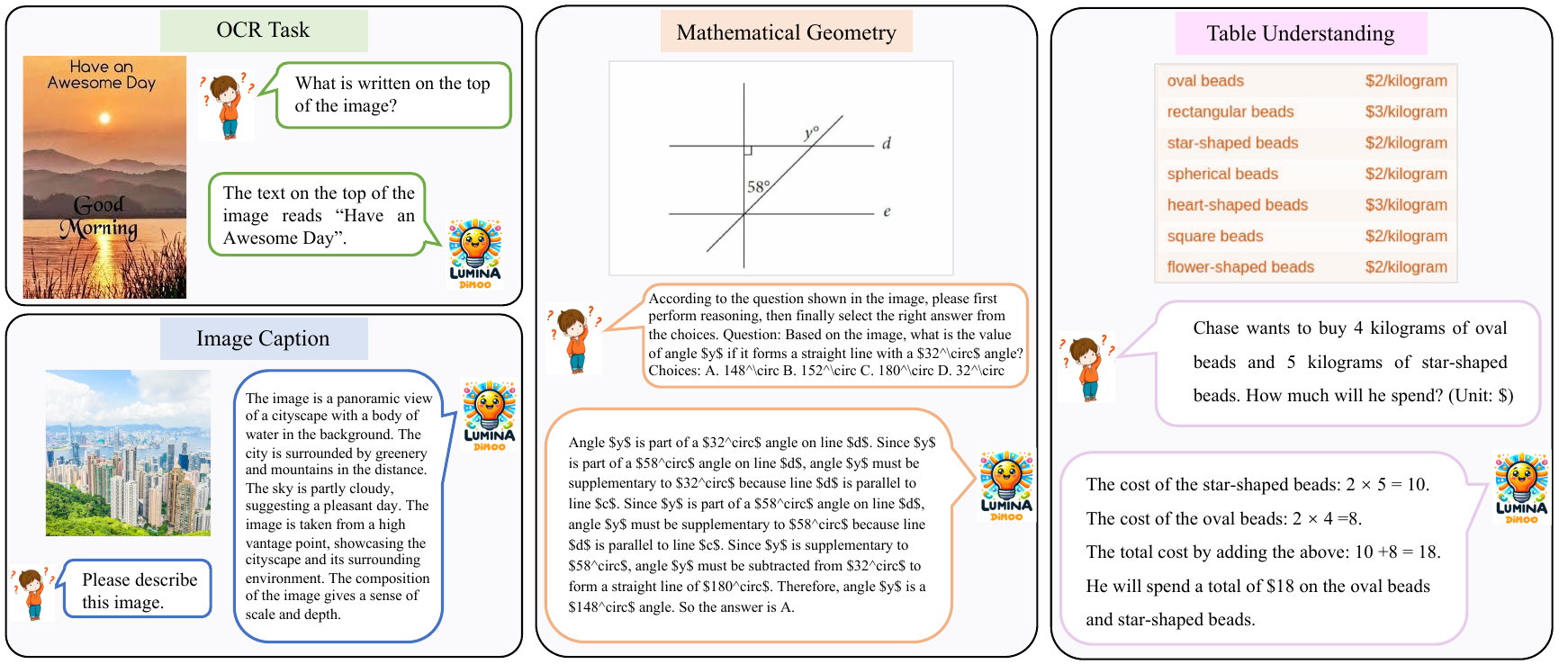}
  \vspace{-0.5cm}
  \caption{\textbf{Visualization of OCR, Image Caption, Mathematical Geometry, and Table Understanding Tasks.}}
    \label{fig:understanding}
    \vspace{-0.1in}
\end{figure*}

\vspace{-0.05in}
\subsubsection{Qualitative Results}
In addition to delivering comparable performance on various image understanding benchmarks, we visualize its capabilities across several understanding tasks, including OCR, captioning, mathematical geometry, and table understanding, as shown in Figure~\ref{fig:understanding}. 
The visualization results demonstrate that Lumina-DiMOO excels in text recognition accuracy, detailed image description, mathematical geometry, and the rational analysis of tables.

\section{Ablation and Extension}
\begin{figure*}[!t]
  \hspace*{-0.065\textwidth}
  \includegraphics[width=1.1\textwidth]{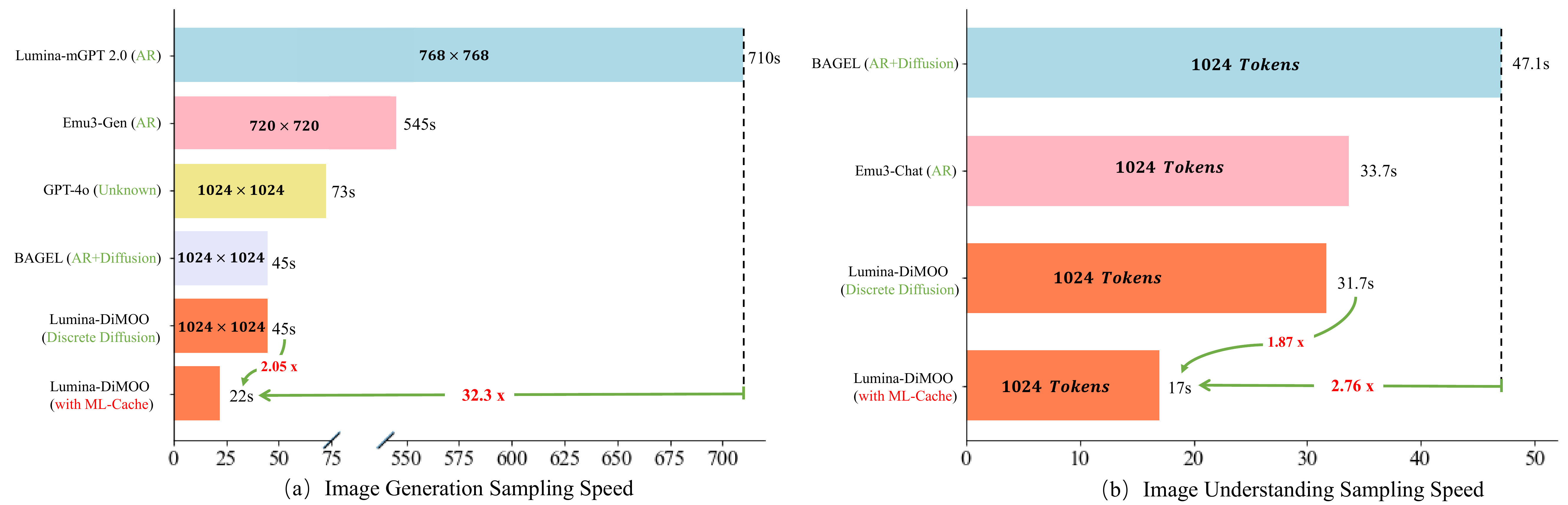}
  \vspace{-0.5cm}
  \caption{\textbf{Comparison of Sampling Time on Text-to-Image and Image Understanding Tasks.} For the text-to-image, Lumina-mGPT 2.0 generates images at a resolution of 768, Emu3 produces images at 720 resolution, while the other models utilize a 1024 resolution. In the image understanding task, all models consistently generate 1024 tokens.}
    \label{fig:speed}
    \vspace{-0.12in}
\end{figure*}

\subsection{Analysis of Sampling Speed}
\paragraph{Comparison with AR and Hybrid AR-Diffusion Models.} For text-to-image generation, we set 64 sampling steps for Lumina-DiMOO. 
As illustrated in Figure~\ref{fig:speed}\textcolor{magenta}{(a)}, Lumina-DiMOO’s sampling efficiency is several times higher than that of the AR models (Lumina-mGPT 2.0 and Emu3), and its sampling speed is roughly on par with BAGEL. 
If the sampling steps for Lumina-DiMOO are further reduced, its speed advantage becomes even more pronounced. 
On the other hand, for image understanding, we configure the block length to 256 and the number of sampling steps to 128 for Lumina-DiMOO. 
We find that the sampling speed advantage for image understanding is reduced, as shown in Figure~\ref{fig:speed}\textcolor{magenta}{(b)}. 
This is because text generation occurs in a block-wise manner, unlike image generation, which employs a single global decoding step. 
As a result, its speed is affected by both the number of blocks and the number of steps. Thus, the speed improvement in image understanding is not as substantial as in image generation. 
These observations highlight the promising potential of Lumina-DiMOO.

\vspace{-0.15in}
\paragraph{Effect of ML-Cache.} Under identical settings, we evaluate the sampling time of Lumina-DiMOO with and without the ML-Cache strategy, as shown in Figure~\ref{fig:speed}. 
The results demonstrate that ML-Cache significantly enhances the sampling process, boosting efficiency by a factor of 2.05 for text-to-image generation and 1.87 for image understanding. 
However, a minor drawback is that ML-Cache increases GPU usage, for example, from 38.9 GB to 45.9 GB when generating a 1024$\times$1024 image.

\subsection{Effect of Initialization From LLaDA}
\label{sec:initial}
In this work, Lumina-DiMOO builds upon LLaDA’s text capabilities and expands its functionalities in multi-modal generation and understanding, consistent with the paradigm of previous works~\citep{liquid,yang2025mmada}. 
However, previous studies~\citep{xin2025lumina} has also found that training from scratch without prior textual knowledge does not impact performance in autoregressive multimodal generation. 
To explore this further, we conduct additional ablation experiments to assess the necessity of inheriting text capabilities within the discrete diffusion framework.

\vspace{-0.15in}
\paragraph{Experimental Setup.} 
We design a comparative experiment with two key components: (1) \textbf{\textit{initializing Lumina-DiMOO using LLaDA-SFT}}~\citep{llada} and (2) \textit{\textbf{training Lumina-DiMOO from scratch}}. 
For model training, we randomly select a dataset from Section~\ref{sec:pretain-data} (Stage-III), consisting of 5M samples for visual generation and 5M samples for visual understanding. 
To conserve training resources, we omit the pre-training stage and directly engage in supervised fine-tuning on 256 resolution. 
For a fair comparison, we keep all training and evaluation parameters constant, except for model initialization.

\vspace{-0.15in}
\paragraph{Results.} In our evaluation of generation and understanding capabilities, we observe that training from scratch falls short in generating images or performing image understanding, often resulting in very large gradient norm during training. 
In contrast, initializing from LLaDA effectively supports both image generation and understanding, clearly demonstrating its superiority without requiring quantitative comparison.

\begin{figure*}[!t]
  \hspace*{-0.05\textwidth}
  \vspace{-0.2in}\includegraphics[width=1.1\textwidth]{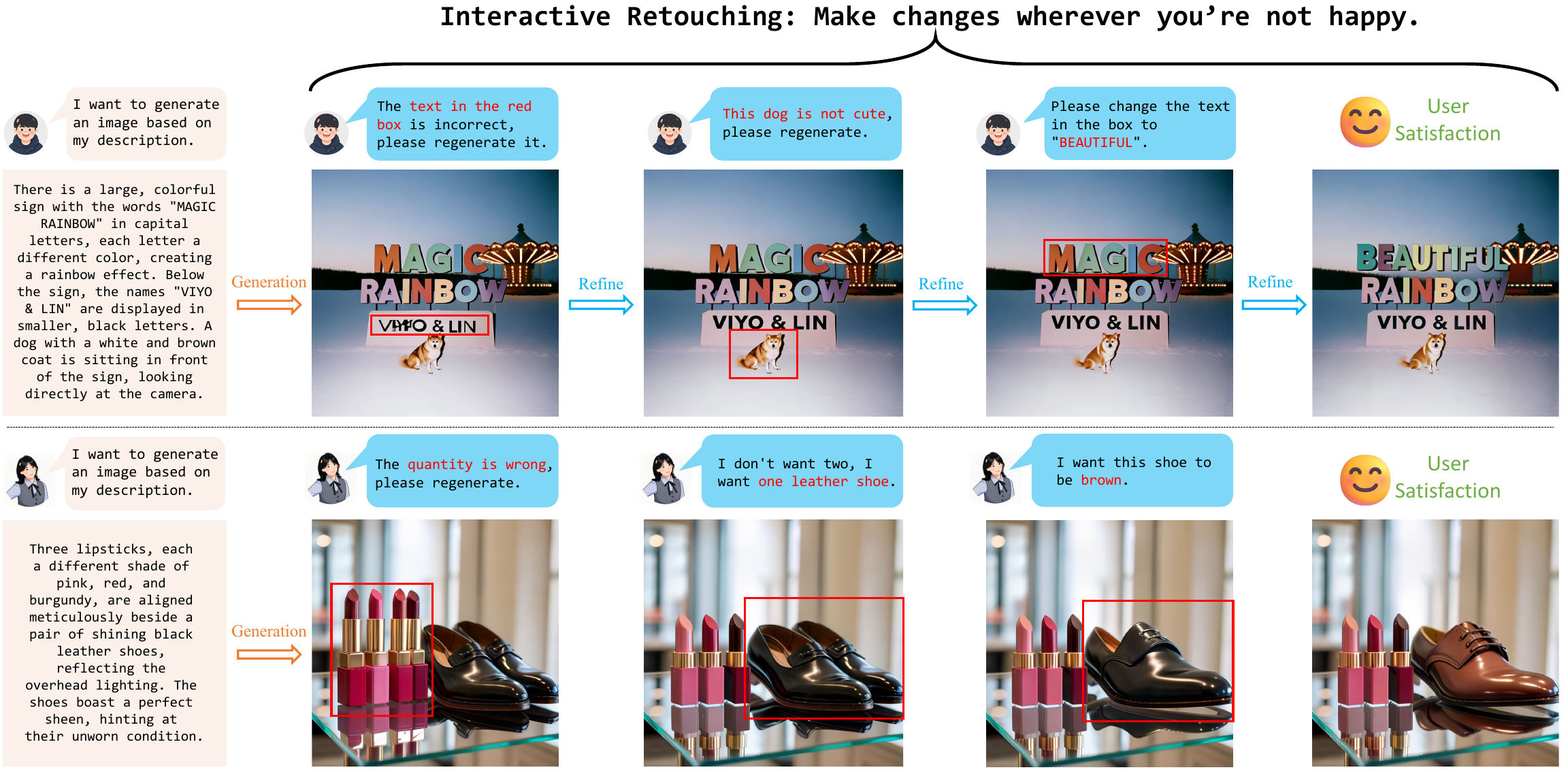}
  \vspace{-0.2cm}
  \caption{\textbf{Illustration of Interactive Retouching.} Users can repeatedly modify specific areas while keeping the surrounding regions unchanged until they reach satisfaction.}
  \vspace{-0.15in}
    \label{fig:interactive_retouching}
\end{figure*}

\subsection{Bringing New Ideas to Image Generation: Interactive Retouching}
Interactive Retouching stands out as a unique feature of Lumina-DiMOO, adept at allowing users to pinpoint specific areas for refinement through precise annotations, as illustrated in Figure~\ref{fig:interactive_retouching}. 
Lumina-DiMOO achieves this due to its unique discrete diffusion modeling paradigm, which allows it to mask user-annotated areas for regeneration. 
This process preserves all information in areas outside of the user's annotations, a feat previously unachievable with diffusion or AR generative models. 
While many commercial editing models exist, such as GPT-4o and Nana-Banana, none offer a 100\% guarantee of maintaining unchanged content outside the user’s specified annotations.

\section{Conclusion}
\label{sec:conclusion}
In this paper, we introduce Lumina-DiMOO, a unified foundation model for multi-modal understanding and generation. 
Lumina-DiMOO delivers top-tier performance on standard multi-modal generation and understanding benchmarks and stands out with its ultra-fast sampling speed and unique interactive retouching features. 
To further advance research in multi-modal and discrete diffusion research, we have open-sourced Lumina-DiMOO to the research community.

While Lumina-DiMOO currently demonstrates strong capabilities in image generation and understanding, our goal is to evolve it into a more comprehensive multi-modal model. 
In the future, we aim to expand Lumina-DiMOO to seamlessly integrate video, audio, and more modalities. 
Achieving this will require substantial research, particularly in creating a versatile tokenizer for diverse data types, designing the model architecture that processes temporal information, and developing advanced training techniques. Let us look forward to a more powerful Lumina-DiMOO.

\bibliography{arxiv}

\begin{thebibliography}{119}
\providecommand{\natexlab}[1]{#1}
\providecommand{\url}[1]{\texttt{#1}}
\expandafter\ifx\csname urlstyle\endcsname\relax
  \providecommand{\doi}[1]{doi: #1}\else
  \providecommand{\doi}{doi: \begingroup \urlstyle{rm}\Url}\fi

\bibitem[Austin et~al.(2021)Austin, Johnson, Ho, Tarlow, and Van Den~Berg]{d3pm}
Jacob Austin, Daniel~D Johnson, Jonathan Ho, Daniel Tarlow, and Rianne Van Den~Berg.
\newblock Structured denoising diffusion models in discrete state-spaces.
\newblock \emph{Advances in Neural Information Processing Systems (NeurIPS)}, 2021.

\bibitem[Bai et~al.(2024)Bai, Ye, Chow, Song, Li, Dong, Zhu, and Yan]{bai2024meissonic}
Jinbin Bai, Tian Ye, Wei Chow, Enxin Song, Xiangtai Li, Zhen Dong, Lei Zhu, and Shuicheng Yan.
\newblock Meissonic: Revitalizing masked generative transformers for efficient high-resolution text-to-image synthesis.
\newblock \emph{arXiv preprint arXiv:2410.08261}, 2024.

\bibitem[Bai et~al.(2023)Bai, Bai, Yang, Wang, Tan, Wang, Lin, Zhou, and Zhou]{Qwen-VL}
Jinze Bai, Shuai Bai, Shusheng Yang, Shijie Wang, Sinan Tan, Peng Wang, Junyang Lin, Chang Zhou, and Jingren Zhou.
\newblock Qwen-vl: A versatile vision-language model for understanding, localization, text reading, and beyond.
\newblock \emph{arXiv preprint arXiv:2308.12966}, 2023.

\bibitem[Bai et~al.(2025)Bai, Chen, Liu, Wang, Ge, Song, Dang, Wang, Wang, Tang, Zhong, Zhu, Yang, Li, Wan, Wang, Ding, Fu, Xu, Ye, Zhang, Xie, Cheng, Zhang, Yang, Xu, and Lin]{Qwen2.5-VL}
Shuai Bai, Keqin Chen, Xuejing Liu, Jialin Wang, Wenbin Ge, Sibo Song, Kai Dang, Peng Wang, Shijie Wang, Jun Tang, Humen Zhong, Yuanzhi Zhu, Mingkun Yang, Zhaohai Li, Jianqiang Wan, Pengfei Wang, Wei Ding, Zheren Fu, Yiheng Xu, Jiabo Ye, Xi~Zhang, Tianbao Xie, Zesen Cheng, Hang Zhang, Zhibo Yang, Haiyang Xu, and Junyang Lin.
\newblock Qwen2.5-vl technical report.
\newblock \emph{arXiv preprint arXiv:2502.13923}, 2025.

\bibitem[Betker et~al.(2023)Betker, Goh, Jing, Brooks, Wang, Li, Ouyang, Zhuang, Lee, Guo, et~al.]{dalle3}
James Betker, Gabriel Goh, Li~Jing, Tim Brooks, Jianfeng Wang, Linjie Li, Long Ouyang, Juntang Zhuang, Joyce Lee, Yufei Guo, et~al.
\newblock Improving image generation with better captions.
\newblock \emph{Computer Science. https://cdn. openai. com/papers/dall-e-3. pdf}, 2\penalty0 (3):\penalty0 8, 2023.

\bibitem[Chang et~al.(2022)Chang, Zhang, Jiang, Liu, and Freeman]{chang2022maskgit}
Huiwen Chang, Han Zhang, Lu~Jiang, Ce~Liu, and William~T Freeman.
\newblock Maskgit: Masked generative image transformer.
\newblock In \emph{Proceedings of the IEEE Conference on Computer Vision and Pattern Recognition (CVPR)}, 2022.

\bibitem[Chang et~al.(2025)Chang, Fang, Xing, Wu, Cheng, Wang, Zeng, Yu, and Chen]{chang2025oneig}
Jingjing Chang, Yixiao Fang, Peng Xing, Shuhan Wu, Wei Cheng, Rui Wang, Xianfang Zeng, Gang Yu, and Hai-Bao Chen.
\newblock Oneig-bench: Omni-dimensional nuanced evaluation for image generation.
\newblock \emph{arXiv preprint arXiv:2506.07977}, 2025.

\bibitem[Changpinyo et~al.(2021)Changpinyo, Sharma, Ding, and Soricut]{changpinyo2021conceptual}
Soravit Changpinyo, Piyush Sharma, Nan Ding, and Radu Soricut.
\newblock Conceptual 12m: Pushing web-scale image-text pre-training to recognize long-tail visual concepts.
\newblock In \emph{Proceedings of the IEEE Conference on Computer Vision and Pattern Recognition (CVPR)}, 2021.

\bibitem[Chen et~al.(2025{\natexlab{a}})Chen, Xu, Pan, Hu, Qin, Goldstein, Huang, Zhou, Xie, Savarese, et~al.]{chen2025blip3}
Jiuhai Chen, Zhiyang Xu, Xichen Pan, Yushi Hu, Can Qin, Tom Goldstein, Lifu Huang, Tianyi Zhou, Saining Xie, Silvio Savarese, et~al.
\newblock Blip3-o: A family of fully open unified multimodal models-architecture, training and dataset.
\newblock \emph{arXiv preprint arXiv:2505.09568}, 2025{\natexlab{a}}.

\bibitem[Chen et~al.(2023)Chen, Yu, Ge, Yao, Xie, Wu, Wang, Kwok, Luo, Lu, et~al.]{chen2023pixart}
Junsong Chen, Jincheng Yu, Chongjian Ge, Lewei Yao, Enze Xie, Yue Wu, Zhongdao Wang, James Kwok, Ping Luo, Huchuan Lu, et~al.
\newblock Pixart-$\alpha$: Fast training of diffusion transformer for photorealistic text-to-image synthesis.
\newblock \emph{Proceedings of the International Conference on Learning Representations (ICLR)}, 2023.

\bibitem[Chen et~al.(2025{\natexlab{b}})Chen, Cai, Chen, Chen, Ji, Wang, Yang, and Wang]{chen2025sharegpt}
Junying Chen, Zhenyang Cai, Pengcheng Chen, Shunian Chen, Ke~Ji, Xidong Wang, Yunjin Yang, and Benyou Wang.
\newblock Sharegpt-4o-image: Aligning multimodal models with gpt-4o-level image generation.
\newblock \emph{arXiv preprint arXiv:2506.18095}, 2025{\natexlab{b}}.

\bibitem[Chen et~al.(2025{\natexlab{c}})Chen, Wu, Liu, Pan, Liu, Xie, Yu, and Ruan]{chen2025januspro}
Xiaokang Chen, Zhiyu Wu, Xingchao Liu, Zizheng Pan, Wen Liu, Zhenda Xie, Xingkai Yu, and Chong Ruan.
\newblock Janus-pro: Unified multimodal understanding and generation with data and model scaling.
\newblock \emph{arXiv preprint arXiv:2501.17811}, 2025{\natexlab{c}}.

\bibitem[Chen et~al.(2024{\natexlab{a}})Chen, Wang, Cao, Liu, Gao, Cui, Zhu, Ye, Tian, Liu, et~al.]{chen2024expanding}
Zhe Chen, Weiyun Wang, Yue Cao, Yangzhou Liu, Zhangwei Gao, Erfei Cui, Jinguo Zhu, Shenglong Ye, Hao Tian, Zhaoyang Liu, et~al.
\newblock Expanding performance boundaries of open-source multimodal models with model, data, and test-time scaling.
\newblock \emph{arXiv preprint arXiv:2412.05271}, 2024{\natexlab{a}}.

\bibitem[Chen et~al.(2024{\natexlab{b}})Chen, Wang, Tian, Ye, Gao, Cui, Tong, Hu, Luo, Ma, et~al.]{chen2024far}
Zhe Chen, Weiyun Wang, Hao Tian, Shenglong Ye, Zhangwei Gao, Erfei Cui, Wenwen Tong, Kongzhi Hu, Jiapeng Luo, Zheng Ma, et~al.
\newblock How far are we to gpt-4v? closing the gap to commercial multimodal models with open-source suites.
\newblock \emph{arXiv preprint arXiv:2404.16821}, 2024{\natexlab{b}}.

\bibitem[Chen et~al.(2024{\natexlab{c}})Chen, Wu, Wang, Su, Chen, Xing, Zhong, Zhang, Zhu, Lu, et~al.]{chen2024internvl}
Zhe Chen, Jiannan Wu, Wenhai Wang, Weijie Su, Guo Chen, Sen Xing, Muyan Zhong, Qinglong Zhang, Xizhou Zhu, Lewei Lu, et~al.
\newblock Internvl: Scaling up vision foundation models and aligning for generic visual-linguistic tasks.
\newblock In \emph{Proceedings of the IEEE/CVF Conference on Computer Vision and Pattern Recognition}, pages 24185--24198, 2024{\natexlab{c}}.

\bibitem[Cho et~al.(2024)Cho, Hu, Baldridge, Garg, Anderson, Krishna, Bansal, Pont{-}Tuset, and Wang]{2024dsg}
Jaemin Cho, Yushi Hu, Jason~M. Baldridge, Roopal Garg, Peter Anderson, Ranjay Krishna, Mohit Bansal, Jordi Pont{-}Tuset, and Su~Wang.
\newblock Davidsonian scene graph: Improving reliability in fine-grained evaluation for text-to-image generation.
\newblock In \emph{Proceedings of the International Conference on Learning Representations (ICLR)}, 2024.

\bibitem[Chu et~al.(2023)Chu, Qiao, Lin, Xu, Yang, Hu, Wei, Zhang, Zhang, Wei, et~al.]{chu2023mobilevlm}
Xiangxiang Chu, Limeng Qiao, Xinyang Lin, Shuang Xu, Yang Yang, Yiming Hu, Fei Wei, Xinyu Zhang, Bo~Zhang, Xiaolin Wei, et~al.
\newblock Mobilevlm: A fast, reproducible and strong vision language assistant for mobile devices.
\newblock \emph{arXiv preprint arXiv:2312.16886}, 2023.

\bibitem[Chu et~al.(2024)Chu, Qiao, Zhang, Xu, Wei, Yang, Sun, Hu, Lin, Zhang, et~al.]{chu2024mobilevlm2}
Xiangxiang Chu, Limeng Qiao, Xinyu Zhang, Shuang Xu, Fei Wei, Yang Yang, Xiaofei Sun, Yiming Hu, Xinyang Lin, Bo~Zhang, et~al.
\newblock Mobilevlm v2: Faster and stronger baseline for vision language model.
\newblock \emph{arXiv preprint arXiv:2402.03766}, 2024.

\bibitem[Dai et~al.(2023)Dai, Li, Li, Tiong, Zhao, Wang, Li, Fung, and Hoi]{instructblip}
Wenliang Dai, Junnan Li, Dongxu Li, Anthony Meng~Huat Tiong, Junqi Zhao, Weisheng Wang, Boyang Li, Pascale Fung, and Steven Hoi.
\newblock Instructblip: Towards general-purpose vision-language models with instruction tuning, 2023.

\bibitem[Deng et~al.(2025)Deng, Zhu, Li, Gou, Li, Wang, Zhong, Yu, Nie, Song, et~al.]{bagel}
Chaorui Deng, Deyao Zhu, Kunchang Li, Chenhui Gou, Feng Li, Zeyu Wang, Shu Zhong, Weihao Yu, Xiaonan Nie, Ziang Song, et~al.
\newblock Emerging properties in unified multimodal pretraining.
\newblock \emph{arXiv preprint arXiv:2505.14683}, 2025.

\bibitem[Devlin(2018)]{devlin2018bert}
Jacob Devlin.
\newblock Bert: Pre-training of deep bidirectional transformers for language understanding.
\newblock \emph{arXiv preprint arXiv:1810.04805}, 2018.

\bibitem[Esser et~al.(2024)Esser, Kulal, Blattmann, Entezari, M{\"u}ller, Saini, Levi, Lorenz, Sauer, Boesel, et~al.]{esser2024scaling}
Patrick Esser, Sumith Kulal, Andreas Blattmann, Rahim Entezari, Jonas M{\"u}ller, Harry Saini, Yam Levi, Dominik Lorenz, Axel Sauer, Frederic Boesel, et~al.
\newblock Scaling rectified flow transformers for high-resolution image synthesis.
\newblock In \emph{Proceedings of the International Conference on Machine Learning (ICML)}, 2024.

\bibitem[Ge et~al.(2024)Ge, Zhao, Zhu, Ge, Yi, Song, Li, Ding, and Shan]{ge2024seed}
Yuying Ge, Sijie Zhao, Jinguo Zhu, Yixiao Ge, Kun Yi, Lin Song, Chen Li, Xiaohan Ding, and Ying Shan.
\newblock Seed-x: Multimodal models with unified multi-granularity comprehension and generation.
\newblock \emph{arXiv preprint arXiv:2404.14396}, 2024.

\bibitem[Ghosh et~al.(2024)Ghosh, Hajishirzi, and Schmidt]{ghosh2024geneval}
Dhruba Ghosh, Hannaneh Hajishirzi, and Ludwig Schmidt.
\newblock Geneval: An object-focused framework for evaluating text-to-image alignment.
\newblock \emph{Advances in Neural Information Processing Systems (NeurIPS)}, 2024.

\bibitem[Grattafiori et~al.(2024)Grattafiori, Dubey, Jauhri, Pandey, Kadian, Al-Dahle, Letman, Mathur, Schelten, Vaughan, et~al.]{llama3}
Aaron Grattafiori, Abhimanyu Dubey, Abhinav Jauhri, Abhinav Pandey, Abhishek Kadian, Ahmad Al-Dahle, Aiesha Letman, Akhil Mathur, Alan Schelten, Alex Vaughan, et~al.
\newblock The llama 3 herd of models.
\newblock \emph{arXiv preprint arXiv:2407.21783}, 2024.

\bibitem[Guo et~al.(2025)Guo, Yang, Zhang, Song, Zhang, Xu, Zhu, Ma, Wang, Bi, et~al.]{guo2025deepseek}
Daya Guo, Dejian Yang, Haowei Zhang, Junxiao Song, Ruoyu Zhang, Runxin Xu, Qihao Zhu, Shirong Ma, Peiyi Wang, Xiao Bi, et~al.
\newblock Deepseek-r1: Incentivizing reasoning capability in llms via reinforcement learning.
\newblock \emph{arXiv preprint arXiv:2501.12948}, 2025.

\bibitem[Guo et~al.(2024)Guo, Zheng, Bai, Li, Wang, Zhu, Li, Neubig, Chen, and Yue]{guo2024mammoth}
Jarvis Guo, Tuney Zheng, Yuelin Bai, Bo~Li, Yubo Wang, King Zhu, Yizhi Li, Graham Neubig, Wenhu Chen, and Xiang Yue.
\newblock Mammoth-vl: Eliciting multimodal reasoning with instruction tuning at scale.
\newblock \emph{arXiv preprint arXiv:2412.05237}, 2024.

\bibitem[Heusel et~al.(2017)Heusel, Ramsauer, Unterthiner, Nessler, and Hochreiter]{heusel2017gans}
Martin Heusel, Hubert Ramsauer, Thomas Unterthiner, Bernhard Nessler, and Sepp Hochreiter.
\newblock Gans trained by a two time-scale update rule converge to a local nash equilibrium.
\newblock \emph{Advances in Neural Information Processing Systems (NeurIPS)}, 2017.

\bibitem[Hu et~al.(2024)Hu, Wang, Fang, Fu, Cheng, and Yu]{hu2024ella}
Xiwei Hu, Rui Wang, Yixiao Fang, Bin Fu, Pei Cheng, and Gang Yu.
\newblock Ella: Equip diffusion models with llm for enhanced semantic alignment.
\newblock \emph{arXiv preprint arXiv:2403.05135}, 2024.

\bibitem[Huang et~al.(2025{\natexlab{a}})Huang, Guo, Wang, Yi, Ma, Cao, and Sheng]{huang2024mvadapter}
Zehuan Huang, Yuanchen Guo, Haoran Wang, Ran Yi, Lizhuang Ma, Yan-Pei Cao, and Lu~Sheng.
\newblock Mv-adapter: Multi-view consistent image generation made easy.
\newblock \emph{Proceedings of the IEEE International Conference on Computer Vision (ICCV)}, 2025{\natexlab{a}}.

\bibitem[Huang et~al.(2025{\natexlab{b}})Huang, Chen, Wang, Li, and Qi]{dcolt}
Zemin Huang, Zhiyang Chen, Zijun Wang, Tiancheng Li, and Guo-Jun Qi.
\newblock Reinforcing the diffusion chain of lateral thought with diffusion language models.
\newblock \emph{arXiv preprint arXiv:2505.10446}, 2025{\natexlab{b}}.

\bibitem[Huang et~al.(2025{\natexlab{c}})Huang, Wu, Lin, Zhang, Chen, and Wu]{huang2025autogeo}
Zihan Huang, Tao Wu, Wang Lin, Shengyu Zhang, Jingyuan Chen, and Fei Wu.
\newblock Autogeo: Automating geometric image dataset creation for enhanced geometry understanding.
\newblock \emph{IEEE Transactions on Multimedia (TMM)}, 2025{\natexlab{c}}.

\bibitem[Ke et~al.(2021)Ke, Wang, Wang, Milanfar, and Yang]{ke2021musiq}
Junjie Ke, Qifei Wang, Yilin Wang, Peyman Milanfar, and Feng Yang.
\newblock Musiq: Multi-scale image quality transformer.
\newblock In \emph{Proceedings of the IEEE International Conference on Computer Vision (ICCV)}, 2021.

\bibitem[Labs(2024)]{flux2024}
Black~Forest Labs.
\newblock Flux.
\newblock \url{https://github.com/black-forest-labs/flux}, 2024.

\bibitem[Laurençon et~al.(2023)Laurençon, van Strien, Bekman, Tronchon, Saulnier, Wang, Karamcheti, Singh, Pistilli, Jernite, and et~al.]{laurencon2023introducing}
Hugo Laurençon, Daniel van Strien, Stas Bekman, Leo Tronchon, Lucile Saulnier, Thomas Wang, Siddharth Karamcheti, Amanpreet Singh, Giada Pistilli, Yacine Jernite, and et~al.
\newblock Introducing idefics: An open reproduction of state-of-the-art visual language model, 2023.
\newblock URL \url{https://huggingface.co/blog/idefics}.

\bibitem[Le et~al.(2025)Le, Pham, Lee, Clark, Kembhavi, Mandt, Krishna, and Lu]{le2024diffusiongenerate}
Duong~H. Le, Tuan Pham, Sangho Lee, Christopher Clark, Aniruddha Kembhavi, Stephan Mandt, Ranjay Krishna, and Jiasen Lu.
\newblock One diffusion to generate them all.
\newblock \emph{Proceedings of the IEEE Conference on Computer Vision and Pattern Recognition (CVPR)}, 2025.

\bibitem[Li et~al.(2023{\natexlab{a}})Li, Wang, Wang, Ge, Ge, and Shan]{li2023seed}
Bohao Li, Rui Wang, Guangzhi Wang, Yuying Ge, Yixiao Ge, and Ying Shan.
\newblock Seed-bench: Benchmarking multimodal llms with generative comprehension.
\newblock \emph{arXiv preprint arXiv:2307.16125}, 2023{\natexlab{a}}.

\bibitem[Li et~al.(2024{\natexlab{a}})Li, Kamko, Akhgari, Sabet, Xu, and Doshi]{li2024playground}
Daiqing Li, Aleks Kamko, Ehsan Akhgari, Ali Sabet, Linmiao Xu, and Suhail Doshi.
\newblock Playground v2. 5: Three insights towards enhancing aesthetic quality in text-to-image generation.
\newblock \emph{arXiv preprint arXiv:2402.17245}, 2024{\natexlab{a}}.

\bibitem[Li et~al.(2025{\natexlab{a}})Li, Kallidromitis, Bansal, Gokul, Kato, Kozuka, Kuen, Lin, Chang, and Grover]{lavida}
Shufan Li, Konstantinos Kallidromitis, Hritik Bansal, Akash Gokul, Yusuke Kato, Kazuki Kozuka, Jason Kuen, Zhe Lin, Kai-Wei Chang, and Aditya Grover.
\newblock {LaViDa}: A large diffusion language model for multimodal understanding.
\newblock \emph{arXiv preprint arXiv:2505.16839}, 2025{\natexlab{a}}.

\bibitem[Li et~al.(2023{\natexlab{b}})Li, Du, Zhou, Wang, Zhao, and Wen]{li2023evaluating}
Yifan Li, Yifan Du, Kun Zhou, Jinpeng Wang, Wayne~Xin Zhao, and Ji-Rong Wen.
\newblock Evaluating object hallucination in large vision-language models.
\newblock In \emph{Proceedings of the Conference on Empirical Methods in Natural Language Processing (EMNLP)}, 2023{\natexlab{b}}.

\bibitem[Li et~al.(2025{\natexlab{b}})Li, Du, Yan, Zhuo, Li, Gao, Ma, and Cheng]{li2025visualcloze}
Zhong-Yu Li, Ruoyi Du, Juncheng Yan, Le~Zhuo, Zhen Li, Peng Gao, Zhanyu Ma, and Ming-Ming Cheng.
\newblock Visualcloze: A universal image generation framework via visual in-context learning.
\newblock \emph{Proceedings of the IEEE International Conference on Computer Vision (ICCV)}, 2025{\natexlab{b}}.

\bibitem[Li et~al.(2024{\natexlab{b}})Li, Li, Shi, Farimani, Kluger, Yang, and Wang]{li2024dual}
Zijie Li, Henry Li, Yichun Shi, Amir~Barati Farimani, Yuval Kluger, Linjie Yang, and Peng Wang.
\newblock Dual diffusion for unified image generation and understanding.
\newblock \emph{arXiv preprint arXiv:2501.00289}, 2024{\natexlab{b}}.

\bibitem[Liao et~al.(2025)Liao, Liu, Wang, Luo, Zhang, Zhao, Wu, Li, Tian, and Huang]{mogao}
Chao Liao, Liyang Liu, Xun Wang, Zhengxiong Luo, Xinyu Zhang, Wenliang Zhao, Jie Wu, Liang Li, Zhi Tian, and Weilin Huang.
\newblock Mogao: An omni foundation model for interleaved multi-modal generation.
\newblock \emph{arXiv preprint arXiv:2505.05472}, 2025.

\bibitem[Lin et~al.(2025)Lin, Li, Cheng, Niu, Ye, He, Yuan, Yu, Wang, Ge, et~al.]{lin2025uniworld}
Bin Lin, Zongjian Li, Xinhua Cheng, Yuwei Niu, Yang Ye, Xianyi He, Shenghai Yuan, Wangbo Yu, Shaodong Wang, Yunyang Ge, et~al.
\newblock Uniworld: High-resolution semantic encoders for unified visual understanding and generation.
\newblock \emph{arXiv preprint arXiv:2506.03147}, 2025.

\bibitem[Liu et~al.(2025{\natexlab{a}})Liu, Broadrick, Niepert, and den Broeck]{copula}
Anji Liu, Oliver Broadrick, Mathias Niepert, and Guy~Van den Broeck.
\newblock Discrete copula diffusion.
\newblock In \emph{Proceedings of the International Conference on Learning Representations (ICLR)}, 2025{\natexlab{a}}.

\bibitem[Liu et~al.(2024{\natexlab{a}})Liu, Zhao, Zhuo, Lin, Qiao, Li, and Gao]{liu2024lumina}
Dongyang Liu, Shitian Zhao, Le~Zhuo, Weifeng Lin, Yu~Qiao, Hongsheng Li, and Peng Gao.
\newblock Lumina-mgpt: Illuminate flexible photorealistic text-to-image generation with multimodal generative pretraining.
\newblock \emph{arXiv preprint arXiv:2408.02657}, 2024{\natexlab{a}}.

\bibitem[Liu et~al.(2024{\natexlab{b}})Liu, Yan, Zaharia, and Abbeel]{liu2024world}
Hao Liu, Wilson Yan, Matei Zaharia, and Pieter Abbeel.
\newblock World model on million-length video and language with ringattention.
\newblock \emph{arXiv preprint arXiv:2402.08268}, 2024{\natexlab{b}}.

\bibitem[Liu et~al.(2023)Liu, Li, Wu, and Lee]{liu2023llava}
Haotian Liu, Chunyuan Li, Qingyang Wu, and Yong~Jae Lee.
\newblock Visual instruction tuning, 2023.

\bibitem[Liu et~al.(2024{\natexlab{c}})Liu, Li, Li, and Lee]{liu2023improvedllava}
Haotian Liu, Chunyuan Li, Yuheng Li, and Yong~Jae Lee.
\newblock Improved baselines with visual instruction tuning.
\newblock In \emph{Proceedings of the IEEE Conference on Computer Vision and Pattern Recognition (CVPR)}, 2024{\natexlab{c}}.

\bibitem[Liu et~al.(2024{\natexlab{d}})Liu, Li, Li, Li, Zhang, Shen, and Lee]{liu2024llavanext}
Haotian Liu, Chunyuan Li, Yuheng Li, Bo~Li, Yuanhan Zhang, Sheng Shen, and Yong~Jae Lee.
\newblock Llava-next: Improved reasoning, ocr, and world knowledge, 2024{\natexlab{d}}.
\newblock URL \url{https://llava-vl.github.io/blog/2024-01-30-llava-next/}.

\bibitem[Liu et~al.(2024{\natexlab{e}})Liu, Li, Wu, and Lee]{liu2024visual}
Haotian Liu, Chunyuan Li, Qingyang Wu, and Yong~Jae Lee.
\newblock Visual instruction tuning.
\newblock \emph{Advances in Neural Information Processing Systems (NeurIPS)}, 2024{\natexlab{e}}.

\bibitem[Liu et~al.(2025{\natexlab{b}})Liu, Ou, Song, Qu, Lam, Xiong, Chen, Neubig, and Yue]{liuharnessing}
Junpeng Liu, Tianyue Ou, Yifan Song, Yuxiao Qu, Wai Lam, Chenyan Xiong, Wenhu Chen, Graham Neubig, and Xiang Yue.
\newblock Harnessing webpage uis for text-rich visual understanding.
\newblock In \emph{Proceedings of the International Conference on Learning Representations (ICLR)}, 2025{\natexlab{b}}.

\bibitem[Liu et~al.(2025{\natexlab{c}})Liu, Liu, Broeck, and Liang]{recap}
Xuejie Liu, Anji Liu, Guy Van~den Broeck, and Yitao Liang.
\newblock Plug-and-play context feature reuse for efficient masked generation.
\newblock \emph{arXiv preprint arXiv:2505.19089}, 2025{\natexlab{c}}.

\bibitem[Liu et~al.(2024{\natexlab{f}})Liu, Duan, Zhang, Li, Zhang, Zhao, Yuan, Wang, He, Liu, et~al.]{liu2024mmbench}
Yuan Liu, Haodong Duan, Yuanhan Zhang, Bo~Li, Songyang Zhang, Wangbo Zhao, Yike Yuan, Jiaqi Wang, Conghui He, Ziwei Liu, et~al.
\newblock Mmbench: Is your multi-modal model an all-around player?
\newblock In \emph{Proceedings of the European Conference on Computer Vision (ECCV)}, 2024{\natexlab{f}}.

\bibitem[Liu et~al.(2025{\natexlab{d}})Liu, Yang, Zhang, Chen, Zou, Wei, Wang, and Zhang]{dllmcache}
Zhiyuan Liu, Yicun Yang, Yaojie Zhang, Junjie Chen, Chang Zou, Qingyuan Wei, Shaobo Wang, and Linfeng Zhang.
\newblock dllm-cache: Accelerating diffusion large language models with adaptive caching.
\newblock \emph{arXiv preprint arXiv:2506.06295}, 2025{\natexlab{d}}.

\bibitem[Lou et~al.(2024)Lou, Meng, and Ermon]{sedd}
Aaron Lou, Chenlin Meng, and Stefano Ermon.
\newblock Discrete diffusion modeling by estimating the ratios of the data distribution.
\newblock In \emph{Proceedings of the International Conference on Machine Learning (ICML)}, pages 32819--32848, 2024.

\bibitem[Luo et~al.(2024)Luo, Shi, Ge, Yang, Wang, and Shan]{luo2024open}
Zhuoyan Luo, Fengyuan Shi, Yixiao Ge, Yujiu Yang, Limin Wang, and Ying Shan.
\newblock Open-magvit2: An open-source project toward democratizing auto-regressive visual generation.
\newblock \emph{arXiv preprint arXiv:2409.04410}, 2024.

\bibitem[Ma et~al.(2025)Ma, Yu, Fang, and Wang]{dkvcache}
Xinyin Ma, Runpeng Yu, Gongfan Fang, and Xinchao Wang.
\newblock dkv-cache: The cache for diffusion language models.
\newblock \emph{arXiv preprint arXiv:2505.15781}, 2025.

\bibitem[Ma et~al.(2024)Ma, Liu, Chen, Liu, Wu, Wu, Pan, Xie, Zhang, Zhao, et~al.]{ma2024janusflow}
Yiyang Ma, Xingchao Liu, Xiaokang Chen, Wen Liu, Chengyue Wu, Zhiyu Wu, Zizheng Pan, Zhenda Xie, Haowei Zhang, Liang Zhao, et~al.
\newblock Janusflow: Harmonizing autoregression and rectified flow for unified multimodal understanding and generation.
\newblock \emph{arXiv preprint arXiv:2411.07975}, 2024.

\bibitem[Mao et~al.(2025)Mao, Yang, and Shou]{mao2025unirl}
Weijia Mao, Zhenheng Yang, and Mike~Zheng Shou.
\newblock Unirl: Self-improving unified multimodal models via supervised and reinforcement learning.
\newblock \emph{arXiv preprint arXiv:2505.23380}, 2025.

\bibitem[Nie et~al.(2025)Nie, Zhu, You, Zhang, Ou, Hu, Zhou, Lin, Wen, and Li]{llada}
Shen Nie, Fengqi Zhu, Zebin You, Xiaolu Zhang, Jingyang Ou, Jun Hu, Jun Zhou, Yankai Lin, Ji-Rong Wen, and Chongxuan Li.
\newblock Large language diffusion models.
\newblock \emph{Proceedings of the International Conference on Machine Learning (ICML)}, 2025.

\bibitem[OpenAI(2025)]{gpt4o}
OpenAI.
\newblock Gpt-image-1.
\newblock \url{https://openai.com/index/introducing-4o-image-generation/}, 2025.

\bibitem[Oquab et~al.(2023)Oquab, Darcet, Moutakanni, Vo, Szafraniec, Khalidov, Fernandez, Haziza, Massa, El-Nouby, et~al.]{oquab2023dinov2}
Maxime Oquab, Timoth{\'e}e Darcet, Th{\'e}o Moutakanni, Huy Vo, Marc Szafraniec, Vasil Khalidov, Pierre Fernandez, Daniel Haziza, Francisco Massa, Alaaeldin El-Nouby, et~al.
\newblock Dinov2: Learning robust visual features without supervision.
\newblock \emph{Transactions on Machine Learning Research (TMLR)}, 2023.

\bibitem[Ou et~al.(2025)Ou, Nie, Xue, Zhu, Sun, Li, and Li]{radd}
Jingyang Ou, Shen Nie, Kaiwen Xue, Fengqi Zhu, Jiacheng Sun, Zhenguo Li, and Chongxuan Li.
\newblock Your absorbing discrete diffusion secretly models the conditional distributions of clean data.
\newblock In \emph{Proceedings of the International Conference on Learning Representations (ICLR)}, 2025.

\bibitem[Pan et~al.(2025)Pan, Shukla, Singh, Zhao, Mishra, Wang, Xu, Chen, Li, Juefei-Xu, Hou, and Xie]{pan2025transfer}
Xichen Pan, Satya~Narayan Shukla, Aashu Singh, Zhuokai Zhao, Shlok~Kumar Mishra, Jialiang Wang, Zhiyang Xu, Jiuhai Chen, Kunpeng Li, Felix Juefei-Xu, Ji~Hou, and Saining Xie.
\newblock Transfer between modalities with metaqueries.
\newblock \emph{arXiv preprint arXiv:2504.06256}, 2025.

\bibitem[Park et~al.(2025)Park, Lai, Hayakawa, Takida, and Mitsufuji]{jumpyoursteps}
Yong-Hyun Park, Chieh-Hsin Lai, Satoshi Hayakawa, Yuhta Takida, and Yuki Mitsufuji.
\newblock Jump your steps: Optimizing sampling schedule of discrete diffusion models.
\newblock In \emph{Proceedings of the International Conference on Learning Representations (ICLR)}, 2025.

\bibitem[Patil et~al.(2024)Patil, Berman, Rombach, and von Platen]{patil2024amused}
Suraj Patil, William Berman, Robin Rombach, and Patrick von Platen.
\newblock amused: An open muse reproduction.
\newblock \emph{arXiv preprint arXiv:2401.01808}, 2024.

\bibitem[Podell et~al.(2024)Podell, English, Lacey, Blattmann, Dockhorn, M{\"u}ller, Penna, and Rombach]{2023SDXL}
Dustin Podell, Zion English, Kyle Lacey, Andreas Blattmann, Tim Dockhorn, Jonas M{\"u}ller, Joe Penna, and Robin Rombach.
\newblock {SDXL}: Improving latent diffusion models for high-resolution image synthesis.
\newblock In \emph{Proceedings of the International Conference on Learning Representations (ICLR)}, 2024.

\bibitem[Pu et~al.(2025)Pu, Zhuo, Zhu, Xie, Zhang, Chen, Gao, Qiao, Dong, and Liu]{pu2025lumina}
Yuandong Pu, Le~Zhuo, Kaiwen Zhu, Liangbin Xie, Wenlong Zhang, Xiangyu Chen, Peng Gao, Yu~Qiao, Chao Dong, and Yihao Liu.
\newblock Lumina-omnilv: A unified multimodal framework for general low-level vision.
\newblock \emph{arXiv preprint arXiv:2504.04903}, 2025.

\bibitem[Qin et~al.(2025)Qin, Zhuo, Xin, Du, Li, Fu, Lu, Yuan, Li, Liu, et~al.]{LuminaImage2024}
Qi~Qin, Le~Zhuo, Yi~Xin, Ruoyi Du, Zhen Li, Bin Fu, Yiting Lu, Jiakang Yuan, Xinyue Li, Dongyang Liu, et~al.
\newblock Lumina-image 2.0: A unified and efficient image generative framework.
\newblock \emph{arXiv preprint arXiv:2503.21758}, 2025.

\bibitem[Qu et~al.(2024)Qu, Zhang, Liu, Wang, Jiang, Gao, Ye, Du, Yuan, and Wu]{qu2024tokenflow}
Liao Qu, Huichao Zhang, Yiheng Liu, Xu~Wang, Yi~Jiang, Yiming Gao, Hu~Ye, Daniel~K Du, Zehuan Yuan, and Xinglong Wu.
\newblock Tokenflow: Unified image tokenizer for multimodal understanding and generation.
\newblock \emph{arXiv preprint arXiv:2412.03069}, 2024.

\bibitem[Radford et~al.(2021)Radford, Kim, Hallacy, Ramesh, Goh, Agarwal, Sastry, Askell, Mishkin, Clark, et~al.]{radford2021learning}
Alec Radford, Jong~Wook Kim, Chris Hallacy, Aditya Ramesh, Gabriel Goh, Sandhini Agarwal, Girish Sastry, Amanda Askell, Pamela Mishkin, Jack Clark, et~al.
\newblock Learning transferable visual models from natural language supervision.
\newblock In \emph{Proceedings of the International Conference on Machine Learning (ICML)}, 2021.

\bibitem[Razzhigaev et~al.(2023)Razzhigaev, Shakhmatov, Maltseva, Arkhipkin, Pavlov, Ryabov, Kuts, Panchenko, Kuznetsov, and Dimitrov]{razzhigaev2023kandinsky}
Anton Razzhigaev, Arseniy Shakhmatov, Anastasia Maltseva, Vladimir Arkhipkin, Igor Pavlov, Ilya Ryabov, Angelina Kuts, Alexander Panchenko, Andrey Kuznetsov, and Denis Dimitrov.
\newblock Kandinsky: An improved text-to-image synthesis with image prior and latent diffusion.
\newblock In \emph{Proceedings of the Conference on Empirical Methods in Natural Language Processing (EMNLP)}, 2023.

\bibitem[Rombach et~al.(2022)Rombach, Blattmann, Lorenz, Esser, and Ommer]{rombach2022high}
Robin Rombach, Andreas Blattmann, Dominik Lorenz, Patrick Esser, and Bj{\"o}rn Ommer.
\newblock High-resolution image synthesis with latent diffusion models.
\newblock In \emph{Proceedings of the IEEE Conference on Computer Vision and Pattern Recognition (CVPR)}, 2022.

\bibitem[Sahoo et~al.(2024)Sahoo, Arriola, Schiff, Gokaslan, Marroquin, Chiu, Rush, and Kuleshov]{mdlm}
Subham Sahoo, Marianne Arriola, Yair Schiff, Aaron Gokaslan, Edgar Marroquin, Justin Chiu, Alexander Rush, and Volodymyr Kuleshov.
\newblock Simple and effective masked diffusion language models.
\newblock \emph{Advances in Neural Information Processing Systems (NeurIPS)}, 2024.

\bibitem[Schuhmann et~al.(2021)Schuhmann, Kaczmarczyk, Komatsuzaki, Katta, Vencu, Beaumont, Jitsev, Coombes, and Mullis]{schuhmann2021laion}
Christoph Schuhmann, Robert Kaczmarczyk, Aran Komatsuzaki, Aarush Katta, Richard Vencu, Romain Beaumont, Jenia Jitsev, Theo Coombes, and Clayton Mullis.
\newblock Laion-400m: Open dataset of clip-filtered 400 million image-text pairs.
\newblock In \emph{Advances in Neural Information Processing Systems Workshops (NeurIPS Workshops)}, 2021.

\bibitem[Sun et~al.(2024{\natexlab{a}})Sun, Jiang, Chen, Zhang, Peng, Luo, and Yuan]{sun2024autoregressive}
Peize Sun, Yi~Jiang, Shoufa Chen, Shilong Zhang, Bingyue Peng, Ping Luo, and Zehuan Yuan.
\newblock Autoregressive model beats diffusion: Llama for scalable image generation.
\newblock \emph{arXiv preprint arXiv:2406.06525}, 2024{\natexlab{a}}.

\bibitem[Sun et~al.(2024{\natexlab{b}})Sun, Cui, Zhang, Zhang, Yu, Wang, Rao, Liu, Huang, and Wang]{emu2}
Quan Sun, Yufeng Cui, Xiaosong Zhang, Fan Zhang, Qiying Yu, Yueze Wang, Yongming Rao, Jingjing Liu, Tiejun Huang, and Xinlong Wang.
\newblock Generative multimodal models are in-context learners.
\newblock In \emph{Proceedings of the IEEE Conference on Computer Vision and Pattern Recognition (CVPR)}, 2024{\natexlab{b}}.

\bibitem[Tan et~al.(2024)Tan, Liu, Yang, Xue, and Wang]{tan2024ominicontrol}
Zhenxiong Tan, Songhua Liu, Xingyi Yang, Qiaochu Xue, and Xinchao Wang.
\newblock Ominicontrol: Minimal and universal control for diffusion transformer.
\newblock \emph{arXiv preprint arXiv:2411.15098}, 2024.

\bibitem[Team(2024)]{team2024chameleon}
Chameleon Team.
\newblock Chameleon: Mixed-modal early-fusion foundation models.
\newblock \emph{arXiv preprint arXiv:2405.09818}, 2024.

\bibitem[Wang et~al.(2024{\natexlab{a}})Wang, Spinelli, Wang, Bai, Qin, and Chen]{wang2024instantstyle}
Haofan Wang, Matteo Spinelli, Qixun Wang, Xu~Bai, Zekui Qin, and Anthony Chen.
\newblock Instantstyle: Free lunch towards style-preserving in text-to-image generation.
\newblock \emph{arXiv preprint arXiv:2404.02733}, 2024{\natexlab{a}}.

\bibitem[Wang et~al.(2024{\natexlab{b}})Wang, Bai, Tan, Wang, Fan, Bai, Chen, Liu, Wang, Ge, Fan, Dang, Du, Ren, Men, Liu, Zhou, Zhou, and Lin]{Qwen2-VL}
Peng Wang, Shuai Bai, Sinan Tan, Shijie Wang, Zhihao Fan, Jinze Bai, Keqin Chen, Xuejing Liu, Jialin Wang, Wenbin Ge, Yang Fan, Kai Dang, Mengfei Du, Xuancheng Ren, Rui Men, Dayiheng Liu, Chang Zhou, Jingren Zhou, and Junyang Lin.
\newblock Qwen2-vl: Enhancing vision-language model's perception of the world at any resolution.
\newblock \emph{arXiv preprint arXiv:2409.12191}, 2024{\natexlab{b}}.

\bibitem[Wang et~al.(2025{\natexlab{a}})Wang, Gao, Gu, Pu, Cui, Wei, Liu, Jing, Ye, Shao, et~al.]{wang2025internvl3_5}
Weiyun Wang, Zhangwei Gao, Lixin Gu, Hengjun Pu, Long Cui, Xingguang Wei, Zhaoyang Liu, Linglin Jing, Shenglong Ye, Jie Shao, et~al.
\newblock Internvl3.5: Advancing open-source multimodal models in versatility, reasoning, and efficiency.
\newblock \emph{arXiv preprint arXiv:2508.18265}, 2025{\natexlab{a}}.

\bibitem[Wang et~al.(2024{\natexlab{c}})Wang, Zhang, Luo, Sun, Cui, Wang, Zhang, Wang, Li, Yu, et~al.]{wang2024emu3}
Xinlong Wang, Xiaosong Zhang, Zhengxiong Luo, Quan Sun, Yufeng Cui, Jinsheng Wang, Fan Zhang, Yueze Wang, Zhen Li, Qiying Yu, et~al.
\newblock Emu3: Next-token prediction is all you need.
\newblock \emph{arXiv preprint arXiv:2409.18869}, 2024{\natexlab{c}}.

\bibitem[Wang et~al.(2025{\natexlab{b}})Wang, Li, Zang, Zhou, Bu, Wang, Lu, Jin, and Wang]{wang2025pref}
Yibin Wang, Zhimin Li, Yuhang Zang, Yujie Zhou, Jiazi Bu, Chunyu Wang, Qinglin Lu, Cheng Jin, and Jiaqi Wang.
\newblock Pref-grpo: Pairwise preference reward-based grpo for stable text-to-image reinforcement learning.
\newblock \emph{arXiv preprint arXiv:2508.20751}, 2025{\natexlab{b}}.

\bibitem[Wei et~al.(2024)Wei, Xiong, Ren, Du, Zhang, and Chen]{wei2024omniedit}
Cong Wei, Zheyang Xiong, Weiming Ren, Xeron Du, Ge~Zhang, and Wenhu Chen.
\newblock Omniedit: Building image editing generalist models through specialist supervision.
\newblock In \emph{Proceedings of the International Conference on Learning Representations (ICLR)}, 2024.

\bibitem[Wei et~al.(2025)Wei, Zhang, Wang, Wei, Guo, and Zhang]{wei2025tiif}
Xinyu Wei, Jinrui Zhang, Zeqing Wang, Hongyang Wei, Zhen Guo, and Lei Zhang.
\newblock Tiif-bench: How does your t2i model follow your instructions?
\newblock \emph{arXiv preprint arXiv:2506.02161}, 2025.

\bibitem[Wu et~al.(2024{\natexlab{a}})Wu, Chen, Wu, Ma, Liu, Pan, Liu, Xie, Yu, Ruan, et~al.]{wu2024janus}
Chengyue Wu, Xiaokang Chen, Zhiyu Wu, Yiyang Ma, Xingchao Liu, Zizheng Pan, Wen Liu, Zhenda Xie, Xingkai Yu, Chong Ruan, et~al.
\newblock Janus: Decoupling visual encoding for unified multimodal understanding and generation.
\newblock \emph{arXiv preprint arXiv:2410.13848}, 2024{\natexlab{a}}.

\bibitem[Wu et~al.(2025{\natexlab{a}})Wu, Zhang, Xue, Liu, Diao, Zhu, Luo, Han, and Xie]{fastdllm}
Chengyue Wu, Hao Zhang, Shuchen Xue, Zhijian Liu, Shizhe Diao, Ligeng Zhu, Ping Luo, Song Han, and Enze Xie.
\newblock Fast-dllm: Training-free acceleration of diffusion llm by enabling kv cache and parallel decoding.
\newblock \emph{arXiv preprint arXiv:2505.22618}, 2025{\natexlab{a}}.

\bibitem[Wu et~al.(2025{\natexlab{b}})Wu, Zheng, Yan, Xiao, Luo, Wang, Li, Jiang, Liu, Zhou, et~al.]{wu2025omnigen2}
Chenyuan Wu, Pengfei Zheng, Ruiran Yan, Shitao Xiao, Xin Luo, Yueze Wang, Wanli Li, Xiyan Jiang, Yexin Liu, Junjie Zhou, et~al.
\newblock Omnigen2: Exploration to advanced multimodal generation.
\newblock \emph{arXiv preprint arXiv:2506.18871}, 2025{\natexlab{b}}.

\bibitem[Wu et~al.(2024{\natexlab{b}})Wu, Jiang, Ma, Liu, Zhao, Yuan, Bai, and Bai]{liquid}
Junfeng Wu, Yi~Jiang, Chuofan Ma, Yuliang Liu, Hengshuang Zhao, Zehuan Yuan, Song Bai, and Xiang Bai.
\newblock Liquid: Language models are scalable and unified multi-modal generators, 2024{\natexlab{b}}.

\bibitem[Wu et~al.(2025{\natexlab{c}})Wu, Zhang, Chen, Tang, Li, Fang, Zhu, Xie, Yin, Yi, et~al.]{wu2024vila}
Yecheng Wu, Zhuoyang Zhang, Junyu Chen, Haotian Tang, Dacheng Li, Yunhao Fang, Ligeng Zhu, Enze Xie, Hongxu Yin, Li~Yi, et~al.
\newblock Vila-u: a unified foundation model integrating visual understanding and generation.
\newblock \emph{Proceedings of the International Conference on Learning Representations (ICLR)}, 2025{\natexlab{c}}.

\bibitem[Xiao et~al.(2024)Xiao, Wang, Zhou, Yuan, Xing, Yan, Wang, Huang, and Liu]{xiao2024omnigen}
Shitao Xiao, Yueze Wang, Junjie Zhou, Huaying Yuan, Xingrun Xing, Ruiran Yan, Shuting Wang, Tiejun Huang, and Zheng Liu.
\newblock Omnigen: Unified image generation.
\newblock \emph{arXiv preprint arXiv:2409.11340}, 2024.

\bibitem[Xie et~al.(2025{\natexlab{a}})Xie, Chen, Chen, Cai, Tang, Lin, Zhang, Li, Zhu, Lu, et~al.]{xie2024sana}
Enze Xie, Junsong Chen, Junyu Chen, Han Cai, Haotian Tang, Yujun Lin, Zhekai Zhang, Muyang Li, Ligeng Zhu, Yao Lu, et~al.
\newblock Sana: Efficient high-resolution image synthesis with linear diffusion transformers.
\newblock \emph{Proceedings of the International Conference on Learning Representations (ICLR)}, 2025{\natexlab{a}}.

\bibitem[Xie et~al.(2025{\natexlab{b}})Xie, Chen, Zhao, Yu, Zhu, Wu, Lin, Zhang, Li, Chen, et~al.]{xie2025sana}
Enze Xie, Junsong Chen, Yuyang Zhao, Jincheng Yu, Ligeng Zhu, Chengyue Wu, Yujun Lin, Zhekai Zhang, Muyang Li, Junyu Chen, et~al.
\newblock Sana 1.5: Efficient scaling of training-time and inference-time compute in linear diffusion transformer.
\newblock \emph{Proceedings of the International Conference on Machine Learning (ICML)}, 2025{\natexlab{b}}.

\bibitem[Xie et~al.(2025{\natexlab{c}})Xie, Mao, Bai, Zhang, Wang, Lin, Gu, Chen, Yang, and Shou]{showo}
Jinheng Xie, Weijia Mao, Zechen Bai, David~Junhao Zhang, Weihao Wang, Kevin~Qinghong Lin, Yuchao Gu, Zhijie Chen, Zhenheng Yang, and Mike~Zheng Shou.
\newblock Show-o: One single transformer to unify multimodal understanding and generation.
\newblock In \emph{Proceedings of the International Conference on Learning Representations (ICLR)}, 2025{\natexlab{c}}.

\bibitem[Xin et~al.(2025{\natexlab{a}})Xin, Yan, Qin, Li, Liu, Li, Huang, Zhou, Zhang, Zhuo, et~al.]{xin2025lumina}
Yi~Xin, Juncheng Yan, Qi~Qin, Zhen Li, Dongyang Liu, Shicheng Li, Victor Shea-Jay Huang, Yupeng Zhou, Renrui Zhang, Le~Zhuo, et~al.
\newblock Lumina-mgpt 2.0: Stand-alone autoregressive image modeling.
\newblock \emph{arXiv preprint arXiv:2507.17801}, 2025{\natexlab{a}}.

\bibitem[Xin et~al.(2025{\natexlab{b}})Xin, Zhuo, Qin, Luo, Cao, Fu, He, Li, Zhai, Liu, et~al.]{xin2025resurrect}
Yi~Xin, Le~Zhuo, Qi~Qin, Siqi Luo, Yuewen Cao, Bin Fu, Yangfan He, Hongsheng Li, Guangtao Zhai, Xiaohong Liu, et~al.
\newblock Resurrect mask autoregressive modeling for efficient and scalable image generation.
\newblock \emph{arXiv preprint arXiv:2507.13032}, 2025{\natexlab{b}}.

\bibitem[Yang et~al.(2025)Yang, Tian, Li, Zhang, Shen, Tong, and Wang]{yang2025mmada}
Ling Yang, Ye~Tian, Bowen Li, Xinchen Zhang, Ke~Shen, Yunhai Tong, and Mengdi Wang.
\newblock Mmada: Multimodal large diffusion language models.
\newblock \emph{Advances in Neural Information Processing Systems (NeurIPS)}, 2025.

\bibitem[Yang et~al.(2022)Yang, Wu, Shi, Lao, Gong, Cao, Wang, and Yang]{yang2022maniqa}
Sidi Yang, Tianhe Wu, Shuwei Shi, Shanshan Lao, Yuan Gong, Mingdeng Cao, Jiahao Wang, and Yujiu Yang.
\newblock Maniqa: Multi-dimension attention network for no-reference image quality assessment.
\newblock In \emph{Proceedings of the IEEE Conference on Computer Vision and Pattern Recognition (CVPR)}, 2022.

\bibitem[Ye et~al.(2025{\natexlab{a}})Ye, Xie, Zheng, Gao, Wu, Jiang, Li, and Kong]{dream}
Jiacheng Ye, Zhihui Xie, Lin Zheng, Jiahui Gao, Zirui Wu, Xin Jiang, Zhenguo Li, and Lingpeng Kong.
\newblock Dream 7{B}, 2025{\natexlab{a}}.
\newblock URL \url{https://hkunlp.github.io/blog/2025/dream}.

\bibitem[Ye et~al.(2025{\natexlab{b}})Ye, He, Li, Lin, Yuan, Yan, Hou, and Yuan]{ye2025imgedit}
Yang Ye, Xianyi He, Zongjian Li, Bin Lin, Shenghai Yuan, Zhiyuan Yan, Bohan Hou, and Li~Yuan.
\newblock Imgedit: A unified image editing dataset and benchmark.
\newblock \emph{arXiv preprint arXiv:2505.20275}, 2025{\natexlab{b}}.

\bibitem[Yi et~al.(2024)Yi, Li, Xin, and Li]{yi2024towards}
Mingyang Yi, Aoxue Li, Yi~Xin, and Zhenguo Li.
\newblock Towards understanding the working mechanism of text-to-image diffusion model.
\newblock \emph{Advances in Neural Information Processing Systems (NeurIPS)}, 2024.

\bibitem[Yin et~al.(2024)Yin, Fu, Zhao, Li, Sun, Xu, and Chen]{yin2024survey}
Shukang Yin, Chaoyou Fu, Sirui Zhao, Ke~Li, Xing Sun, Tong Xu, and Enhong Chen.
\newblock A survey on multimodal large language models.
\newblock \emph{National Science Review}, 2024.

\bibitem[You et~al.(2025)You, Nie, Zhang, Hu, Zhou, Lu, Wen, and Li]{lladav}
Zebin You, Shen Nie, Xiaolu Zhang, Jun Hu, Jun Zhou, Zhiwu Lu, Ji-Rong Wen, and Chongxuan Li.
\newblock {LLaDA-V}: Large language diffusion models with visual instruction tuning.
\newblock \emph{arXiv preprint arXiv:2505.16933}, 2025.

\bibitem[Yu et~al.(2025{\natexlab{a}})Yu, Chow, Yue, Pan, Wu, Wan, Li, Tang, Zhang, and Zhuang]{yu2025anyedit}
Qifan Yu, Wei Chow, Zhongqi Yue, Kaihang Pan, Yang Wu, Xiaoyang Wan, Juncheng Li, Siliang Tang, Hanwang Zhang, and Yueting Zhuang.
\newblock Anyedit: Mastering unified high-quality image editing for any idea.
\newblock In \emph{Proceedings of the IEEE Conference on Computer Vision and Pattern Recognition (CVPR)}, 2025{\natexlab{a}}.

\bibitem[Yu et~al.(2025{\natexlab{b}})Yu, Ma, and Wang]{dimple}
Runpeng Yu, Xinyin Ma, and Xinchao Wang.
\newblock Dimple: Discrete diffusion multimodal large language model with parallel decoding.
\newblock \emph{arXiv preprint arXiv:2505.16990}, 2025{\natexlab{b}}.

\bibitem[Yue et~al.(2024)Yue, Ni, Zhang, Zheng, Liu, Zhang, Stevens, Jiang, Ren, Sun, et~al.]{yue2024mmmu}
Xiang Yue, Yuansheng Ni, Kai Zhang, Tianyu Zheng, Ruoqi Liu, Ge~Zhang, Samuel Stevens, Dongfu Jiang, Weiming Ren, Yuxuan Sun, et~al.
\newblock Mmmu: A massive multi-discipline multimodal understanding and reasoning benchmark for expert agi.
\newblock In \emph{Proceedings of the IEEE Conference on Computer Vision and Pattern Recognition (CVPR)}, 2024.

\bibitem[Zhang et~al.(2024)Zhang, Hu, Xu, Yan, Xu, Jin, Zhang, and Huang]{zhang2024tinychart}
Liang Zhang, Anwen Hu, Haiyang Xu, Ming Yan, Yichen Xu, Qin Jin, Ji~Zhang, and Fei Huang.
\newblock Tinychart: Efficient chart understanding with visual token merging and program-of-thoughts learning.
\newblock \emph{arXiv preprint arXiv:2404.16635}, 2024.

\bibitem[Zhang et~al.(2023)Zhang, Rao, and Agrawala]{zhang2023adding}
Lvmin Zhang, Anyi Rao, and Maneesh Agrawala.
\newblock Adding conditional control to text-to-image diffusion models.
\newblock In \emph{Proceedings of the IEEE International Conference on Computer Vision (ICCV)}, 2023.

\bibitem[Zhang et~al.(2025)Zhang, Wei, Jiang, Zhang, Guo, Tong, Liu, Zhou, Wei, Zhang, et~al.]{zhang2024mavis}
Renrui Zhang, Xinyu Wei, Dongzhi Jiang, Yichi Zhang, Ziyu Guo, Chengzhuo Tong, Jiaming Liu, Aojun Zhou, Bin Wei, Shanghang Zhang, et~al.
\newblock Mavis: Mathematical visual instruction tuning.
\newblock \emph{Proceedings of the International Conference on Learning Representations (ICLR)}, 2025.

\bibitem[Zhao et~al.(2024)Zhao, Ma, Chen, Si, Wu, An, Yu, Zhang, Li, and Chang]{zhao2024ultraedit}
Haozhe Zhao, Xiaojian~Shawn Ma, Liang Chen, Shuzheng Si, Rujie Wu, Kaikai An, Peiyu Yu, Minjia Zhang, Qing Li, and Baobao Chang.
\newblock Ultraedit: Instruction-based fine-grained image editing at scale.
\newblock \emph{Advances in Neural Information Processing Systems (NeurIPS)}, 2024.

\bibitem[Zheng et~al.(2024)Zheng, Feng, Si, She, Lin, Jiang, and Wang]{zheng2024multimodal}
Mingyu Zheng, Xinwei Feng, Qingyi Si, Qiaoqiao She, Zheng Lin, Wenbin Jiang, and Weiping Wang.
\newblock Multimodal table understanding.
\newblock In \emph{Proceedings of the Annual Meeting of the Association for Computational Linguistics (ACL)}, pages 9102--9124, 2024.

\bibitem[Zhou et~al.(2025)Zhou, YU, Babu, Tirumala, Yasunaga, Shamis, Kahn, Ma, Zettlemoyer, and Levy]{transfusion}
Chunting Zhou, LILI YU, Arun Babu, Kushal Tirumala, Michihiro Yasunaga, Leonid Shamis, Jacob Kahn, Xuezhe Ma, Luke Zettlemoyer, and Omer Levy.
\newblock Transfusion: Predict the next token and diffuse images with one multi-modal model.
\newblock In \emph{Proceedings of the International Conference on Learning Representations (ICLR)}, 2025.

\bibitem[Zhu et~al.(2025{\natexlab{a}})Zhu, Wang, Nie, Zhang, Wu, Hu, Zhou, Chen, Lin, Wen, and Li]{llada1.5}
Fengqi Zhu, Rongzhen Wang, Shen Nie, Xiaolu Zhang, Chunwei Wu, Jun Hu, Jun Zhou, Jianfei Chen, Yankai Lin, Ji-Rong Wen, and Chongxuan Li.
\newblock {LLaDA} 1.5: Variance-reduced preference optimization for large language diffusion models.
\newblock \emph{arXiv preprint arXiv:2505.19223}, 2025{\natexlab{a}}.

\bibitem[Zhu et~al.(2025{\natexlab{b}})Zhu, Wang, Chen, Liu, Ye, Gu, Tian, Duan, Su, Shao, et~al.]{zhu2025internvl3}
Jinguo Zhu, Weiyun Wang, Zhe Chen, Zhaoyang Liu, Shenglong Ye, Lixin Gu, Hao Tian, Yuchen Duan, Weijie Su, Jie Shao, et~al.
\newblock Internvl3: Exploring advanced training and test-time recipes for open-source multimodal models.
\newblock \emph{arXiv preprint arXiv:2504.10479}, 2025{\natexlab{b}}.

\bibitem[Zhu et~al.(2024)Zhu, Zhu, Liu, Ou, Mou, and Tang]{zhu2024llava}
Yichen Zhu, Minjie Zhu, Ning Liu, Zhicai Ou, Xiaofeng Mou, and Jian Tang.
\newblock Llava-phi: Efficient multi-modal assistant with small language model.
\newblock \emph{arXiv preprint arXiv:2401.02330}, 2024.

\bibitem[Zhuo et~al.(2024)Zhuo, Du, Xiao, Li, Liu, Huang, Liu, Zhao, Wang, Ma, et~al.]{zhuo2024lumina}
Le~Zhuo, Ruoyi Du, Han Xiao, Yangguang Li, Dongyang Liu, Rongjie Huang, Wenze Liu, Lirui Zhao, Fu-Yun Wang, Zhanyu Ma, et~al.
\newblock Lumina-next: Making lumina-t2x stronger and faster with next-dit.
\newblock \emph{Advances in Neural Information Processing Systems (NeurIPS)}, 2024.

\bibitem[Zhuo et~al.(2025)Zhuo, Zhao, Paul, Liao, Zhang, Xin, Gao, Elhoseiny, and Li]{zhuo2025reflection}
Le~Zhuo, Liangbing Zhao, Sayak Paul, Yue Liao, Renrui Zhang, Yi~Xin, Peng Gao, Mohamed Elhoseiny, and Hongsheng Li.
\newblock From reflection to perfection: Scaling inference-time optimization for text-to-image diffusion models via reflection tuning.
\newblock \emph{Proceedings of the IEEE International Conference on Computer Vision (ICCV)}, 2025.

\end{thebibliography}

\end{document}